\documentclass{article}
\usepackage{caption} 

\PassOptionsToPackage{numbers, sort&compress}{natbib}

\usepackage[main, final]{neurips_2026}

\usepackage[utf8]{inputenc} %
\usepackage[T1]{fontenc}    %
\usepackage{graphicx}       %
\usepackage[colorlinks=true,linkcolor=ForestGreen,citecolor=ForestGreen,urlcolor=NavyBlue]{hyperref}  %
\usepackage[dvipsnames]{xcolor}  %
\usepackage{url}            %
\usepackage{booktabs}       %
\usepackage{amsmath}        %
\usepackage{amsfonts}       %

\definecolor{LeadRed}{rgb}{0.78, 0.07, 0.27}      %
\definecolor{AdviseTeal}{rgb}{0.00, 0.50, 0.50}   %
\newcommand{\markersize}{\fontsize{10}{10}\selectfont}
\newcommand{\equalcontrib}{\raisebox{0.4ex}{\markersize$\ast$}}
\newcommand{\leadmark}{\raisebox{0.4ex}{\markersize\textcolor{LeadRed}{$\boldsymbol{\ast}$}}}
\newcommand{\advisemark}{\raisebox{0.4ex}{\markersize\textcolor{AdviseTeal}{$\boldsymbol{\dagger}$}}}
\newcommand{\aff}[1]{\textsuperscript{#1}}
\usepackage{amssymb}        %
\usepackage{nicefrac}       %
\usepackage{microtype}      %
\usepackage{xspace}         %
\usepackage[capitalize]{cleveref}
\crefname{appendix}{appendix}{appendices}
\Crefname{appendix}{Appendix}{Appendices}
\usepackage{multirow}       %
\usepackage{colortbl}       %
\usepackage{tikz}           %
\usetikzlibrary{arrows.meta}
\usepackage{pifont}         %
\usepackage{enumitem}
\usepackage{wrapfig}        %
\usepackage{float}          %
\usepackage[skins,breakable]{tcolorbox}  %
\usepackage[font=footnotesize,labelfont=bf,aboveskip=2pt,belowskip=-10pt]{caption}

\usepackage{tabularx}  %
\newcolumntype{C}{>{\centering\arraybackslash}X}
\definecolor{wkblue}{RGB}{210, 230, 250}
\definecolor{wkgreen}{RGB}{226, 240, 217}
\definecolor{demphcolor}{RGB}{144, 144, 144}
\definecolor{jwblue}{rgb}{0.20, 0.55, 0.90}

\newcommand{\tablestyle}[2]{\setlength{\tabcolsep}{#1}\renewcommand{\arraystretch}{#2}\centering\footnotesize}
\newlength\savewidth

\makeatletter\renewcommand\paragraph{\@startsection{paragraph}{4}{\z@}{.5em\@plus1ex\@minus.2ex}{-.5em}{\normalfont\normalsize\bfseries}}
\preto\align{\small}
\expandafter\preto\csname align*\endcsname{\small}
\preto\equation{\par\nobreak\small\noindent}

\newcommand{\model}{ThinkMorph-SSL\xspace}
\newcommand{\pretext}{Pretext-270K\xspace}
\newcommand{\instruct}{Instruct-180K\xspace}

\title{ThinkMorph-SSL: Scaling Interleaved Multimodal Reasoning with Pretext Tasks}

\author{%
  Jiawei Gu\aff{1}\leadmark \quad Linjie Li\aff{1}\leadmark \quad Yiming Liu\aff{2}\equalcontrib \quad Yunzhuo Hao\aff{3}\equalcontrib \quad Huichen Wang\aff{1}\equalcontrib \\
  \bfseries Guanzheng Chen\aff{4}\equalcontrib \quad Zhichao Peng\aff{5} \quad  Luxin Xu\aff{6} \quad Xinyu Zhang\aff{6} \quad Li Luo\aff{7} \\
  \bfseries Disen Lan\aff{6} \quad Zican Hu\aff{8} \quad Mingyang Song\aff{6} \quad David Valente\aff{4} \quad Alex Jinpeng Wang\aff{5} \\
  \bfseries Yafu Li\aff{9} \quad Ganqu Cui\aff{2} \quad Zhengyuan Yang\aff{10} \quad Qizhe Xie\aff{4}\advisemark \\
  \bfseries Yejin Choi\aff{11}\advisemark \quad Ranjay Krishna\aff{1}\advisemark \quad Yu Cheng\aff{9}\advisemark \\[6pt]
  \normalfont\small
  \aff{1}University of Washington \quad \aff{2}Tsinghua University \quad \aff{3}Zhejiang University \\
  \normalfont\small \aff{4}National University of Singapore \quad \aff{5}Central South University \quad \aff{6}Fudan University \\
  \normalfont\small \aff{7}The University of Hong Kong \quad \aff{8}Nanjing University \quad \aff{9}The Chinese University of Hong Kong \\
  \normalfont\small \aff{10}University of Rochester \quad \aff{11}Stanford University \\[4pt]
  \normalfont\small
  \leadmark\,Project lead \qquad \equalcontrib\,Equal contribution \qquad \advisemark\,Equal advising
}

\begin{document}

\maketitle

\vspace{-18pt}
{\centering\small\hypersetup{urlcolor=AdviseTeal}
  \url{https://thinkmorph.github.io/ssl/}\par
}
\vspace{4pt}

\begin{abstract}
People reason with pictures as well as words. Interleaved multimodal reasoning gives models a similar way to solve problems by creating intermediate images and using them to guide further reasoning. Training models with this ability usually relies on task-specific demonstrations, which are difficult to collect.
We introduce \model, a two-stage training recipe that uses images themselves as a source of supervision. The first stage, \emph{pretext reasoning}, converts visual pretext tasks, such as image colorization and denoising, into \emph{instruction-free} interleaved traces. Models describe a corrupted or transformed image, generate a restored or canonical image, and interpret the result. The second stage uses automatically constructed demonstrations for question-driven instruction tuning.
At the same instruction-tuning budget, models trained with pretext reasoning outperform those trained with instruction tuning alone. The full recipe improves over the base model by an average of 11.33 accuracy points across twelve benchmarks.
These results show that supervision derived from images can build the visual skills needed for interleaved reasoning and improve learning from task-specific demonstrations.

\end{abstract}

\section{Introduction}
\label{sec:intro}

Sketching and imagery are basic forms of thinking, used for reasoning~\cite{goel1995sketches}, ideation~\cite{tversky2003sketches}, and problem-solving~\cite{tversky2009thinking}. Unlike language alone, they can express visuo-spatial structure directly~\cite{zhang1994representations}. An auxiliary line can simplify a geometry proof~\cite{larkin1987diagram}.
A spatial puzzle can become easier after mentally rotating an object~\cite{kosslyn1978measuring,hegarty1992mental}, and diagrams help tell stories and communicate scientific ideas. These visual representations are working states, not merely illustrations. They reorganize the problem and make later inference easier~\cite{kirsh2010thinking}. As multimodal language models mature, we increasingly ask them to solve math~\cite{lu2021inter,fu2024isobench}, scientific~\cite{wang2024charxiv,emma}, and multimodal~\cite{fu2024blink,wu2023vstar,yue2025mmmu} tasks where people naturally rely on such intermediate visual reasoning.

This motivates \emph{interleaved reasoning}: creating intermediate visual representations and using them to guide subsequent language-based inference. Unified models~\cite{bagel,thinkmorph,sensenova_u1} support this exchange natively.
However, coordinating visual and language steps requires training traces where intermediate images support the question being answered. Constructing these traces involves task-specific pipelines and data review~\cite{zebracot,shao2024visualcot}. ThinkMorph, for example, designs interleaved traces around four tasks, each with tailored visual manipulations~\cite{thinkmorph}. These pipelines can generate more examples, but covering new tasks often requires new designs, making interleaved supervision \underline{difficult to scale}.

\begin{figure}[ht]
  \centering
  \includegraphics[width=0.98\textwidth]{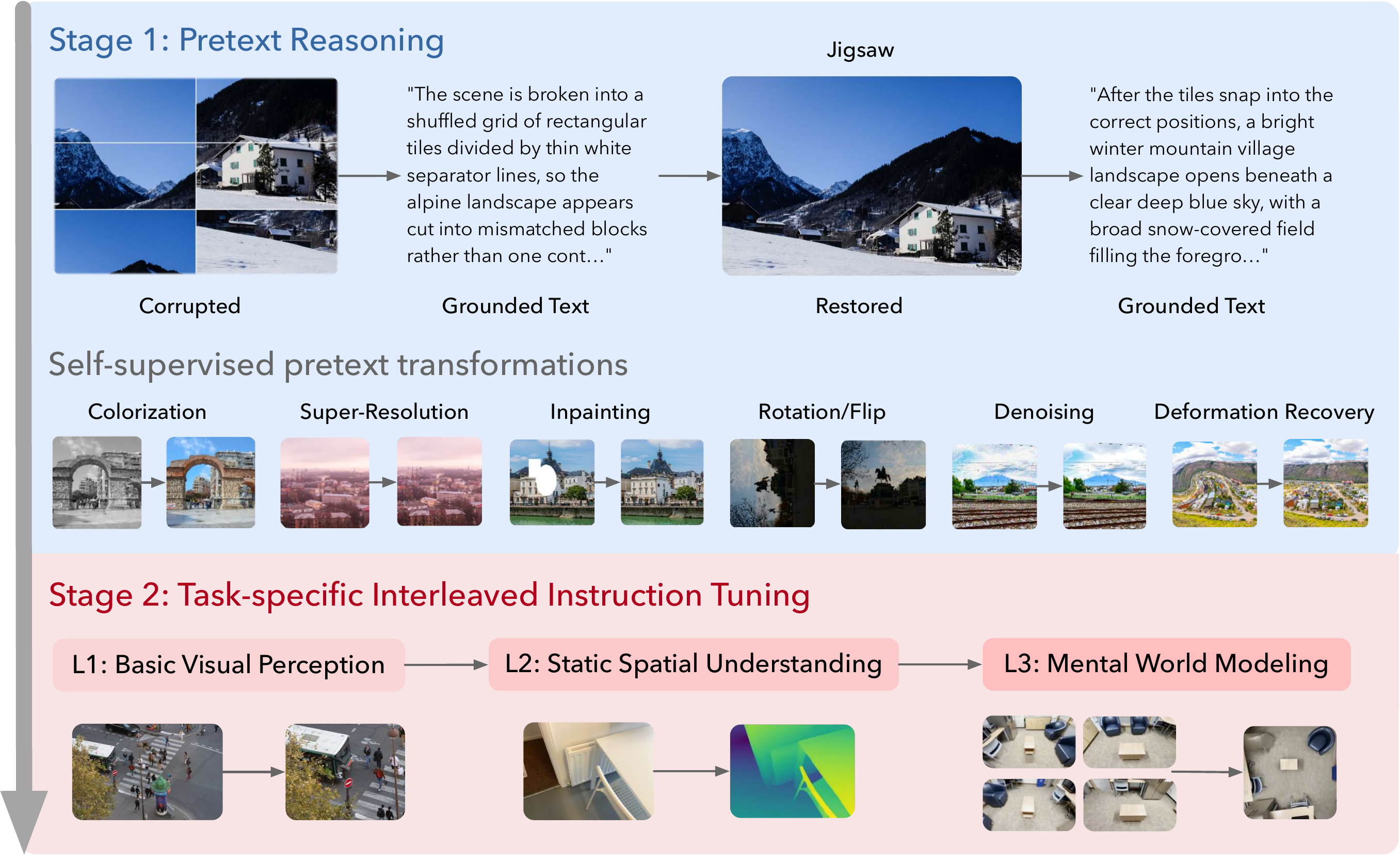}
  \caption{\textbf{Overview of the two training stages.} \emph{Stage~1 (top)} uses seven instruction-free visual pretext tasks to practice producing and reading intermediate images. \emph{Stage~2 (bottom)} uses question-driven instruction tuning to teach the model to use image content to advance reasoning across basic visual perception, static spatial understanding, and mental world modeling.}
  \label{fig:overview}
\end{figure}

\textbf{Images themselves can supply training targets for interleaved reasoning.} In classical self-supervised learning (SSL), pretext tasks such as jigsaw reconstruction~\cite{noroozi2016jigsaw} and image inpainting~\cite{pathak2016context} use an image's original structure or content as supervision.
We introduce \model, a two-stage recipe illustrated in Figure~\ref{fig:overview}. Stage~1, \textbf{\emph{pretext reasoning}}, converts seven visual pretext tasks into question-free interleaved traces, hereafter called \emph{instruction-free}, that train the model to read a transformed image, generate a restored or canonical visual state, and reason over the result (\pretext). Stage~2 uses instruction tuning on question-driven traces constructed with nine automatic, task-specific pipelines (\instruct).

Our experiments show that instruction-free pretext practice makes subsequent instruction tuning more effective. With the tuning budget held fixed, this practice adds an average of \textbf{6.49 points} in accuracy on three representative benchmarks, and its benefits generalize across unified backbones. Surprisingly, instruction-free traces still yield slightly better results than explicitly instructed ones, despite the added difficulty of inferring the task itself. Furthermore, the full two-stage recipe outperforms the baseline by \textbf{11.33 points} on average across twelve benchmarks spanning three levels: \textit{basic visual perception}, \textit{static spatial understanding}, and \textit{mental world modeling}. Improvements extend to nine benchmarks outside the instruction-tuning task families, supporting generalization to new tasks.

These findings suggest that interleaved reasoning can benefit from supervision beyond task-specific demonstrations. This brings the self-supervised learning principle to interleaved multimodal reasoning: images themselves can provide training supervision, reducing the need to identify suitable tasks and design demonstrations for them.

\section{A Two-Stage Recipe for Scaling Interleaved Reasoning}
\label{sec:method}

The recipe trains a unified multimodal backbone to generate and reuse visual states within a single autoregressive reasoning trace. Its key design is to separate scalable, task-independent supervision from the task-specific kind. 
\Cref{sec:method-setup} formalizes the trace format, and \Cref{sec:method-training} describes the two training stages, pretext reasoning and task-specific instruction tuning.

\subsection{Interleaved Reasoning Format}
\label{sec:method-setup}

We formulate interleaved multimodal reasoning as autoregressive generation over text and image segments. Given input \(x\), the model generates a trace
\begin{align*}
\mathcal{S}=(S_1,\ldots,S_K), \qquad
p_{\theta}(\mathcal{S}\mid x)=\prod_{k=1}^{K}p_{\theta}(S_k\mid x,S_{<k}),
\end{align*}
where each \(S_k\) is either a text segment \(T_k\) or an image segment \(I_k\). Image segments are enclosed by \texttt{<gen\_image>} delimiters, allowing the same model to alternate between language and image generation.
Unlike text-only chain-of-thought or tool-augmented reasoning, the model itself generates intermediate visual states and conditions later reasoning on them.

\subsection{Two Sources of Interleaved Supervision}
\label{sec:method-training}

Both sources of supervision share the same interleaved autoregressive interface (\Cref{fig:overview}). In Stage~1, pretext reasoning builds task-independent traces from unlabeled images; in Stage~2, task-specific instruction tuning uses traces that connect intermediate images to questions and answers. The model thus first learns the mechanics of producing and reading visual states, then how to use them for vision-centric reasoning.

\paragraph{Stage 1: pretext reasoning.}
This stage constructs \emph{instruction-free} traces from corrupted--clean image pairs (\Cref{fig:overview}, top): the input is a corrupted or transformed image \(I_c\) with no instruction, not even a question. The target trace is
\begin{align*}
I_c \rightarrow T_0 \rightarrow I_0 \rightarrow T_1,
\end{align*}
where \(T_0\) describes the visible evidence, \(I_0\) is the restored or canonical image, and \(T_1\) describes the resulting visual state. This can be factorized as
\begin{align*}
p_\theta(T_0,I_0,T_1\mid I_c)
=
p_\theta(T_0\mid I_c)
p_\theta(I_0\mid I_c,T_0)
p_\theta(T_1\mid I_c,T_0,I_0).
\end{align*}
We convert classical pretext tasks---jigsaw, colorization, super-resolution, masked inpainting, rotation/flip recovery, denoising, and deformation recovery---into such traces. Unlike edit-style instruction tuning, these traces do not name the operation; the model must infer the transformation from the visual input, generate the intermediate state, and interpret it.

\paragraph{Stage 2: task-specific interleaved instruction tuning.}
This stage keeps the same interface but changes the goal from restoration to task solving. Each example contains an image, a question, and a target trace with reasoning text, one or more intermediate visual states, and a final answer:
\begin{align*}
(x^{\text{img}},x^{\text{text}}) \rightarrow T_0 \rightarrow I_0 \rightarrow \cdots \rightarrow T_K.
\end{align*}
Here, the generated image is chosen because it helps answer the question. Instruction tuning therefore teaches when to construct a visual state, what state to construct, and how to use it for the final answer. These tasks progress from inspecting visible evidence to reconstructing unseen scene states. We organize them into three levels, referred to as L1, L2, and L3 throughout the paper:
\begin{itemize}
    \item \textbf{L1 (Basic Visual Perception)} concerns directly visible content, such as objects, attributes, and local evidence.
    \item \textbf{L2 (Static Spatial Understanding)} concerns the spatial organization of the visible scene, including depth, layout, relative position, and counts.
    \item \textbf{L3 (Mental World Modeling)} requires reconstructing scene states beyond the visible input, such as alternate viewpoints and cross-view correspondences.
\end{itemize}

\paragraph{Training objective.}
Both stages train the model on the same interleaved trace format, with losses applied only to the target text and image segments. The exact token-level and image-generation losses depend on the unified multimodal backbone.

\section{Training Corpus}
\label{sec:data}

We curate two datasets. \pretext, used in Stage~1, is generated automatically from raw images with programmatic corruptions and contains instruction-free Text--Image--Text traces. \instruct, used in Stage~2, relies on task-specific pipelines to generate questions, intermediate visual states, and answers across basic visual perception, static spatial understanding, and mental world modeling. Both share the same text-image trace interface but differ in supervision, and \Cref{fig:data-stats} in the Appendix summarizes their composition.

\subsection{\pretext: Instruction-Free Interleaved Reasoning Data}
\label{sec:data-pretext}

The corpus contains 270{,}896 traces, each derived from a corrupted--clean image pair (\Cref{fig:overview}, top). It is balanced across seven programmatic transformations, including \textbf{(1) Jigsaw}~\cite{noroozi2016jigsaw}: restore scrambled patches to the original layout;  \textbf{(2) Colorization}~\cite{zhang2016colorful}: recover color from grayscale;  \textbf{(3) Super-resolution}~\cite{dong2014superresolution}: reconstruct details from low resolution;  \textbf{(4) Masked inpainting}~\cite{pathak2016context}: complete missing regions; \textbf{(5) Rotation and flip recovery}~\cite{gidaris2018rotation}: restore canonical orientation; \textbf{(6) Denoising}~\cite{vincent2008denoising}: remove noise while preserving content; and \textbf{(7) Deformation recovery}~\cite{doersch2015context}: undo local geometric warps. 
Together, these tasks cover complementary forms of visual-state recovery, from appearance restoration to spatial and structural correction. Appendix~\ref{app:appendix-pretext-examples} provides examples.

\paragraph{Source images.}
We sample images from eight public datasets spanning natural images, web images, charts, screenshots, documents, indoor scenes, panoramas, and 3D environments: SA-1B~\cite{kirillov2023sam}, PixMo-Cap~\cite{deitke2024molmo}, VSI-590K~\cite{yang2025cambrian}, LLaVA-OneVision-1.5-Instruct-Data~\cite{an2025llava}, DeepVision-103K~\cite{deepvision}, Matterport3D~\cite{matterport3d,dit360}, Realsee3D~\cite{realsee3d}, and 360+x~\cite{360x}. This diversity keeps the corpus from being tied to a single image domain, and the source-agnostic pipeline accepts new image pools.

\paragraph{Automatic construction pipeline.}
For each image, we apply one of the seven transformations to create a corrupted input and use the original image as the clean target. We then synthesize the two text segments with a strong multimodal model~\cite{openai2025gpt5}: first captioning the clean image, then generating \(T_0\) for the corrupted view and \(T_1\) for the restored view using both images and the reference caption.

Importantly, the inputs carry no instruction such as ``denoise this image'' or ``restore the jigsaw.'' The model must infer the transformation from visual evidence and autoregressive context, which keeps the corpus closer to pretraining than to image-editing instruction tuning.

\begin{figure}[t]
  \centering
  \includegraphics[width=\textwidth]{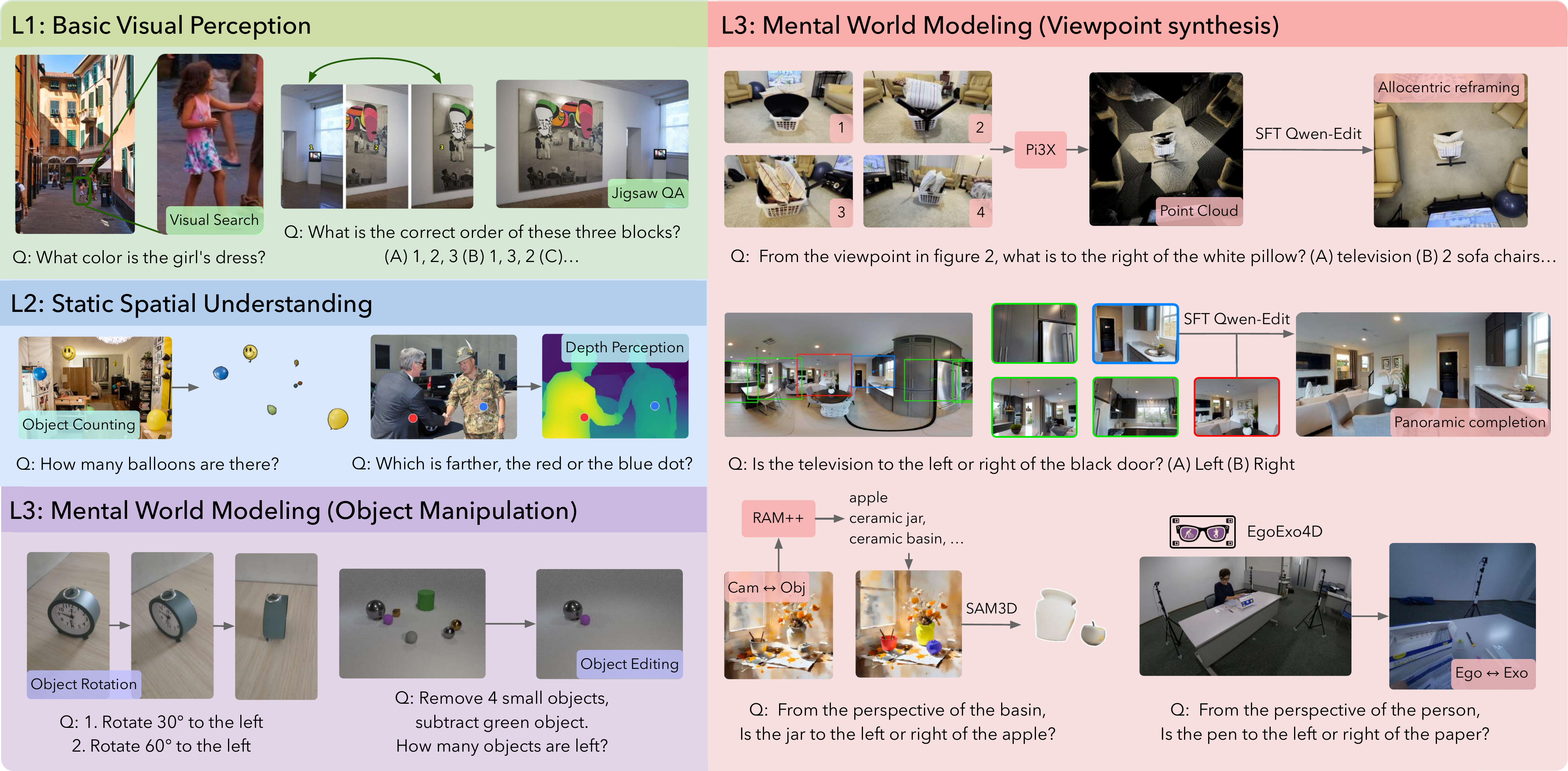}
  \caption{\textbf{\instruct construction pipelines.} The pipelines are grouped into three levels of vision-centric reasoning: basic visual perception, static spatial understanding, and mental world modeling. Each panel shows how one pipeline constructs the intermediate visual state used in an interleaved reasoning example, such as a localized crop, a reordered view, a depth map, an edited object state, or a synthesized viewpoint. Full details are provided in Appendix~\ref{app:appendix-instruct}.}
  \label{fig:instruct-pipelines}
\end{figure}

\subsection{\instruct: Task-Specific Interleaved Reasoning Data}
\label{sec:data-instruct}

Whereas \pretext teaches the mechanics of visual states, this corpus teaches how to use them for downstream tasks. It contains 180{,}490 traces from \textbf{nine construction pipelines}, of which 62{,}100 target basic visual perception, 15{,}144 static spatial understanding, and 103{,}246 mental world modeling. Each example includes an input image, a question, reasoning text, one or more intermediate or final image segments, and an answer. \Cref{fig:data-stats} in the Appendix shows the per-task distribution.

\paragraph{Task coverage.}
Basic visual perception tasks teach the model to locate and inspect visible evidence, including visual search~\cite{wu2023vstar} and Jigsaw QA~\cite{wang2025jigsawr1}. Static spatial tasks require organizing observed evidence into spatial structure, including spatial counting~\cite{gupta2019lvis} and depth reasoning~\cite{fu2024blink,zhou2025roborefer} from a single image. Mental world-modeling tasks require constructing new visual states beyond the original input, including bird's-eye view synthesis \cite{mindcube}, panorama \cite{360x,matterport3d,dit360,realsee3d,mmsi}, alternate-view rendering~\cite{sam3d}, object rotation~\cite{zhan2025actial0,yu2023mvimgnet}, and ego--exo perspective transfer~\cite{egoexo4d}.

\paragraph{Task-specific pipelines.}
Each task family uses a dedicated pipeline (Figure~\ref{fig:instruct-pipelines}), since supervision comes from different sources. Visual search uses object-dense scenes with target regions highlighted by zooms or boxes. Jigsaw QA reuses a prior pipeline~\cite{thinkmorph}, and counting derives gold counts from LVIS instance masks~\cite{gupta2019lvis}. Depth reasoning uses depth annotations or estimates. Viewpoint-centered tasks rely on geometry, including yaw and field-of-view metadata for panoramas, reconstruction and rendering for bird's-eye views, paired ego--exo frames or 3D lifting for perspective transfer, and calibrated multi-view data for object rotation. 
Whenever possible, the ground-truth answers are derived from geometry, annotations, or metadata rather than being synthesized by a language model.

\paragraph{Reasoning trace synthesis and quality control.}
After constructing the input, intermediate visual state, and ground-truth answer, we synthesize traces in a shared interleaved format. Each visual state is placed before the reasoning step that uses it, enforcing the intended causal order of first constructing or inspecting the state and then answering from it. We apply task-specific quality checks, such as yaw/FOV validation for panoramas, coverage checks for bird's-eye-view renders, IoU-based deduplication for segmentation, and multi-model judging for visual search. The shared synthesis prompt further prevents premature answer leakage, requires grounded references to generated images, and disallows describing an intermediate image before it appears.

\section{Experiments}
\label{sec:experiments}

\subsection{Setup}
\label{sec:experiments-setup}

\paragraph{Training.}
We build \model on BAGEL-7B-MoT~\cite{bagel} and InternVL-U-4B~\cite{tian2026internvlu}. Both backbones generate text and images. Stage~1 trains on \pretext, and Stage~2 continues from the Stage-1 checkpoint on \instruct. Both stages use the same system prompt with a \texttt{<gen\_image>} delimiter. Appendix~\ref{app:appendix-training} gives the full details.

\paragraph{Evaluation.}
We evaluate on twelve benchmarks at three levels: \textbf{L1} basic visual perception (V$^\star$~\cite{wu2023vstar}, HR-4K, HR-8K~\cite{hrbench}, MME-RW-L~\cite{zhang2024mme}, RealWorldQA~\cite{xai2024grok15v}), \textbf{L2} static spatial understanding (BLINK~\cite{fu2024blink}, CV-Bench~\cite{tong2024cambrian1}, SAT~\cite{ray2025satdynamicspatialaptitude}), and \textbf{L3} mental world modeling (MMSI~\cite{mmsi}, MindCube~\cite{mindcube}, OmniSpatial~\cite{omnispatial}, SpatialViz~\cite{wang2026spatialvizbenchcognitivelygroundedbenchmarkdiagnosing}). Only V$^\star$, MMSI, and MindCube share a source dataset or task family with \instruct, so the other nine are out-of-domain. Benchmark questions and templates never appear in our training data. We score every model with VLMEvalKit~\cite{vlmevalkit} and use GPT-5~\cite{openai2025gpt5} as the LLM judge. Appendix~\ref{app:appendix-eval} gives the full protocol.

\paragraph{Baselines.} We compare against closed-source APIs (GPT-5~\cite{openai2025gpt5}, Claude Sonnet~4.6~\cite{anthropic2026claudesonnet46}, Gemini~2.5 Flash~\cite{google2025gemini25flash}), open-source vision--language models (Qwen2.5-VL-7B/32B~\cite{qwen25vl}), and open-source unified models (Janus-Pro~\cite{januspro}, Chameleon~\cite{chameleon}, SenseNova-U1-8B~\cite{sensenova_u1}, LongCat-Next-A3B~\cite{longcatnext}, Show-o2-7B~\cite{Showo2}). We also run the untrained BAGEL backbone under several inference settings.

\subsection{Inference-time Effort Does Not Yield Effective Interleaved Reasoning}

A unified model might reason across text and images if it simply tries harder at inference time. We test this with more text effort, then more visual effort, and finally a ground-truth image to separate image quality from image use. The results suggest that training must teach two abilities: producing and reading intermediate images, and using them to advance reasoning.

\begin{table}[ht]
\centering
\footnotesize
\newcommand{\tabgain}[1]{{\scriptsize\textcolor{ForestGreen}{(+#1)}}}
\newcommand{\tabloss}[1]{{\scriptsize\textcolor{BrickRed}{($-$#1)}}}
\begin{minipage}[b]{0.47\linewidth}
\centering
\caption{\textbf{More text reasoning effort at inference time.} Accuracy (\%) averaged over each level. Parentheses give the change from the row above.}
\label{tab:base-reasoning}
\vspace{2pt}
\setlength{\tabcolsep}{3pt}
\begin{tabular}{@{}lrrr@{}}
\toprule
Setting & L1 & L2 & L3 \\
\midrule
No-think & 50.35 & 56.17 & 36.42 \\
Think-4k & 51.40 \tabgain{1.05} & 64.67 \tabgain{8.50} & 25.40 \tabloss{11.02} \\
Think-16k & 53.86 \tabgain{2.46} & 64.23 \tabloss{0.44} & 26.83 \tabgain{1.43} \\
\bottomrule
\end{tabular}
\end{minipage}\hfill
\begin{minipage}[b]{0.51\linewidth}
\centering
\caption{\textbf{More visual reasoning effort at inference time.} Spatial is a random 500-question subset of \instruct with ground-truth (GT) intermediate images.} 
\label{tab:base-forcing}
\vspace{2pt}
\setlength{\tabcolsep}{3pt}
\begin{tabular}{@{}lrrr@{}}
\toprule
Setting & V$^\star$ & MMSI & Spatial \\
\midrule
Text reasoning (16k) & 61.26 & 29.00 & 47.40 \\
Forced interleaved (16k) & 61.26 & 27.20 \tabloss{1.80} & 41.20 \tabloss{6.20} \\
\quad + GT image & -- & -- & 42.00 \tabgain{0.80} \\
\bottomrule
\end{tabular}
\end{minipage}
\end{table}

\paragraph{More text reasoning adds little.}
Adding text reasoning with a 4k token budget to direct answers helps L1 and L2 but hurts L3 (\Cref{tab:base-reasoning}). Many L3 outputs end without a final answer and are scored as wrong. Quadrupling the budget to 16k changes the three averages little, and L3 remains 9.59 points below direct answers. \Cref{tab:results-perception,tab:results-spatial} report all three settings and measure gains against the baseline.

\paragraph{Forced image generation does not help.}
We next increase visual reasoning effort while keeping the text-token budget at 16k. In \Cref{tab:base-forcing}, the model must generate an image and then continue in text. This forced interleaved reasoning ties text reasoning on V$^\star$ and loses on MMSI. Visual reasoning effort at inference time does not improve on text reasoning.

\paragraph{Even adding a ground-truth image does not help.}
The previous comparison does not separate image quality from image use. We replace the generated image with a verified ground-truth image on 500 spatial questions from \instruct, as if the model had produced it. Forcing an image costs 6.20 points on these questions, and the ground-truth image wins back only 0.80 points. These results suggest that \textbf{following the interleaved format is not enough}. The model must also learn to use what an intermediate image shows to advance its later text reasoning.

\subsection{Main Results}
\label{sec:experiments-main}

We compare its performance with the baseline and other models, examine whether its answers rely on the generated images, and measure the cost of interleaved reasoning at inference time.

\begin{table}[H]
\centering
\caption{\textbf{Level 1 benchmarks (Basic Visual Perception).} HR-4K / HR-8K are HRBench at 4K / 8K resolutions; MME-RW-L is the average of the Reasoning and Perception sub-scores. $\Delta$ is computed against BAGEL think-4k.}
\label{tab:results-perception}
\footnotesize
\setlength{\tabcolsep}{3pt}
\renewcommand{\arraystretch}{1.05}
\begin{tabularx}{\textwidth}{l*{6}{C}}
\toprule
& \multicolumn{5}{c}{Level 1: Basic Visual Perception} & Avg \\
\cmidrule(lr){2-6} \cmidrule(lr){7-7}
Model & V$^\star$ & HR-4K & HR-8K & MME-RW-L & RealWorldQA & L1 \\
\midrule
\multicolumn{7}{l}{\textit{Closed-source}} \\
GPT-5 & 69.63 & 67.50 & 65.50 & 56.80 & 73.86 & 66.66 \\
Claude Sonnet 4.6 & 66.49 & 72.50 & 64.50 & 46.95 & 71.50 & 64.39 \\
Gemini 2.5 Flash & 76.96 & 82.75 & 78.50 & 51.02 & 75.16 & 72.88 \\
\midrule
\multicolumn{7}{l}{\textit{Open-source}} \\
Qwen2.5-VL-7B & 71.20 & 61.38 & 58.75 & 43.10 & 61.96 & 59.28 \\
Qwen2.5-VL-32B & 75.92 & 66.62 & 64.38 & 45.28 & 63.01 & 63.04 \\
\midrule
\multicolumn{7}{l}{\textit{Unified}} \\
Janus-Pro & 38.22 & 44.88 & 39.63 & 23.19 & 41.83 & 37.55 \\
Chameleon & 28.27 & 26.25 & 24.88 & 24.13 & 36.21 & 27.95 \\
SenseNova-U1-8B & 61.26 & 56.38 & 52.25 & 39.52 & 69.67 & 55.82 \\
LongCat-Next-A3B & 76.44 & 65.38 & 62.00 & 40.78 & 71.76 & 63.27 \\
Show-o2-7B & 42.93 & 49.75 & 43.25 & 41.01 & 64.05 & 48.20 \\
\midrule
\multicolumn{7}{l}{\textit{BAGEL inference variants}} \\
BAGEL no-think & 52.88 & 46.88 & 39.50 & 41.88 & 70.59 & 50.35 \\
BAGEL think-4k & 58.64 & 59.38 & 48.88 & 25.64 & 64.44 & 51.40 \\
BAGEL think-16k & 61.26 & 63.00 & 50.50 & 25.64 & 68.89 & 53.86 \\
\midrule
\rowcolor{gray!15}
\textbf{\model (Ours)} & 68.06 & 67.00 & 55.38 & 45.28 & 70.46 & 61.24 \\
$\boldsymbol{\Delta}$ \textbf{baseline} & \textcolor{ForestGreen}{\textbf{+9.42}} & \textcolor{ForestGreen}{\textbf{+7.62}} & \textcolor{ForestGreen}{\textbf{+6.50}} & \textcolor{ForestGreen}{\textbf{+19.64}} & \textcolor{ForestGreen}{\textbf{+6.02}} & \textcolor{ForestGreen}{\textbf{+9.84}} \\
\bottomrule
\end{tabularx}
\end{table}
\begin{table}[t]
\centering
\caption{\textbf{Level 2 (Static Spatial Understanding) and Level 3 (Mental World Modeling) benchmarks.} CV-Bench is split into CV-2D and CV-3D. $\Delta$ is computed against BAGEL think-4k.}
\label{tab:results-spatial}
\vspace{4pt}
\footnotesize

\tablestyle{3pt}{1.0}
\resizebox{\textwidth}{!}{%
\begin{tabular}{@{}lcccccccccc@{}}
\toprule
& \multicolumn{4}{c}{Level 2: Static Spatial Understanding} & \multicolumn{4}{c}{Level 3: Mental World Modeling} & \multicolumn{2}{c}{Avg} \\
\cmidrule(lr){2-5} \cmidrule(lr){6-9} \cmidrule(lr){10-11}
Model & BLINK & CV-2D & CV-3D & SAT & MMSI & MindCube & OmniSpat & SpatViz & L2 & L3 \\
\midrule
\multicolumn{11}{l}{\textit{Closed-source}} \\
GPT-5 & 67.70 & 77.68 & 87.58 & 81.33 & 42.50 & 42.33 & 57.34 & 53.47 & 78.57 & 48.91 \\
Claude Sonnet 4.6 & 63.28 & 69.25 & 82.33 & 44.67 & 32.70 & 43.18 & 50.29 & 44.58 & 64.88 & 42.69 \\
Gemini 2.5 Flash & 67.23 & 74.21 & 80.75 & 65.33 & 35.90 & 42.20 & 53.62 & 29.66 & 71.88 & 40.35 \\
\midrule
\multicolumn{11}{l}{\textit{Open-source}} \\
Qwen2.5-VL-7B & 54.87 & 69.97 & 64.00 & 45.33 & 28.80 & 30.95 & 41.88 & 30.76 & 58.54 & 33.10 \\
Qwen2.5-VL-32B & 59.55 & 74.28 & 83.33 & 43.33 & 27.60 & 33.10 & 47.75 & 35.25 & 65.12 & 35.93 \\
\midrule
\multicolumn{11}{l}{\textit{Unified}} \\
Janus-Pro & 38.51 & 68.74 & 66.75 & 22.00 & 27.80 & 37.50 & 28.70 & 18.64 & 49.00 & 28.16 \\
Chameleon & 16.52 & 27.36 & 47.50 & 18.67 & 5.10 & 28.85 & 22.90 & 21.53 & 27.51 & 19.60 \\
SenseNova-U1-8B & 68.60 & 81.45 & 90.33 & 72.00 & 38.70 & 64.13 & 45.66 & 31.27 & 78.10 & 44.94 \\
LongCat-Next-A3B & 56.34 & 73.17 & 80.75 & 27.33 & 28.80 & 39.90 & 40.18 & 28.56 & 59.40 & 34.36 \\
Show-o2-7B & 43.35 & 67.95 & 74.08 & 26.00 & 29.00 & 40.00 & 40.51 & 30.68 & 52.84 & 35.05 \\
\midrule
\multicolumn{11}{l}{\textit{BAGEL inference variants}} \\
BAGEL no-think & 63.91 & 70.92 & 43.17 & 46.67 & 31.60 & 34.71 & 45.79 & 33.56 & 56.17 & 36.42 \\
BAGEL think-4k & 53.39 & 72.81 & 85.17 & 47.33 & 27.40 & 29.66 & 31.57 & 12.97 & 64.67 & 25.40 \\
BAGEL think-16k & 52.97 & 74.30 & 83.00 & 46.67 & 29.00 & 29.66 & 35.09 & 13.56 & 64.23 & 26.83 \\
\midrule
\rowcolor{gray!15}
\textbf{\model (Ours)} & 58.02 & 74.79 & 86.75 & 56.00 & 34.40 & 60.38 & 45.27 & 33.22 & 68.89 & 43.32 \\
$\boldsymbol{\Delta}$ \textbf{baseline} & \textcolor{ForestGreen}{\textbf{+4.63}} & \textcolor{ForestGreen}{\textbf{+1.98}} & \textcolor{ForestGreen}{\textbf{+1.58}} & \textcolor{ForestGreen}{\textbf{+8.67}} & \textcolor{ForestGreen}{\textbf{+7.00}} & \textcolor{ForestGreen}{\textbf{+30.72}} & \textcolor{ForestGreen}{\textbf{+13.70}} & \textcolor{ForestGreen}{\textbf{+20.25}} & \textcolor{ForestGreen}{\textbf{+4.22}} & \textcolor{ForestGreen}{\textbf{+17.92}} \\
\bottomrule
\end{tabular}}
\end{table}

\paragraph{Gains are consistent across all twelve benchmarks.}
\Cref{tab:results-perception,tab:results-spatial} show that \model improves on the baseline on every benchmark. The improvement spans visual perception and spatial reasoning. L3 has the largest average gain at 17.92 points, compared with 9.84 on L1 and 4.22 on L2. The larger L3 gain may reflect a division of labor between generation and perception. A unified model can construct the unseen state, then inspect it visually, turning part of mental simulation into a perceptual judgment. The gains extend beyond the training task families. The three in-domain benchmarks gain 15.71 points on average. All \textbf{nine out-of-domain benchmarks} also improve, by 9.87 points on average. At the benchmark level, MindCube gains the most at 30.72 points. SpatialViz has the largest gain among out-of-domain benchmarks at 20.25 points.
These gains combine pretext reasoning and instruction tuning.

\begin{figure}[!ht]
  \centering
  \includegraphics[width=0.95\linewidth]{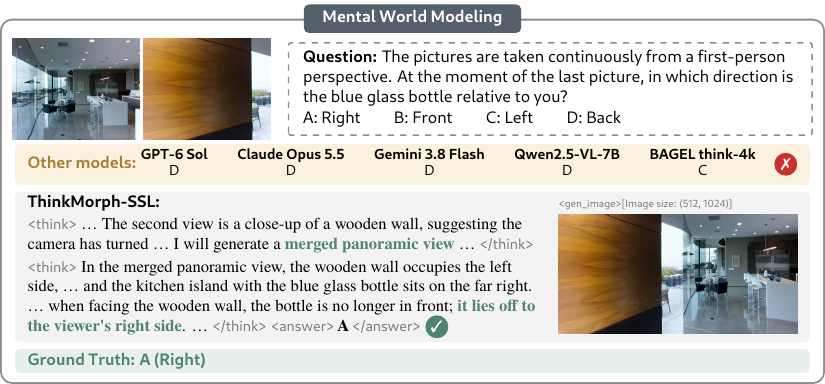}
  \caption{\textbf{A qualitative example from MMSI.}}
  \label{fig:case-study}
\end{figure}

\paragraph{\model rivals larger models on spatial reasoning.}
\Cref{fig:case-study} presents a qualitative example from MMSI. The final frame faces a wooden wall, while the bottle appears in the earlier view. A unified backbone, an open-source VLM, and three closed-source models answer left or back. Our model instead merges the views into a panorama and answers right. The panorama places the wall and bottle in one view, making their spatial relation available to the next reasoning step. Interleaved reasoning here combines both abilities: producing and reading an intermediate image, then using its content to derive the answer.

The advantage extends beyond this example to the spatial benchmark averages in \Cref{tab:results-spatial}. Among unified models, our model exceeds four competitors on both spatial levels. The lead also holds against open-source VLMs, including both the 7B and 32B versions of Qwen2.5-VL. The 32B model trails by 7.39 points on mental world modeling. The comparison with closed-source models is mixed. Our model outperforms Claude Sonnet~4.6 on both spatial levels. Its lead on static spatial understanding is 4.01 points. It also exceeds Gemini~2.5 Flash by 2.97 points on mental world modeling. On the in-domain MindCube benchmark, our model exceeds GPT-5 by 18.05 points.

\begin{figure}[ht]
\centering
\captionsetup{position=bottom,belowskip=0pt}
\begin{minipage}[t]{0.49\linewidth}
\centering
\parbox[c][0.65\linewidth][c]{\linewidth}{\centering\includegraphics[width=\linewidth]{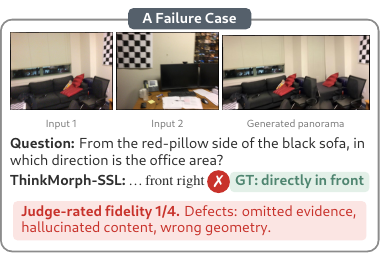}}
\caption{\textbf{A low-fidelity panorama leads to an error.} Key objects are missing or duplicated.}
\label{fig:case-failure}
\end{minipage}\hfill
\begin{minipage}[t]{0.49\linewidth}
\centering
\parbox[c][0.65\linewidth][c]{\linewidth}{\centering\includegraphics[width=\linewidth]{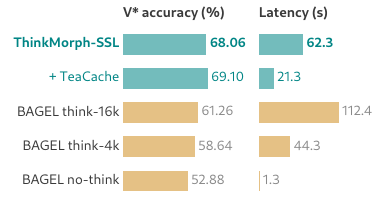}}
\caption{\textbf{Accuracy and latency on V$^\star$.} TeaCache reduces latency without lowering accuracy.}
\label{fig:cost}
\end{minipage}
\end{figure}

\begin{wraptable}{r}{0.43\linewidth}
\vspace{\dimexpr-0.5\intextsep-10pt\relax}
\centering
\captionsetup{position=top,aboveskip=0pt,belowskip=5pt}
\caption{\textbf{Does the image matter?} Spatial (train): 150 spatial problems from \instruct. MMSI (test): 150 test problems. Eight traces per problem. Left: within-problem accuracy w/ vs.\ w/o images. Right: error rates w/ vs.\ w/o critical defects.}
\label{tab:image-use}
\footnotesize
\setlength{\tabcolsep}{1.5pt}
\renewcommand{\arraystretch}{1.15}
\begin{tabular*}{\linewidth}{@{\extracolsep{\fill}}lccccc@{}}
\toprule
 & \multicolumn{3}{c}{Accuracy (\%)} & \multicolumn{2}{c}{Error rate (\%)} \\
\cmidrule(lr){2-4}\cmidrule(l){5-6}
Split & w/ img & w/o img & $\Delta$ & defect & none \\
\midrule
Spatial (train) & 69.4 & 32.2 & +37.2 & 40.7 & 15.4 \\
MMSI (test) & 31.3 & 18.5 & +12.8 & 67.8 & 47.4 \\
\bottomrule
\end{tabular*}
\par\vspace{-8pt}
\end{wraptable}

\paragraph{The model leverages its own generated images for reasoning.}
We analyze sampled traces from both splits in \Cref{tab:image-use}, comparing image use, image defects, and answer accuracy. We first check whether traces with images are more accurate. The model chooses whether to generate an image. Higher accuracy alone cannot show that the image caused the gain. For questions with both image and image-free traces, we compare accuracy between the two groups. Accuracy is 37.2 points higher with images on the training subset and 12.8 on MMSI. We then check whether image defects accompany answer errors. A judge flags critical defects without seeing the final answer. Traces with defective images have higher error rates on both splits. On MMSI, the gap is 20.4 points.

The failure in \Cref{fig:case-failure} further illustrates this link between image defects and answer errors. The panorama omits the desk and monitors from Input~2. It also duplicates the sofa, red pillow, and chair. The model answers front right instead of directly in front. The model appears to use the generated scene as evidence, making its answer vulnerable to missing or distorted spatial information.

\paragraph{The cost of interleaved reasoning can be reduced without losing accuracy.}
\label{sec:experiments-cost}
More inference-time effort alone did not yield effective interleaved reasoning in \Cref{tab:base-reasoning,tab:base-forcing}. After training, we apply inference-time acceleration to reduce the model's generation cost. \Cref{fig:cost} compares accuracy and end-to-end latency on V$^\star$. The uncached model is more accurate but slower than the baseline. Image generation accounts for nearly all its FLOPs, making it the main optimization target. We apply TeaCache~\cite{liu2024timestep} to reuse computation across image-denoising steps. This reduces total FLOPs by \textbf{77\%} without lowering accuracy. Latency drops by 41.0 seconds to 21.3 seconds, below the baseline's 44.3 seconds. \model can incorporate compatible accelerators for both text and image generation, with TeaCache serving as one initial example.

\subsection{Does Pretext Reasoning Improve Interleaved Reasoning?}
\label{sec:experiments-ablation}
\label{sec:experiments-analysis}

The main results combine pretext reasoning and instruction tuning. Here, we isolate the contribution of pretext reasoning, then examine the training schedule and the use of instructions in Stage~1.

\begin{figure}[ht]
  \centering
  \includegraphics[width=\linewidth]{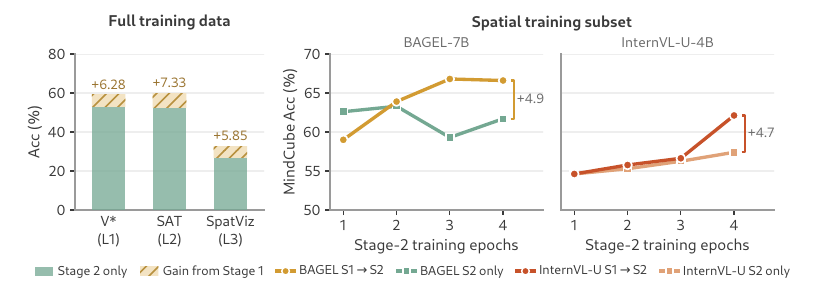}
  \caption{\textbf{Ablations of pretext reasoning.} \textbf{Left:} Stage-2-only accuracy (solid) and the gain from Stage~1 (hatched). Middle and right: MindCube accuracy with and without Stage~1 on BAGEL-7B and InternVL-U-4B.}
  \label{fig:ablation-staging}
\end{figure}

\paragraph{Pretext reasoning helps at the same instruction-tuning budget.}
\Cref{fig:ablation-staging} compares the two-stage training recipe of \model with Stage~2 alone in two settings.
\textbf{(1)~Full training data.} We evaluate one representative benchmark per level. The staged run starts from the same Stage-1 checkpoint used for the main results in \Cref{tab:results-perception,tab:results-spatial}. Both runs use \instruct with matched Stage-2 training budgets. Pretext reasoning helps across all three task levels, with an average gain of \textbf{6.49 points}.
\textbf{(2)~Across backbones.} We compare BAGEL-7B and InternVL-U-4B using identical data and training settings. Stage~2 uses the spatial training subset of \instruct, with the Stage-1 pretext-training budget unchanged. On MindCube, the staged BAGEL run leads from the second epoch and finishes 4.9 points ahead. On InternVL-U, the gain stays below 0.5 points for the first three epochs, then reaches 4.72 points in the fourth. These results suggest that pretext reasoning helps across backbones, with its benefit emerging as instruction tuning progresses.

\begin{wrapfigure}{r}{0.38\linewidth}
\vspace{-\intextsep}
\centering
\includegraphics[width=\linewidth]{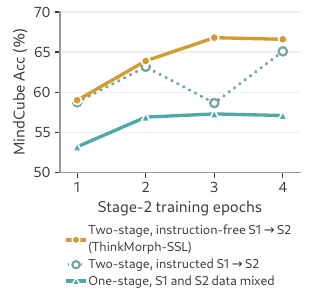}
\caption{\textbf{Effects of training order and pretext instructions.} MindCube accuracy under staged versus mixed training and instruction-free versus instructed pretext training, with matched instruction-tuning budgets.}
\label{fig:instruction-free}
\vspace{-\intextsep}
\end{wrapfigure}

\paragraph{Pretext reasoning should come before instruction tuning.}
Following the spatial-subset setting above, \Cref{fig:instruction-free} compares staged training with mixing both corpora at the same instruction-tuning budget. The mixed run stays near 57\%, while staged training reaches 66.6\%. Pretext reasoning therefore provides better preparation when learned before instruction tuning. These results suggest complementary roles for the two stages. Stage~1 practices producing and reading intermediate images. Stage~2 then teaches the model to use their content to advance text reasoning.

\paragraph{Instruction-free traces transfer better than instructed ones.}
Stage-1 inputs provide neither a question nor an instruction. This makes the task harder: the model must infer the pretext task from the image, without a text instruction specifying the transformation. Following the same spatial-subset setting, we add instructions only to Stage-1 inputs and keep Stage~2 unchanged. \Cref{fig:instruction-free} compares the two variants. Surprisingly, the instruction-free run leads at every checkpoint, with a modest mean advantage of 2.7 points. This suggests that Stage~1 can build the ability to produce and read intermediate images without explicit task instructions. This instruction-free preparation transfers slightly better to Stage~2, where the model learns to use image content to advance its text reasoning.

\section{Related Work}
\label{sec:related}

Our work connects three lines of research. First, recent unified multimodal models support both visual understanding and image generation within a shared interface, using mixed-modal autoregression, diffusion- or continuous-token generation, decoupled visual pathways, or modality-bridging mechanisms \citep{chameleon, Showo, Showo2, VILAU, Emu3, UniFluid, BLIP3o, Janus, JanusFlow, januspro, LatentUM, MetaQueries, MetaMorph, BeyondLanguageModeling, bagel}. These works focus primarily on building the unified interface itself; in contrast, we study how to convert a pretrained unified model into an interleaved reasoner.

Second, self-supervised vision pretraining has long used image-derived supervision, including rotation prediction, jigsaw puzzles, colorization, inpainting, masked image modeling, self-distillation, and image--text alignment \citep{gidaris2018rotation, noroozi2016jigsaw, zhang2016colorful, pathak2016context, he2021masked, bao2022beit, SimMIM, iBOT, caron2021emerging, DINOv2, radford2021learning, SigLIP}. These methods typically learn representations, tokenizers, or alignment spaces. Pretext reasoning instead recasts these tasks as autoregressive Text--Image--Text traces, so the model learns to produce and consume visual states within its own reasoning trajectory.

Third, interleaved multimodal reasoning has explored visual grounding for chain-of-thought, external sketchpads and tools, latent visual thoughts, supervised text-image reasoning traces, and benchmarks that expose failures in visual simulation and spatial reasoning \citep{VisualCoTPrompting, shao2024visualcot, GroundedCoT, ViperGPT, MMREACT, VisualSketchpad, Latte, OpenThinkImg, AdaReasoner, PixelReasoner, DeepEyes, mintcot, bigverdi2024perceptiontokensenhancevisual, Mirage, LatentSketchpad, MullTokens, STARE, ViCBench, CausalSpatial, li2025tirbench}. Zebra-CoT and ThinkMorph develop supervised interleaved traces that train models to generate and reason from intermediate images~\citep{zebracot,thinkmorph}. Most of this work depends on task-specific supervised traces or predefined tools. We add a stage before such supervision, in which traces built from unlabeled images teach the model to construct visual states and continue from them. Instruction tuning then adapts these abilities to question-driven reasoning. We provide a fuller discussion in Appendix~\ref{app:related}.

\section{Discussion and Conclusion}
\label{sec:discussion}

We introduced \model, a two-stage recipe for scaling interleaved multimodal reasoning through self-supervised visual practice. \pretext provides instruction-free traces for producing and interpreting intermediate images, while \instruct supplies automatically constructed demonstrations for question-driven instruction tuning. Pretext training improves subsequent learning at a fixed instruction-tuning budget, and the full recipe improves performance across perception, spatial understanding, and mental world modeling. These findings show that supervision need not be designed around downstream questions to benefit interleaved reasoning. Images themselves can provide scalable training targets that make learning from task-specific demonstrations more effective.

\paragraph{Limitations.}
Our pretext pipeline constructs supervision automatically, but we have not systematically measured how performance changes with more pretext data, training compute, or larger models. We study seven visual pretext tasks and evaluate mainly perception and spatial reasoning; whether the recipe extends to other tasks and domains remains open. Interleaved reasoning also has substantial inference costs, which may limit its use when compute or latency is constrained. Finally, the base model's pretraining data are not public, so overlap with our source data cannot be ruled out.

\paragraph{Broader impact.}
We openly release the datasets, model checkpoints, and code to support research on interleaved multimodal reasoning. Better spatial reasoning could help assistive systems describe scenes and explain layouts. The model may also reproduce biases in its pretrained backbone and the public datasets used here; we have not audited those biases.

\begin{ack}
This project was partially funded by Samsung and by SRC CoCoSys.
\end{ack}

\bibliographystyle{plain}
\bibliography{main}

\appendix
\raggedbottom
\clearpage

\providecommand{\titpill}[2][gray]{%
  \tcbox[
    on line, boxrule=0pt, boxsep=0pt,
    left=4pt, right=4pt, top=1pt, bottom=1pt,
    arc=2pt,
    colback=#1!10, colframe=#1!10,
  ]{\scriptsize\textcolor{#1!55!black}{\strut\textbf{#2}}}%
}

\newtcolorbox{titcard}[1][]{%
  enhanced,
  colback=white, colframe=black!18,
  boxrule=0.4pt, arc=2.5mm,
  left=12pt, right=12pt, top=10pt, bottom=10pt,
  before skip=10pt, after skip=10pt,
  #1
}

\newtcolorbox{titcardwide}[1][]{%
  enhanced,
  colback=white, colframe=black!18,
  boxrule=0.4pt, arc=2.5mm,
  left=12pt, right=12pt, top=10pt, bottom=10pt,
  before skip=10pt, after skip=10pt,
  #1
}

\newcommand{\titimg}[3]{%
  \begin{minipage}[t]{#1}%
    \centering
    \includegraphics[width=\linewidth,height=2.6cm,keepaspectratio]{#2}%
    \par\smallskip
    {\scriptsize\textsc{\textcolor{black!50}{#3}}}%
  \end{minipage}%
}

\newcommand{\titimgbig}[3]{%
  \begin{minipage}[t]{#1}%
    \centering
    \includegraphics[width=\linewidth,height=3.6cm,keepaspectratio]{#2}%
    \par\smallskip
    {\scriptsize\textsc{\textcolor{black!50}{#3}}}%
  \end{minipage}%
}

\newcommand{\titcardheader}[3]{%
  \noindent
  {\textbf{#1}}\,\textcolor{black!35}{$\bullet$}\,\textit{\small Example~#2}\hfill #3%
  \par\smallskip
  {\color{black!12}\rule{\linewidth}{0.4pt}}%
  \par\medskip
}

\newcommand{\titthought}[2]{%
  \par\smallskip
  {\scriptsize\textsc{\textcolor{black!50}{#1}}}\par
  \vspace{1pt}%
  {\footnotesize\raggedright\setlength{\baselineskip}{1.20\baselineskip}#2\par}%
}

\newcommand{\titqa}[2]{%
  \par\medskip
  {\color{black!12}\rule{\linewidth}{0.4pt}}%
  \par\smallskip
  {\footnotesize\textbf{\textcolor{black!60}{Q.}}\,#1\par
   \smallskip
   \textbf{\textcolor{black!60}{A.}}\,#2\par}%
}

\newcommand{\tittasksection}[2]{%
  \par\bigskip
  \noindent{\large\textbf{#1}}\hfill{\small\color{black!50}#2}%
  \par\smallskip
  {\color{black!18}\rule{\linewidth}{0.6pt}}%
  \par\smallskip
}

\section{Overview of the Appendix}

This appendix is organized as follows:
\begin{itemize}
  \item \Cref{app:related} extends the related-work discussion.
  \item \Cref{app:appendix-pretext-composition} reports the composition of \pretext{} and \instruct{}.
  \item \Cref{app:appendix-instruct} details the task-specific \instruct{} pipelines and the shared trace synthesis prompt.
  \item \Cref{app:appendix-instruct-examples} shows example traces per \instruct{} task, and \Cref{app:appendix-pretext-examples} shows one Text--Image--Text trace per pretext task.
  \item \Cref{app:appendix-training} reports training and evaluation details: hyperparameters of the main and ablation runs, the loss, system prompts, compute, the VLMEvalKit configuration, per-benchmark scoring, and judge prompts.
  \item \Cref{app:appendix-analyses} presents qualitative good cases and failure modes of \model, data overlap, and alternative routes without visual-state training.
\end{itemize}

\section{Extended Related Work}\label{app:related}
\paragraph{Unified multimodal models.}
Recent unified multimodal models support visual understanding and image generation within a shared interface. Existing approaches vary in how they connect the two capabilities, from early-fusion mixed-modal modeling \citep{chameleon} to autoregressive token-based unification \citep{Showo, Showo2, VILAU, Emu3} and continuous-token or diffusion-based variants \citep{UniFluid, BLIP3o}; another line decouples or harmonizes visual pathways \citep{Janus, JanusFlow, januspro} or introduces latent or query-based bridges between modalities \citep{LatentUM, MetaQueries}, and a complementary line studies adaptation- or pretraining-based recipes \citep{MetaMorph, BeyondLanguageModeling}. We build on BAGEL, which shows that large-scale interleaved pretraining can yield emergent unified capabilities \citep{bagel}. These works primarily address how to build or adapt the unified interface itself. In contrast, \model studies the orthogonal training bridge from generic unified competence to interleaved reasoning behavior: pretext reasoning is inserted between unified pretraining and question-driven instruction tuning to teach the model the mechanics of Text--Image--Text continuation before supervised reasoning.

\paragraph{Self-supervised visual pretraining.}
A long line of work learns visual representations without human labels by constructing supervision from the image itself. Early pretext tasks include instance and rotation discrimination \citep{ExemplarCNN, gidaris2018rotation}, spatial-context and jigsaw puzzles \citep{doersch2015context, noroozi2016jigsaw}, color and inpainting reconstructions \citep{zhang2016colorful, pathak2016context}, and motion-based signals from video \citep{EgoMotion, Tracking}. These ideas later scaled through masked image modeling \citep{he2021masked, bao2022beit, SimMIM, iBOT}, self-distillation \citep{caron2021emerging, DINOv2}, and image--text alignment \citep{radford2021learning, SigLIP}. Despite their diversity, these objectives primarily produce an encoder, a tokenizer, or an alignment space; the model is not trained to use visual generation as part of its own computation. Pretext reasoning revisits this paradigm but changes the interface: each restoration task is reformulated as an autoregressive Text--Image--Text trace. The resulting supervision trains a unified model to continue interleaved perception and generation trajectories rather than merely produce a visual representation.

\paragraph{Interleaved multimodal reasoning.}
Cognitive science has long argued that human spatial reasoning often relies on mental imagery and visual simulation rather than purely verbal deduction \citep{shepard1971mental, kosslyn1980image}. This motivates models that can reason through intermediate visual states instead of expressing every step only in text. Prior work brings visual content into reasoning in three main ways. The first augments textual chain-of-thought with visual grounding \citep{VisualCoTPrompting, shao2024visualcot, GroundedCoT} or with external programmatic\cite{duan2025codeplotcot}, agent-style, and sketchpad tools \citep{ViperGPT, MMREACT, VisualSketchpad}; recent work further trains models to invoke vision specialists or pixel-space operations adaptively \citep{Latte, OpenThinkImg, AdaReasoner}, often through reinforcement learning over images \citep{PixelReasoner, DeepEyes, active-o3, o3}. These methods improve evidence gathering and interpretability, but the reasoning trace remains primarily textual or tool-mediated, and its visual operations are bounded by predefined tools or experts.

The second line incorporates visual state directly into the reasoning trajectory. Some methods train on supervised interleaved text-image reasoning traces \citep{mintcot, zebracot, thinkmorph}; others introduce latent visual tokens as task-specific intermediates, mental imagery, internal sketchpads, or modality-agnostic thought tokens \citep{bigverdi2024perceptiontokensenhancevisual, Mirage, LatentSketchpad, MullTokens}. Complementary work studies test-time scaling and representation bottlenecks for such multimodal reasoning processes \citep{UniT, LatentUM}. Recent benchmarks further motivate this direction, showing that current models struggle when answers depend on intermediate visual states, visual simulation, or causal-spatial structure rather than language-only reasoning \citep{zhou2025visualizingstepreasoningmira, STARE, ViCBench, CausalSpatial,li2025tirbench}.

The third line studies interleaved multimodal generation, where mixed image-text content is the final artifact rather than an intermediate reasoning process \citep{Mogao, LLMI, WanWeaver, ImageGenCoT}. This direction is complementary to our goal, since we use generated visual states as intermediate thoughts to improve multimodal understanding and reasoning. Across these lines, the dominant recipe relies on supervised reasoning traces, which are costly to scale and limited in coverage. We target the stage that prior work largely skips. Pretext reasoning provides instruction-free training, before any question-driven supervision, that teaches a unified model the mechanics of Text--Image--Text continuation, while \instruct then fits within the supervised interleaved-reasoning lineage as a question-driven corpus.

\section{Data Composition}
\label{app:appendix-pretext-composition}

\begin{figure}[t]
  \centering
  \includegraphics[width=0.9\textwidth]{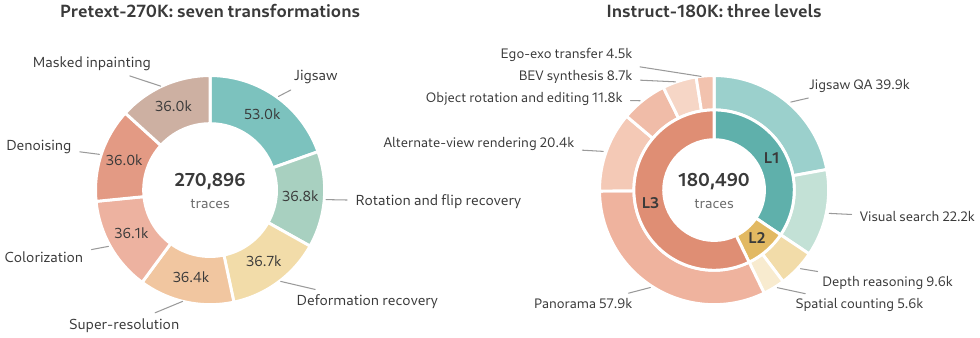}
  \caption{\textbf{Corpus composition.} Left: traces per pretext transformation in \pretext. Right: traces in \instruct per level (inner ring; L1 basic visual perception, L2 static spatial understanding, L3 mental world modeling) and per construction pipeline (outer ring; nine pipelines).}
  \label{fig:data-stats}
\end{figure}

\Cref{fig:data-stats} shows the composition of both corpora. \pretext{} draws its images from SA-1B, PixMo-Cap, VSI-590K, LLaVA-OneVision-1.5-Instruct-Data, and DeepVision-103K for all seven transformations, and additionally from Realsee3D, 360+x, and Matterport3D for Jigsaw.

\section{Implementation Details of the \instruct{} Curation Pipeline}
\label{app:appendix-instruct}

\Cref{tab:app-level-mapping} maps each level of \instruct{} to its task families and to
the kind of intermediate image its traces construct.

\begin{table}[ht]
\centering
\caption{\textbf{Role of the intermediate visual state at each level of \instruct.}}
\label{tab:app-level-mapping}
\small
\begin{tabular}{@{}>{\raggedright\arraybackslash}p{0.2\linewidth}>{\raggedright\arraybackslash}p{0.21\linewidth}>{\raggedright\arraybackslash}p{0.27\linewidth}>{\raggedright\arraybackslash}p{0.24\linewidth}@{}}
\toprule
Level & Role of the visual state & Task families & Intermediate image \\
\midrule
L1: Basic visual perception & Focus on visible evidence & Visual search, Jigsaw QA & Target boxed or zoomed in the scene; unscrambled image \\
L2: Static spatial understanding & Organize visible spatial relations & Spatial counting, depth reasoning & Instance crops; depth map \\
L3: Mental world modeling & Reconstruct unseen or transformed states & Bird's-eye view synthesis, panorama, alternate-view rendering, object rotation, ego--exo perspective transfer & Bird's-eye view; panorama; alternate view; rotated object; another agent's view \\
\bottomrule
\end{tabular}
\end{table}

Mental world modeling further splits into viewpoint synthesis, where the
camera moves while the scene stays fixed, and object manipulation, where
the camera stays while the object changes. Within viewpoint synthesis the
subsections are ordered by how far the synthesized view departs from the
input egocentric frame. Panorama stays at the same camera
position with a wider field of view. Bird's-eye view synthesis leaves the
agent for a world-anchored overhead. Ego--exo perspective transfer and
alternate-view rendering both adopt another entity's viewpoint; the first
learns from real paired ego/exo captures and the second from single-image
3D lifts, since no single source covers both.

\subsection{Interleaved chain-of-thought generation}
\label{app:cot-generation}

Each row arrives with the question stem, the gold label, input images, and
zero or more reasoning images that have already been rendered or
photographed. A separate vision-language pass writes the trace a student
model is trained to mimic. For $N$ attached reasoning images the writer
returns $N+1$ text blocks alternating with those images. The first block
is written before any generated view is consulted: it inspects the inputs,
states what visualization will come next, and explains why. Each later
block is written after its corresponding image: it describes what that
view makes explicit, ties it back to the inputs, and either plans the
next view or concludes.

The writer is invoked through two cascaded prompts. An optional first
pass paraphrases the question and answer to diversify surface form
without changing semantics. The main pass casts the model as an agent
that generates the intermediate views during reasoning, under three
constraints: it must not refer to the views as externally supplied, it
must not leak the gold answer before the final block, and it must use
descriptive view names (bird's-eye view, depth map) rather than ordinal
labels. The model sees the stem, the answer, a per-task suggested
reasoning approach, and the input and reasoning images in alternation.
When per-row metadata indicates which reasoning view corresponds to which
input, that correspondence is attached as a hidden hint usable only for
internal consistency. Outputs that omit any of the $N+1$ expected blocks
are rejected and re-run.

Tasks that do not commit to a visual reasoning sequence skip reasoning
images. The model returns a single continuous-prose chain under a lighter
system prompt that forbids bullet inventories and mechanical enumeration.

Per-task templates supply both the suggested reasoning approach and the
short captions used to label the attached reasoning images. The fixed
system roles below are shared across tasks. The template excerpts that
follow are representative rather than exhaustive.

\begin{tcolorbox}[breakable, colback=gray!4, colframe=gray!50,
                  title={\texttt{REWRITE\_SYSTEM\_PROMPT} (optional question--answer paraphrase).},
                  label=box:rewrite-system]
\small
\begin{verbatim}
Rewrite the given question and answer with diverse wording and sentence
structure while preserving correctness.

Rules:
1. Preserve the original meaning -- do not alter spatial relations,
   change counts, swap entities, or introduce ambiguity. All entities,
   constraints, and spatial details must be retained.
2. Vary vocabulary, sentence structure, and phrasing style freely. Do
   not add hints toward the answer.
3. For multiple-choice questions, keep the option order unchanged; you
   may rephrase option text if the meaning is preserved.
4. Output a JSON object with keys: "question", "answer".
\end{verbatim}
\end{tcolorbox}

\begin{tcolorbox}[breakable, colback=gray!4, colframe=gray!50,
                  title={\texttt{CORE\_SYSTEM\_PROMPT} (interleaved reasoning images).},
                  label=box:core_system]
\small
\begin{verbatim}
You are an advanced spatial reasoning AI with a unique capability: you
can generate intermediate visualizations (e.g., bird's-eye views, side
profiles, depth maps, rotated perspectives, or step-by-step
constructions) to help understand spatial relationships.

Your Task:
You will be given:

1. Input Images: One or more images related to the task (e.g.,
   multiple viewpoints of a scene, a puzzle layout, a diagram, etc.).
2. Intermediate Generated Views: A sequence of visual aids you
   "generate" during reasoning (e.g., new perspectives, step-by-step
   assembly stages, annotated overlays, etc.).
3. A spatial reasoning question and its correct answer.

Your Goal:
Generate textual reasoning that connects these images step-by-step.
Since you have the capability to "generate" these intermediate views,
you must explain why you need them and what they show.

For N intermediate views, you need N+1 reasoning blocks:

* reasoning_thought_0: BEFORE generating any view
  - Analyze the initial Input Images and questions.
  - Explicitly state what type of view you will generate next and why.

* reasoning_thought_1: AFTER generating the first view
  - Now you can see the generated image.
  - Describe the type of view you generated (e.g., "This top-down
    perspective reveals...") and analyze its content.
  - Synthesize this with the original inputs.
  - State what you will generate next (if needed).

* reasoning_thought_2: AFTER generating the second view
  - Describe the new view and what information it provides.
  - Synthesize all observations to derive the final answer.

* (Continue this pattern for more views if needed.)

Output Format: Return your reasoning as a JSON object with the
required keys.

Important Notes:

1. Never describe a view before generating it. In reasoning_thought_0
   you only state your intention to generate a specific type of view.
2. Dynamic Terminology: Do not use generic labels like "Image 1".
   Instead, use descriptive terms based on the view type, such as
   "the bird's-eye view", "the depth map", "the side-angle
   perspective", or "the zoomed-in visualization".
3. Cross-Referencing: Intermediate views help with structure but may
   lack specific details. If you need, please refer back to the Input
   Images for fine-grained object identification.
4. Step-by-Step Deduction: Your conclusion must be the logical result
   of combining the input images with your generated visualizations.
5. Do not state, hint at, or assume the correct answer in the
   reasoning process. Do not refer to intermediate views as "provided"
   or "given"; describe them as views you generate during reasoning.
\end{verbatim}
\end{tcolorbox}

\begin{tcolorbox}[breakable, colback=gray!4, colframe=gray!50,
                  title={\texttt{TEXT\_ONLY\_SYSTEM\_PROMPT} (no intermediate images).},
                  label=box:text_only_system]
\small
\begin{verbatim}
You are an advanced spatial reasoning AI.

Your Task:
You will be given:

1. Input Images: One or more images related to the task.
2. A spatial reasoning question and its correct answer.

Your Goal:
Generate a natural chain-of-thought reasoning that explains how to
arrive at the correct answer by analyzing the input images.

Output Format: Return your reasoning as a JSON object with exactly one
key: "reasoning".

Important Notes:

1. Think naturally. Write your reasoning as a continuous, flowing
   thought process -- as if you are talking through the problem.
   Avoid rigid numbered lists, bullet-by-bullet image inventories, or
   mechanical enumeration patterns like "Count = 1, Count = 2".
2. Build toward the answer progressively. Do not state or assume the
   correct answer at the beginning. Let your conclusion emerge from
   your observations and spatial analysis.
\end{verbatim}
\end{tcolorbox}

\begin{tcolorbox}[breakable, colback=gray!4, colframe=gray!50,
                  title={Counting with mask overlay (\texttt{lvis\_mask\_counting}).},
                  label=box:lvis_mask_counting_prompt]
\small
\begin{verbatim}
Task: Count how many instances of the queried object category are in
the image.
reasoning_thought_0 (BEFORE generating any view): Observe the input
image, identify the queried object category, and note any challenges
for counting (occlusion, clutter, similar-looking objects). Explain
why you will generate a mask overlay that isolates only the target
instances on a white background.
reasoning_thought_1 (AFTER generating the mask overlay): The generated
mask overlay shows only the target instances on a white background
with black contours around each instance. Count each distinct instance
visible in the overlay carefully and state the final count.
\end{verbatim}
\end{tcolorbox}

\begin{tcolorbox}[breakable, colback=gray!4, colframe=gray!50,
                  title={Top-down counting on multi-view scenes (\texttt{topdown\_counting}).},
                  label=box:topdown_counting_prompt]
\small
\begin{verbatim}
Task: Count how many instances of the queried object category are in
the full scene.
reasoning_thought_0: Briefly inspect the multi-view input images, and
explain why you will generate a top-down view.
reasoning_thought_1: Describe what the generated top-down view
reveals, count the queried category carefully in that view, and
finish with the final count consistent with the question.
\end{verbatim}
\end{tcolorbox}

\begin{tcolorbox}[breakable, colback=gray!4, colframe=gray!50,
                  title={RefSpatial depth with two generated views (\texttt{refspatial\_point\_depth}).},
                  label=box:refspatial_prompt]
\small
\begin{verbatim}
Task: Answer a RefSpatial depth question using the available generated
depth-support view or views. When there are two generated views, use
the first to localize the correct point or points and the second to
make the final depth judgment.
For reasoning_thought_0, briefly inspect the input image and
question, explain what depth relation must be resolved, and state
what kind of depth-support view you will form next. Do not answer yet.
\end{verbatim}
\end{tcolorbox}

\begin{tcolorbox}[breakable, colback=gray!4, colframe=gray!50,
                  title={Ego--exo perspective transfer (\texttt{egoexo\_perspective\_taking}).},
                  label=box:egoexo_prompt]
\small
\begin{verbatim}
Task: Answer a perspective-taking spatial question from a third-person
scene image.
The input image shows the scene from an outside view, while the
generated view represents the queried person's first-person
perspective.
Do not mention raw field names such as `images`, `reasoning_image_0`,
or any hidden metadata.
Do not mention markers, highlights, or annotations that are not
visible to the final solver.
Instead, refer naturally to the outside scene view and the generated
first-person view.
reasoning_thought_0 (BEFORE generating any view): Inspect the
third-person image, identify which person the question refers to,
locate the relevant object or objects mentioned in the question, and
explain why the answer cannot be judged directly from the outside
camera.
State that you will generate a first-person view from that person's
position and facing direction so the relation can be judged from
their own perspective.
Do not reveal the final answer yet.
reasoning_thought_1 (AFTER generating the first view): Analyze the
generated first-person view from that person's perspective.
Use it to determine the asked relation precisely, such as left versus
right, above versus below, or front versus behind.
Keep the reasoning grounded in visible scene evidence and perspective
transformation only, then finish with the correct answer.
\end{verbatim}
\end{tcolorbox}

\begin{tcolorbox}[breakable, colback=gray!4, colframe=gray!50,
                  title={Panorama merge (\texttt{panorama\_spatial}).},
                  label=box:panprama_prompt]
\small
\begin{verbatim}
Task: Answer a spatial reasoning question about an indoor or outdoor
scene captured from two partially overlapping viewpoints (left view
and right view from the same camera position).
The merged panoramic view provides a wider field of view that reveals
the full spatial layout.
reasoning_thought_0 (BEFORE generating any view): Inspect the two
input images carefully.
Identify the key objects, scene areas, and spatial cues mentioned in
the question.
Note how the two views overlap at the center (shared region) and what
each view uniquely shows on its periphery.
Explain that you will generate a merged panoramic view to see the
complete spatial layout and resolve the relationship asked. Do not
answer yet.
reasoning_thought_1 (AFTER generating the merged view): Analyze the
generated merged panoramic view.
Use it to determine the full spatial layout -- identify where objects
and scene areas are located relative to each other and relative to
the camera.
Cross-reference with the original input images for fine-grained
object identification.
Derive the final answer through step-by-step spatial reasoning
grounded in the visual evidence.
\end{verbatim}
\end{tcolorbox}

\begin{tcolorbox}[breakable, colback=gray!4, colframe=gray!50,
                  title={Mindcube-style left/right with two top-downs (\texttt{template\_02\_directional\_query}).},
                  label=box:mindcube_prompt]
\small
\begin{verbatim}
Task: Answer a left/right query for the target object from one
queried input viewpoint.
Do not mention raw field names such as `reasoning_image_0`,
`reasoning_image_1`, `prompt_metadata`, or
`reasoning_image_1_matches_image_index` in the output reasoning.
Do not mention any hidden hint, metadata, annotation, field name, or
externally provided correspondence in the output reasoning.
Instead, refer to the visual evidence naturally as the first
generated top-down view, the rotated top-down view, or the
corresponding original image.
reasoning_thought_0 (BEFORE generating any view): Inspect the queried
input image and the other input images briefly. State that you will
first form a bird's-eye view of the queried viewpoint, which helps
convert the egocentric left/right question into a rotation on the
top-down layout. Do not answer yet.
reasoning_thought_1 (AFTER generating the first view): Analyze that
first bird's-eye view, then apply the 90-degree rotation rule
(clockwise for left questions, anticlockwise for right questions in
this dataset) to explain why you need the second generated view.
Keep this step focused on forming the rotated view itself, without
yet naming which original image it matches.
reasoning_thought_2 (AFTER generating the second view): State which
original input image has the same orientation as the rotated view,
then use that matched original input image to identify what lies in
front in the aligned view, because that matched original image is
the normal-perspective version of the rotated view.
So the actual object judgment should be made from the matched
original input image, then translated back into the answer for the
queried viewpoint.
Use the original images to confirm object identity and appearance
details, and keep the explanation grounded in visual reasoning only.
\end{verbatim}
\end{tcolorbox}

Downstream, modality-specific post-filters (for example the panorama CoT
consistency check in \Cref{app:panorama-post}) assume these fields are
already present on each row.

\subsection{Basic visual perception}

\subsubsection{Visual search}
\label{sec:appendix-instruct-visual-search}

Visual search targets a single object inside an object-dense scene. Images
must contain at least 135 annotated bounding boxes to ensure sufficient object
density for unambiguous target selection. For each candidate target, a
templated GPT call conditioned on the full image with a red bounding box drawn
around the target or a high-resolution crop of that object generates a
four-option multiple-choice question spanning fine-grained attribute
recognition, spatial relationships, OCR, and visual reasoning. The correct
answer must be verifiable from the cropped view alone, so the intermediate
image---the full scene with the red bounding box rendered---is by construction
the visual artifact required to answer. A four-model judge ensemble
(GPT-4.1, Gemini-2.5~Pro, GPT-5.1, and Gemini-3-Pro-Preview) then rejects
items on any of four failure criteria: visual mismatch or ambiguity
(\texttt{VIS\_FAIL}), poor question logic (\texttt{LOGIC\_FAIL}), incorrect
ground-truth answer (\texttt{ANS\_ERR}), or absence of visual evidence in the
crop (\texttt{NO\_EVID}).

\subsection{Static spatial understanding}

\subsubsection{Spatial counting}
\label{sec:appendix-instruct-object-counting}

Object-counting supervision is sampled from LVIS~\cite{gupta2019lvis}, which provides
high-quality instance segmentation masks for a large open vocabulary. For
each image we enumerate instances whose mask area clears a minimum pixel
threshold, crop each instance with a small margin from its axis-aligned
box, and pair the crops with questions whose gold counts come from the
mask inventory alone, with no language-model labelling of the number. The
task tests whether the model can aggregate fragmented visual evidence
into a verifiable count.

\subsubsection{Depth reasoning}
\label{sec:appendix-instruct-depth-perception}

Depth-oriented questions are subsampled from the RefSpatial corpus
introduced in RoboRefer~\cite{zhou2025roborefer}, whose annotations include depth-related
probes. We keep the original question and answer semantics but replace
the depth or disparity visualization: scalar maps are re-coloured with
the viridis colormap for perceptual uniformity and clearer contrast in
printed traces, without changing the underlying geometry.

\subsection{Mental world modeling}

\subsubsection{Viewpoint synthesis}
\label{sec:appendix-instruct-panorama-completion}
\label{sec:appendix-instruct-viewpoint-synthesis}

\paragraph{Panorama.}
The smallest departure from the input is to stay at the same camera
position and widen the field of view. Supervision for this comes from
equirectangular panoramas (Matterport3D~\cite{matterport3d,dit360}, RealSee-3D~\cite{realsee3d}, 360+x~\cite{360x} scans).
Because we can synthesize pairs of perspective crops at controlled yaw,
pitch, and field of view, the yaw separation between the two views is
known exactly, which lets us derive ground-truth answers for many
spatial questions analytically rather than asking a vision-language
model to read them off pixels. Each panorama becomes a stream of
multiple-choice questions whose images are two perspective crops and
whose reasoning aid is the side-by-side composite of those crops.

We synthesize two question families on the same view-extraction
frontend, distinguished by who decides the answer. A template-driven
family covers cases where view metadata does not pin down the answer
(compass premises, measurement, object identification, scene
navigation, camera motion); a vision-language model writes the
question and picks the gold option, gated by metadata-consistency
checks on the subset where geometry is verifiable. For the geometric subset where metadata determines the answer, we use an analytical generation pipeline: the model only names objects and writes prose around an analytically fixed direction. The pipeline runs in four stages.

\begin{itemize}[leftmargin=1.2em, itemsep=0.5pt]
    \item 
\textbf{Step 1: Perspective view extraction.}
For each panorama we sample a base pair (left, right) sharing the
camera position but offset in yaw. Per-view FOV is drawn from
$U(60^\circ, 90^\circ)$, the desired angular overlap from
$U(-5^\circ, 50^\circ)$ (negative values indicate a gap), and a small
pitch jitter from $U(-5^\circ, 5^\circ)$. The yaw separation that
realizes the requested overlap is
\begin{align*}
  \mathrm{yaw}_{\mathrm{sep}} = \max\!\Big(0,\, \tfrac{1}{2}(\mathrm{fov}_L + \mathrm{fov}_R) - \mathrm{overlap}\Big),
\end{align*}
placed symmetrically about the panorama's reference yaw. Each crop is
rendered by direct equirectangular-to-perspective resampling: pixels
are lifted to camera-frame rays, rotated by the requested yaw and
pitch, converted back to spherical coordinates, and bilinearly sampled
from the panorama. No distortion model is needed because the source is
already rectified to equirectangular convention.

To support multi-image mental spatial reasoning and cannot-be-determined questions, we extend each scene with additional crops on both sides of the base pair. From each base view, we sample a chain of crops with random yaw offsets, stopping when a candidate overlaps with the opposite base view above a small threshold. The resulting auxiliary views tile the half-sphere away from the base pair without re-entering it. We save the side-by-side composite of the left and right chains as the reasoning-aid image attached to every question for the scene.

\item\textbf{Step 2: Question synthesis.}
The template catalog covers photographer-relative directions, compass
premises, object--object relations, camera motion (translation and
rotation), measurement, and object identification, plus a reserved
cannot-be-determined category for non-overlapping pairs. Templates are
sampled per scene with category-specific weights. For every scene we
compute camera deltas from the view metadata,
\begin{align*}
  \Delta\mathrm{yaw} = (\mathrm{yaw}_2 - \mathrm{yaw}_1 + 180) \bmod 360 - 180,
  \quad
  \Delta\mathrm{pitch} = \mathrm{pitch}_2 - \mathrm{pitch}_1,
\end{align*}
and inject them into the prompt in plain English (``the viewpoint
shifts to the right by $|\Delta\mathrm{yaw}|^\circ$'') so the model
has ground-truth camera motion when answering rotation or translation
questions. A forbidden-token guard rejects any output mentioning
stitching, panoramas, yaw, or file paths. Compass questions are
further constrained by a rule that a cardinal premise on view~1 fixes
only one axis: the perpendicular axis is determinable only with an
orthogonal reference, and ``in front of a wall'' is depth rather than
a cardinal direction.

For samples whose answer can be derived from yaw alone
under the equal-distance assumption that anchor and target sit at the
centers of their respective views at the same depth from the
panorama's optical center. Setting $\theta_A = 0$ and
$\theta_B = (\mathrm{yaw}_B - \mathrm{yaw}_A) \bmod 360^\circ$
normalized into $[-180^\circ, 180^\circ)$, the world-frame direction
from anchor to target is
\begin{align*}
  \mathrm{direction}_{\mathrm{world}} = \operatorname{atan2}\!\big(\sin\theta_B - \sin\theta_A,\; \cos\theta_B - \cos\theta_A\big),
\end{align*}
which is rotated into the anchor's local frame using a fixed lookup of
forward directions for a whitelist of oriented objects (TV, bed, sofa,
sink, kitchen island). Cardinal-chain templates instead carry a
cardinal premise from view A to view B by adding the clockwise yaw
delta and snapping to the nearest $45^\circ$ (8-direction) or
$90^\circ$ (4-direction) bin.

Pure yaw-based direction is fragile when anchor and target sit on
different walls. A yaw-clockwise target on the back wall is
geometrically back from the camera rather than to the right. For
relational templates that ask about a target in view B relative to an
anchor in view A, we replace the bearing rule with the depth-and-yaw
hybrid in \Cref{tab:depth-yaw}. The model is asked only for a 3-way
depth comparison (closer, same, or further), which we combine with
the sign of $\Delta\mathrm{yaw}$ to produce an 8-direction label. The
$25^\circ$ dead-zone reflects that base pairs are typically separated
by $20$--$80^\circ$: pairs below the dead-zone with a clear depth
signal read as straight front or back, while pairs in the dead-zone
with no depth signal yield no usable direction and are rejected.

\begin{table}[ht]
  \centering
  \caption{Depth--yaw mapping to 8-direction labels for the relational
  templates. The $25^\circ$ dead-zone rejects ambiguous pairs.}
  \label{tab:depth-yaw}
  \begin{tabular}{lccc}
    \toprule
    \textbf{Depth} & $|\Delta\mathrm{yaw}| < 25^\circ$ & $\Delta\mathrm{yaw} > 0$ & $\Delta\mathrm{yaw} < 0$ \\
    \midrule
    closer  & front  & front-right & front-left \\
    same    & reject & right       & left       \\
    further & back   & back-right  & back-left  \\
    \bottomrule
  \end{tabular}
\end{table}

A first pass typically leaves coverage gaps in some categories, which
we patch with targeted regeneration that re-runs the scene loop on a
restricted template pool. The regeneration shares the per-scene RNG
seeding, so its output concatenates with the main pass without
overlap.

\item \textbf{Step 3: Verification.}
Every candidate clears four cumulative gates before it is written. The
forbidden-leak guard disqualifies any stem or option that mentions
stitching, panoramas, yaw, or file paths. The direction--metadata
consistency check parses the answer text of verifiable templates
(camera rotation, translation, photographer-relative direction) for
direction keywords and compares them against $\Delta\mathrm{yaw}$ and
$\Delta\mathrm{pitch}$: yaw shifts above $3^\circ$ require a
horizontal direction with the correct sign and reject answers that
are purely vertical, and pitch shifts work analogously. Uniqueness
grounding sends every named noun phrase from the questions back to
the model in a separate call asking it to count visible candidate
matches in its image, and only objects flagged as uniquely present
survive; we found object naming from a single forward pass to be the
largest source of unverifiable references. Finally, a reverse-solve
gate re-runs each question blind against a fresh model call with
the exact images and options under a prompt that requires explicit
object localization, angular and depth judgment, 8-direction
classification, and option matching, and items where the
reverse-solver disagrees with the generator are dropped.

\item\textbf{Step 4: Post-processing.}
\label{app:panorama-post}
After CoT generation we apply a conservative line-by-line filter as a
final defense against CoT-level inconsistencies that the per-question
gates in Step 3 cannot see. The filter drops rows whose reasoning
chain contradicts the rotation direction encoded in the gold option,
whose CoT is empty or under twenty whitespace tokens, or whose gold
letter does not extract to a non-empty option text in the prompt.
\end{itemize}
\paragraph{Bird's-eye view synthesis.}
The next departure leaves the input agent's position altogether and
synthesizes a world-anchored overhead view. Source scenes come from
Mindcube~\cite{mindcube}. An overhead view makes object positions, distances,
and ego-relative directions directly readable, so the model can
consult it as an intermediate image during reasoning. To be usable as
a visual thought, the synthesized map must be geometrically faithful
so that the original answer survives the view substitution, stable
under sparse-view inputs, and realistic enough that a unified VLM
treats it as a photograph rather than a diagram.

\begin{figure}[ht]
  \centering
  \includegraphics[width=\linewidth]{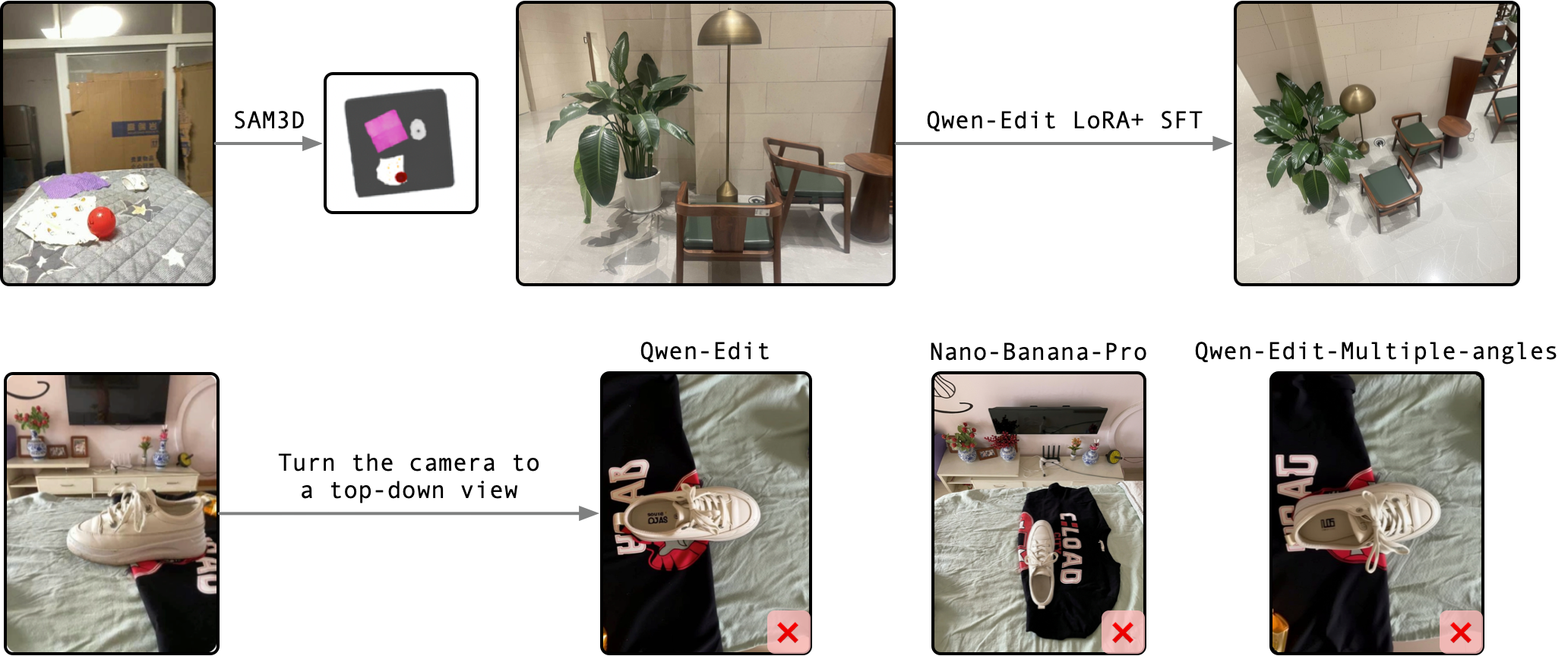}
  \caption{Representative failures of off-the-shelf and fine-tuned
  candidates on real scenes. \textit{Top:} a SAM3D-based composite of
  the small objects on the bed yields a generic synthetic-looking
  platform, and a Qwen-Edit model further fine-tuned (LoRA + SFT) on
  synthetic top-down pairs only resamples a different angle of that
  fabricated scene rather than depicting the actual input.
  \textit{Bottom:} given the prompt ``turn the camera to a top-down
  view'', Qwen-Edit, Nano-Banana-Pro, and a multi-angle Qwen-Edit
  fine-tune each return a wrong camera angle and drift in object
  orientation. The picture is consistent across many scenes.}
  \label{fig:candidate-failures}
\end{figure}

To our knowledge no public tool produces a sparse-view top-down image
of an indoor scene under these constraints. We tried three natural
candidates from adjacent fields, summarized in \Cref{tab:candidates}
and illustrated in \Cref{fig:candidate-failures}.

\begin{table}[ht]
  \centering
  \caption{Off-the-shelf candidates we considered for sparse-view
  top-down construction, and the way each one fails.}
  \label{tab:candidates}
  \small
  \begin{tabular}{>{\raggedright\arraybackslash}p{0.18\linewidth} >{\raggedright\arraybackslash}p{0.32\linewidth} >{\raggedright\arraybackslash}p{0.40\linewidth}}
    \toprule
    \textbf{Candidate} & \textbf{What it does} & \textbf{Why it fails in our setting} \\
    \midrule
    SAM3 + SAM3D &
    Detect objects, recover per-object 3D shape and pose, composite into a top-down view. &
    Requires an explicit object vocabulary, ingests views one at a time, and the rendered composite is visibly synthetic. \\
    \addlinespace
    Image-edit / view-conversion (Qwen-Edit, Nano Banana Pro) &
    Treat the top-down view as image-to-image editing. &
    Object orientations drift, items disappear, and the camera angle is inconsistent across runs. Fine-tuning Qwen-Edit on synthetic top-down pairs improves in-distribution scenes but still hallucinates content, and inference slows as the input view count grows. \\
    \addlinespace
    Novel-view synthesis (Gaussian Splatting, Stable Virtual Camera) &
    Reconstruct or interpolate views from the input set. &
    Built for small camera displacements; rotating to nadir is far outside their regime, and sparse inputs blur quickly under any large rotation. \\
    \bottomrule
  \end{tabular}
\end{table}

Candidates that are strong on appearance tend to be weak on geometric
consistency, and candidates that are strong on geometry produce
outputs that do not look photographic. We split the problem along
that boundary. A reconstruction stage commits to where objects are
and how they are oriented relative to a horizontal support plane, and
a generative stage paints that layout to look photographic.
Steps 1--3 build the geometric layout, and Step 4 renders it and
paints in the texture.

\begin{itemize}[leftmargin=1.2em, itemsep=0.5pt]
    \item 
\textbf{Step 1: Filtered point-cloud reconstruction.}
Let $\mathcal{I} = \{I_t\}_{t=1}^{T}$ denote the input views. Running
Pi3 on $\mathcal{I}$ produces dense 3D points with per-point
confidence scores $w_i \in [0,1]$, RGB colors copied from the source
pixels, and camera-to-world matrices. We discard low-confidence
points and points adjacent to depth discontinuities. The latter
removes the boundary halos Pi3 produces around silhouettes, which
would otherwise bleed into the top-down rendering. The result is a
colored point cloud
\begin{align*}
  \mathcal{P} = \{(\mathbf{x}_i,\, \mathbf{c}_i,\, w_i)\}_{i=1}^{N},
\end{align*}
which we subsample with probability proportional to $w_i$ to bias
survival toward high-confidence geometry.

\item 
\textbf{Step 2: Scene anchor and rendering radius.}
We need a center $\mathbf{o}$ for the overhead camera to look at and
a radius $R$ that sets its framing. Both are estimated coarse-to-fine
so distant outliers cannot drag either off the scene.

For the coarse center we take the least-squares intersection of the
input camera rays,
\begin{align*}
  \mathbf{q}_0 = \arg\min_{\mathbf{q}} \sum_t
  \big\|(\mathbf{I} - \mathbf{d}_t \mathbf{d}_t^\top)(\mathbf{q} - \mathbf{o}_t)\big\|_2^2,
\end{align*}
where $\mathbf{o}_t \in \mathbb{R}^3$ and
$\mathbf{d}_t \in \mathbb{R}^3$ are the origin and unit direction of
ray $t$, $\mathbf{I}$ is the $3 \times 3$ identity, and
$\mathbf{I} - \mathbf{d}_t \mathbf{d}_t^\top$ is the projection onto
the plane orthogonal to $\mathbf{d}_t$. The minimizer $\mathbf{q}_0$
is the approximate point of common visual focus. When camera poses
are missing we fall back to the confidence-weighted centroid of the
cloud.

To separate foreground from background, we sort point distances to
$\mathbf{q}_0$ and look for the first large gap relative to the
median adjacent difference. The radius before that gap is taken as
the foreground extent and the rest is discarded as distant
background. We then refine the center to the confidence-weighted
mean over the foreground subset $\mathcal{F}$ and set $R$ to a fixed
percentile of distances from $\mathcal{F}$ to the refined center.
Using a percentile rather than the maximum keeps framing stable
across scenes by ignoring the tail of weakly reconstructed points.

\item 
\textbf{Step 3: Overhead camera frame.}
We recover the top-down direction by fitting the dominant horizontal
plane to the cloud with RANSAC (PCA fallback on degenerate clouds).
When camera poses are available we sharpen the fit with two
pose-aware constraints: the candidate normal must roughly align with
the average camera up direction, and most camera centers must lie on
the same side of the plane. Together these push the fit toward the
actual support surface and reject vertical structures such as walls.
The normal's sign is resolved by flipping it toward the side that
contains more cameras.

With the oriented normal $\mathbf{n}$ we place the overhead camera at
$\mathbf{t}_{\mathrm{top}} = \mathbf{o} + h\mathbf{n}$ with forward
direction $-\mathbf{n}$ and height $h = sR$, optionally clamped so
the camera clears any tall reconstructed geometry. We align its
right axis with the reference camera when one is available, so all
rendered top-downs share a consistent orientation across scenes;
otherwise we derive the right axis from a canonical world up vector.
Camera intrinsics are a standard pinhole, with the field of view
widened on square outputs to avoid clipping the source aspect ratio.

\item 
\textbf{Step 4: Rendering and photographic texturing.}
We project the surviving points into the overhead camera using
standard pinhole projection, drop points behind the camera, and
resolve per-pixel collisions by keeping the nearest hit. The result
is geometrically faithful but visually sparse, so it serves as
structural conditioning rather than as the final image. As a cheap
consistency check, we also render the point cloud from the original
reference camera pose and discard the scene if the cloud and the
pose disagree there, before paying the cost of texturing.

A LoRA-fine-tuned Qwen-Image-Edit-2511~\cite{qwenedit} turns the sparse rendering
into a photographic top-down. We feed the sparse rendering as the
image-conditioning input and train the LoRA~\cite{hu2021loralowrankadaptationlarge} to reproduce a real
overhead photograph of the same scene. The training set is a
hand-collected corpus of 25 scene groups, each containing the
multi-view inputs, the reconstructed point cloud, and a real
overhead photograph, used as (sparse rendering, photograph) pairs.
A narrow rank-32 LoRA combined with heavy resampling lets the model
pick up our rendering style without erasing its pretraining. Full
hyperparameters are in \Cref{tab:lora-config}. At inference we feed
the sparse rendering with a fixed top-down instruction prompt and
read off the photographic image.

\begin{table}[ht]
  \centering
  \caption{LoRA training configuration for the photographic texturing
  model.}
  \label{tab:lora-config}
  \small
  \begin{tabular}{p{0.30\linewidth} p{0.62\linewidth}}
    \toprule
    \textbf{Setting} & \textbf{Value} \\
    \midrule
    Base model & Qwen-Image-Edit-2511 DiT (trainable); Qwen-Image text encoder and VAE (frozen) \\
    Adapter & LoRA, rank 32 \\
    LoRA target modules (DiT) & attention $q$/$k$/$v$ projections (including cross-modal variants for text--image fusion), output projections, MLP layers, modulation layers \\
    Precision & bf16 with gradient checkpointing \\
    Learning rate & peak $10^{-4}$, cosine decay to $10^{-6}$, 100-step warmup \\
    Schedule & 20 epochs; each example resampled 20 times per epoch \\
    Input resolution & up to $1024^2$ pixels \\
    Hardware & 4 GPUs, data-parallel \\
    \bottomrule
  \end{tabular}
\end{table}

For each scene we save the photographic top-down, the sparse
rendering as a structural reference, the colored point cloud, and a
metadata file with the refined center, rendering radius, plane
normal, and camera intrinsics and extrinsics, sufficient to project
3D annotations onto the top-down image without rerunning
reconstruction.
\end{itemize}
\paragraph{Ego--exo perspective transfer (Ego-Exo4D).}

Ego-Exo4D~\cite{egoexo4d} records every activity simultaneously with a
head-mounted Aria camera and several stationary GoPro cameras, so
the photographer's first-person view is paired frame-for-frame with
several third-person views of the same scene. Given a third-person
photograph in which the photographer is visible, the answer to
``what is to your left from this person's perspective?'' reduces to
a direct ego-frame fact.

We convert each Ego-Exo4D take into multiple-choice
perspective-taking questions. The third-person image is the model's
input, and the photographer's ego frame is attached as a reasoning
aid. The third-person image used as input is unmarked. We do draw a
green marker on the photographer's projected head, but only on a
parallel copy of the same exo frame, kept in metadata for reviewer
inspection. Marking the input would let any model identify the
photographer trivially from color and bypass the perspective-taking
signal.

The pipeline runs in three steps. First, a per-take nine-pack exports
three ego frames, three exo frames, and three exo-marked frames.
Then a two-stage GPT pipeline writes one or two perspective-taking
questions per frame. A deterministic post-processor then selects a
balanced subset across takes and answer axes.
\begin{itemize}[leftmargin=1.2em, itemsep=0.5pt]
\item 
\textbf{Step 1: Nine-pack construction.}
For every take we select three ego frames that jointly maximize
in-focus quality and visual diversity. Per-frame sharpness is the
variance of the Laplacian on a central
$\tfrac{3}{4}\!\times\!\tfrac{3}{4}$ crop of the rectified ego
frame. Visual diversity is one minus the maximum cosine similarity
between $96 \times 96$ z-normalized appearance vectors. Picks are
made greedily by maximizing
$0.45 \cdot \mathrm{sharpness} + 0.55 \cdot \mathrm{novelty}$ subject
to a minimum temporal gap between selected frames (default 120
frames at 30\,fps), which prevents the top-$k$ from collapsing onto
a single sub-second cluster on temporally redundant activity videos.

The Aria RGB stream is fisheye and is rectified to a regular
perspective view before scoring. When the official Project Aria
calibration is available we use it directly; otherwise we estimate
fisheye disc parameters from the first frame and resample onto a
perspective grid. The exporter runs the same rectification at two
resolutions: a smaller one for frame scoring and a larger one for
the on-disk image.

Drawing the marker on the third-person frame requires projecting the
photographer's 3D world position into the exo image at the selected
ego frame. Using the per-take time synchronization, we map each
ego-frame index to a trajectory time and read the device world
position $\mathbf{p}_{\mathrm{world}}$ off the closed-loop
trajectory. We then transform $\mathbf{p}_{\mathrm{world}}$ into the
exo camera frame as
$\mathbf{p}_{\mathrm{cam}} = \mathbf{R}_{\mathrm{cam,world}}(\mathbf{p}_{\mathrm{world}} - \mathbf{t}_{\mathrm{world,cam}})$
with the GoPro extrinsics, and project it through the
Kannala--Brandt fisheye model. Writing
$\mathbf{p}_{\mathrm{cam}} = (x, y, z)$, the model maps it to image
coordinates $(u, v)$ through
\begin{align*}
  a = x/z, \quad b = y/z, \quad r = \sqrt{a^2 + b^2}, \quad \theta = \arctan r,
\end{align*}
\begin{align*}
  \theta_d = \theta + k_0\theta^3 + k_1\theta^5 + k_2\theta^7 + k_3\theta^9,
\end{align*}
\begin{align*}
  (u, v) = \big(f_x \cdot (\theta_d/r) \cdot a + c_x,\; f_y \cdot (\theta_d/r) \cdot b + c_y\big),
\end{align*}
where $(f_x, f_y)$, $(c_x, c_y)$, and $\{k_0, \ldots, k_3\}$ are
the per-camera intrinsics. A point projects to a marker only when
it lies in front of the camera ($z > 0$) and $(u, v)$ falls inside
the exo frame. Otherwise no marker is drawn and the frame is
exported with the marker layer suppressed. Takes with missing or
invalid timesync, trajectory, or calibration files are skipped by
default.

\item 
\textbf{Step 2: Two-stage VQA generation.}
From each frame in the nine-pack we generate up to two
perspective-taking questions. The single-object axis question picks
one visible object near a single edge of the photographer's view,
with the answer drawn from \{left, right, above, below\}. The
object-pair relation question picks a target and a reference, with
the answer drawn from \{left, right, above, below, front, behind\}.
Both branches use a two-stage GPT pipeline that routes images so
that the spatial fact and the question wording are decided in
different stages from different views.

In the candidate stage the model sees only the ego image and
proposes the spatial fact: a target object (and, for the pair
branch, a reference) plus the chosen direction label. Question
wording is forbidden at this stage. In the finalize stage the model
sees only the exo image with the marker on the photographer and
rewrites the candidate fact into a perspective-taking question,
naming the photographer naturally by clothing, role, or position,
picking an opener from a small bank of templates, and forcing the
either-or option form. The finalize stage never sees the ego image,
so it cannot redo the spatial reasoning, only re-phrase it.

This split removes two failure modes we saw when a single prompt
did both: the model invented spatial answers to match the marker
rather than the ego frame, and the question stem leaked the answer
through phrases like ``the person is on the left, so X must be on
his left''. With the split, the spatial fact is fixed by the
candidate stage and the finalize stage is constrained to keep the
same target and chosen axis. Each candidate's chosen axis fixes a
binary either-or option pair (left/right, above/below, or
front/behind), which determines the gold letter directly.

\item 
\textbf{Step 3: Balanced subset selection.}
The raw corpus is naturally imbalanced. Single-object axis questions
skew toward whichever direction the activity faces in each take,
and some takes contribute many more frames than others. A
deterministic post-processor produces a balanced subset by greedy
selection.

We first reduce to at most one row per (take, frame) by random
selection between the single-object and pair candidates of the same
frame, so no frame contributes twice. We then pick rows in two
phases. Phase 1 walks the takes in shuffled order and adds, from
each take, the row whose addition would most reduce the imbalance
across the three axis pairs (left vs.\ right, above vs.\ below,
front vs.\ behind), with under-represented pair totals and smaller
absolute counts as tie-breakers. This yields at most one row per
take in an order that keeps the answer distribution close to
per-axis balanced. If the target subset size is larger than the
number of distinct takes, Phase 2 pulls from the remaining pool
with the same scoring rule but without the per-take constraint. The
end result is a subset with near 50/50 splits on each axis pair and
a maximally diverse take base for the requested size.
\end{itemize}
\paragraph{Alternate-view rendering (SAM3D).}
When only a single image is available, we cannot rely on a paired
second camera. Instead we lift the image to 3D, pick any directional
entity in the scene (a person, a chair, a TV) as the viewer, and
render the rest of the scene from the viewer's inferred pose. The
supervision target is a question like ``from the chair's point of
view, is the lamp to the left or right of the table?''. Every gold
answer is computed analytically from 3D vector geometry. The
language model, when invoked at all, only plays validator and
natural-language polisher.

The construction emits three layered reasoning images per question:
a who-is-who image showing role-colored masks on the original
image, an orientation image showing the viewer's basis pasted at
the viewer's centroid, and the viewer-perspective render with the
viewer hidden. A model trained to consume them works through them
as an interleaved visual chain of thought during reasoning. The
pipeline runs in four steps. Before describing them we note the
front-end choice.

\begin{itemize}[leftmargin=1.2em, itemsep=0.5pt]
\item 
\textbf{Detection and segmentation front end.}
Before settling on RAM++~\citep{ram}, SAM3~\cite{sam3}, and SAM3D-Objects~\cite{sam3d} we tried several
alternatives. Combining RAM++ with Grounding DINO or Florence-2
produced boxes that were too loose for clean single-object splats.
SAM3 alone gave high mask quality when it fired, but recall was too
low: many RAM++ tags never produced a usable instance, which blocked
triplet discovery. A hybrid that piped RAM++ through WeDetect into
SAM3 improved coverage somewhat, but WeDetect boxes were often
clipped at the object boundary, and those truncated masks
propagated into incomplete 3D geometry. The
RAM++ $\to$ SAM3 $\to$ OriAnyV2~\cite{orianyv2} stack we ship traded a small loss in
vocabulary recall for tighter masks and reliable orientation, which
matters more for downstream rendering.

\item 
\textbf{Step 1: 2D perception.}
For each input image we run RAM++ to produce a list of
open-vocabulary tags, then call SAM3 once per tag with the tag as
the segmentation prompt to obtain per-instance masks at the
original image resolution. A tag-aware non-maximum suppression
removes both within-tag duplicates (a single chair detected many
times) and tag-internal hierarchical duplicates (a sofa mask
containing its cushions). Any mask whose bounding box covers more
than $60\%$ of the image is dropped, since these are almost always
background panels (walls, floors, sky) that the tagger picked up.

For every surviving instance we predict its 3D orientation with
OriAnyV2, which returns the object's canonical-front azimuth,
elevation, and roll relative to the camera frame, together with a
binary directional-front flag. Only instances flagged as directional
(TVs, sofas, beds) can serve as viewers whose forward direction is
read off the predicted orientation; non-directional instances
(spheres, bowls, lamps) either play the target role or, if used as
the viewer, fall back to a look-toward-the-targets view direction.

\item 
\textbf{Step 2: Per-instance 3D reconstruction.}
Each (image, instance mask) pair is reconstructed independently
into a 3D Gaussian splat by SAM3D-Objects. For this branch of the
corpus we use MS COCO 2017~\cite{lin2015microsoftcococommonobjects} (train and val splits) as the only
image source, so reconstruction statistics stay tied to a single
domain. The model outputs a per-object Gaussian set in a local
frame together with a rigid-plus-isotropic transform
$(\mathbf{R}, \mathbf{t}, s)$ that takes the object into a shared
canonical frame. We compose multiple objects into a scene by
applying these transforms to the per-Gaussian centers, rotating
each Gaussian's covariance quaternion accordingly, and concatenating
the Gaussian buffers. Per-object scales are clamped to be at least
$1.1\times$ the merged kernel size to absorb mild scale
inconsistency between independent reconstructions. After composition
we apply a fixed axis-swap to align the scene with a $+Y$-up
convention and re-normalize so the active Gaussians' largest extent
is unit length. Reconstructing per-instance rather than fitting a
whole scene at once keeps each splat small and decouples
reconstruction failures across unrelated regions, at the cost of
mild per-object scale drift that the clamping and re-normalization
absorb.

\item 
\textbf{Step 3: Viewer-perspective rendering.}
For each candidate viewer--target triplet (one viewer plus the two
targets that will appear in the question) we place a virtual camera
at the viewer's center, looking along the viewer's forward
direction, and render the remaining objects through that camera.
For directional viewers the forward direction is the negative of
the first column of the predicted orientation matrix (the camera
looks out of the object's front); for non-directional viewers it is
a look-at vector toward the centroid of the targets.

We jointly fit the camera position and field of view so that the
targets fit with a small margin and the viewer does not clip the
camera. A small ladder of backoff distances (multiples of the scene
diagonal) and angular widths is swept, and we accept the smallest
pair $(\mathrm{backoff}, \mathrm{fov})$ for which every target AABB
corner $\mathbf{v} = \mathbf{p} - \mathbf{c}_{\mathrm{cam}}$
satisfies
\begin{align*}
  \operatorname{atan2}\!\big(|\mathbf{v}\cdot\mathbf{r}|,\; \mathbf{v}\cdot\mathbf{f}\big) < \mathrm{fov}/2
  \quad \text{and} \quad
  \mathbf{v}\cdot\mathbf{f} > 0,
\end{align*}
where $\mathbf{f}$ is the camera forward axis and
$\mathbf{r} = \mathbf{f} \times \mathbf{u}$ its right axis built
from the global up vector. The smallest fitting pair preserves the
illusion that the camera sits at the viewer object. Larger
backoffs and wider angular extents are only allowed when no smaller
pair fits.

We hide the viewer in the render by zeroing the opacity of its
Gaussians, so the viewer disappears from the image while the rest
of the buffer stays intact.

Two sanity checks finalize the candidate. Each target's AABB must
project to a non-empty bounding box in the rendered view (i.e.,
lie in front of the camera and inside the image), and the two
target bboxes must not overlap in render coordinates, since
overlap means the question is not visually answerable from the
render alone. Candidates failing either check are dropped.

\item 
\textbf{Step 4: Question synthesis.}
For each accepted candidate we emit one or two multiple-choice
questions from a small set of deterministic templates. The current
templates ask (i) whether one target is to the left or right of the
other from the viewer's standpoint, and (ii) which of the two
targets the viewer is facing more toward. Both gold answers are
computed analytically from the camera basis $(\mathbf{f}, \mathbf{r})$
and the relative vectors
$\mathbf{rel}_X = \mathbf{c}_X - \mathbf{c}_{\mathrm{viewer}}$.

For the left/right question we project each target's offset onto
the viewer's right axis,
\begin{align*}
  x_a = \mathbf{rel}_a \cdot \mathbf{r}, \quad
  x_b = \mathbf{rel}_b \cdot \mathbf{r},
\end{align*}
and pick ``left'' iff $x_a < x_b$, subject to a minimum normalized
margin
$|x_a - x_b| / (\|\mathbf{rel}_a\| + \|\mathbf{rel}_b\|) \ge \delta_{\mathrm{LR}}$
so close-call configurations are rejected. For the facing question
we measure each target's angular offset from the viewer's forward
axis,
\begin{align*}
  \mathrm{angle}_X = \arccos\!\big((\mathbf{rel}_X \cdot \mathbf{f}) / \|\mathbf{rel}_X\|\big),
\end{align*}
and pick whichever target has the smaller angle, subject to a
minimum angular margin between the two angles. Rejecting close-call
candidates makes the gold answers safe to use as supervision, since
a small reconstruction error nudging the two targets in opposite
directions cannot silently flip the answer.

A vision-language model is invoked twice, optionally. It can
validate a candidate by viewing the three reasoning artefacts
produced in Step 3 and either accepting it or rejecting it as
visually ambiguous, under a system prompt that asks for
conservative rejection. It can also polish the question stem into
more natural language while the gold answer and option pair stay
frozen. The model is never given authority over the answer.

Object phrasings in the stem are disambiguated automatically.
Viewer choice is restricted to instances the stem can refer to
unambiguously, which means singletons of their tag, instances with
a unique color within their tag group, or strictly leftmost or
rightmost in bbox order. When multiple instances of the same tag
remain, the chosen one is referred to as left, right, leftmost,
rightmost, or middle by its centroid-x rank within the group. The
stem therefore stays unambiguous even when several chairs sit in
the same scene.
\end{itemize}

\subsubsection{Object manipulation}

The remaining family of mental world modeling keeps the camera
fixed and transforms the object itself. We supervise the simplest
case: rigid rotation of the object about a vertical axis, which the
model must imagine and produce as a second view of the same object
from another angle.

\paragraph{Object rotation (Viewpoint-100K).}
The model is given one image of an object and the supervision
target is a second view of the same object after a horizontal
rotation, with mostly the same subject framing. The supervision
comes from Viewpoint-100K~\cite{zhan2025actial0}, which provides one hundred
thousand calibrated pairs of multi-view frames of the same physical
object, built from MVImgNet~\cite{yu2023mvimgnet} (roughly 6.5M real multi-view
frames with object masks, intrinsics, extrinsics, depth, and point
clouds per subject). For each sample, two views of the same
instance are drawn with horizontal separation between $20^\circ$
and $100^\circ$, over 10{,}813 distinct object identities across
205 classes.

We repackage those pairs as edit-the-input examples. The first
frame is the image to be transformed, the second is the supervision
target or auxiliary visual thought, and the text prompt describes
the relative rotation in natural language rather than exposing raw
calibration tuples. The geometric relationship between the pair is
unchanged from the public release; only the prompt and field
layout are adapted to our trace format.

\section{\instruct{} Examples}
\label{app:appendix-instruct-examples}

\raggedbottom
\noindent We show Text--Image--Text traces for each \instruct task, organized by cognitive level: one per basic visual perception task and two per spatial task. Each card pairs the input scene with its rendered intermediate image; Thought~0 (under the input) shows the model's plan before generation, Thought~1 (under the intermediate) describes the synthesized view. The question stem and gold answer appear at the bottom of each card.

\medskip
\subsection*{Level 1 (basic visual perception)}
\noindent Tasks at this level test direct reading of single-image content---locating a target among distractors, or recovering a global layout from shuffled fragments.
\par\medskip

\subsubsection*{Visual search (L1)}

{\centering\includegraphics[width=0.85\linewidth]{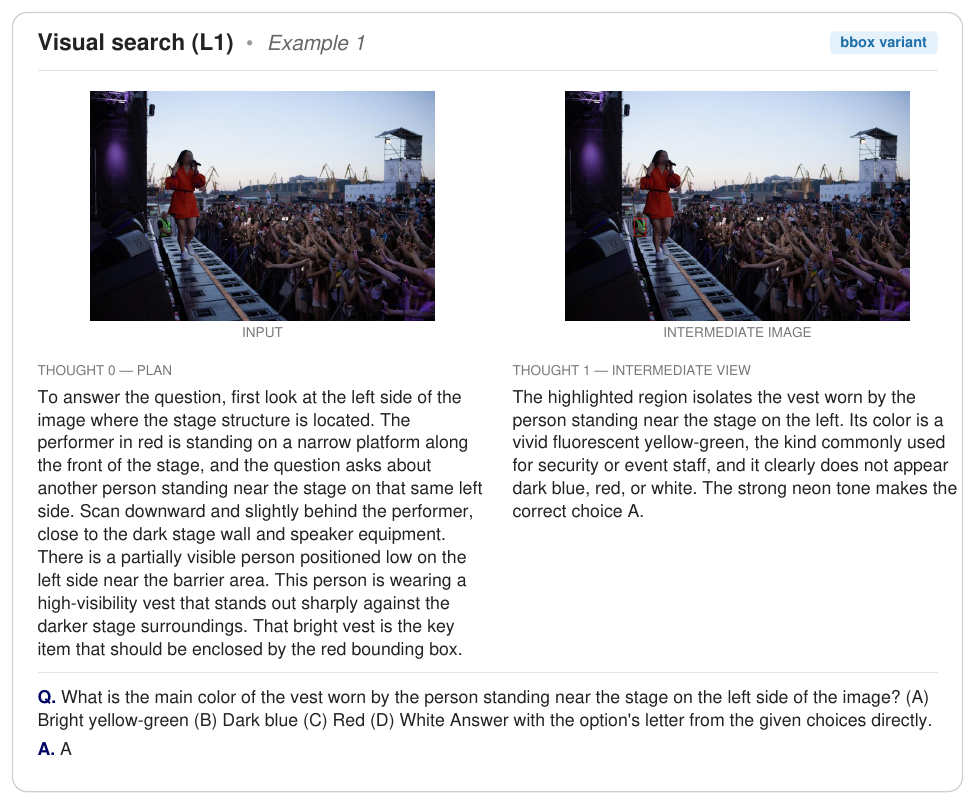}\par}\medskip

\medskip
\subsubsection*{Jigsaw QA (L1)}

{\centering\includegraphics[width=0.85\linewidth]{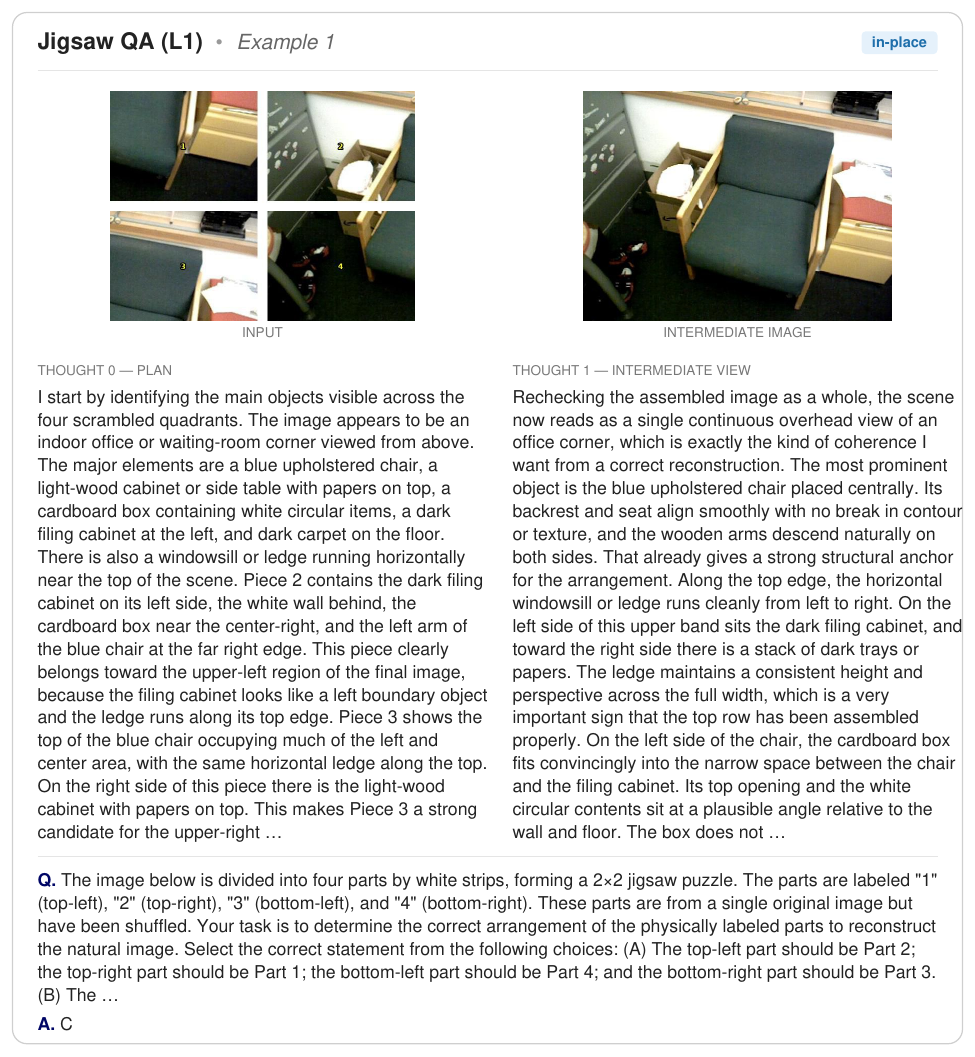}\par}\medskip

\medskip
\subsection*{Level 2 (static spatial understanding)}
\noindent Tasks at this level reason over spatial relationships within the observed scene---object counts and relative depths between visible regions.
\par\medskip

\subsubsection*{Spatial counting (L2)}

{\centering\includegraphics[width=0.85\linewidth]{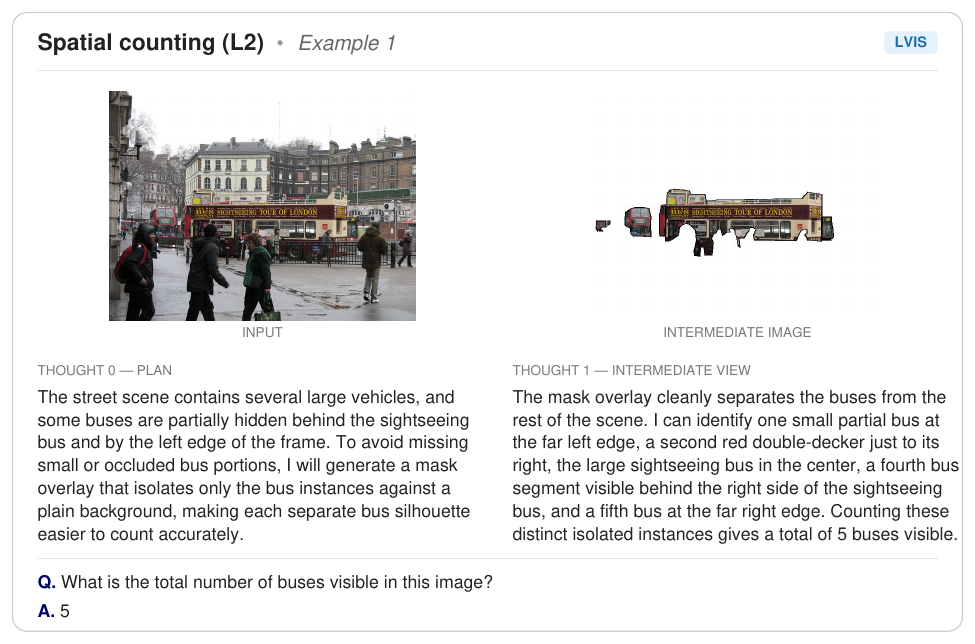}\par}\medskip
{\centering\includegraphics[width=0.85\linewidth]{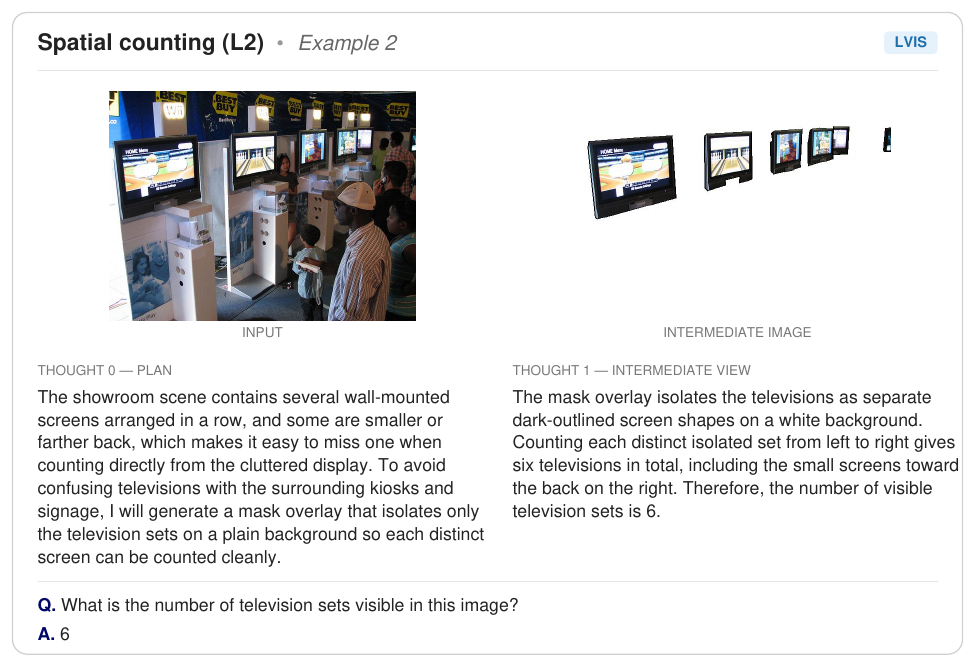}\par}\medskip

\medskip
\subsubsection*{Depth reasoning (L2)}

{\centering\includegraphics[width=0.85\linewidth]{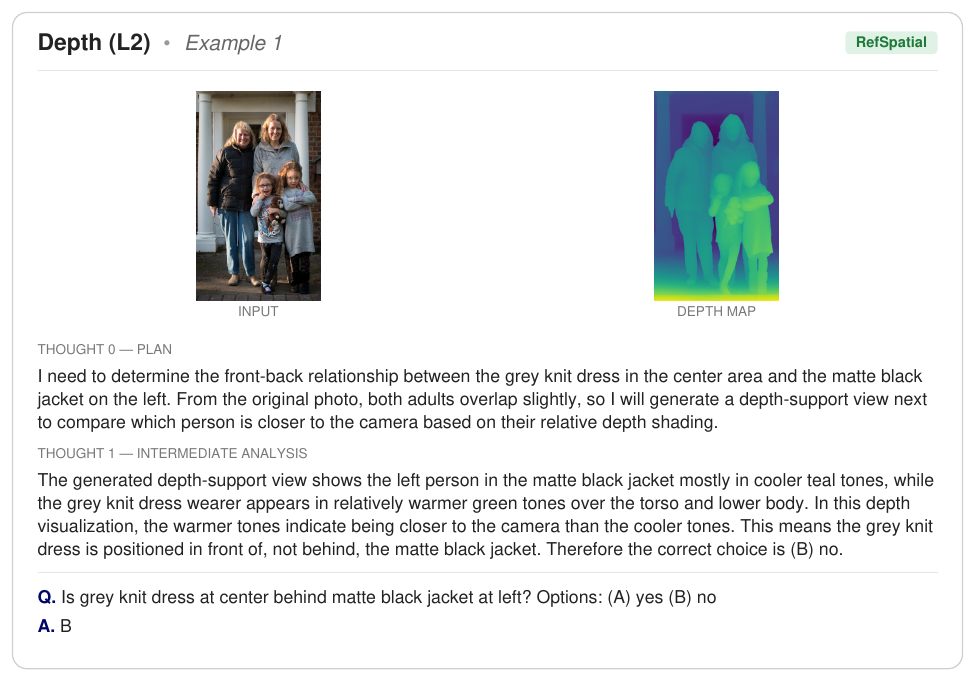}\par}\medskip
{\centering\includegraphics[width=0.85\linewidth]{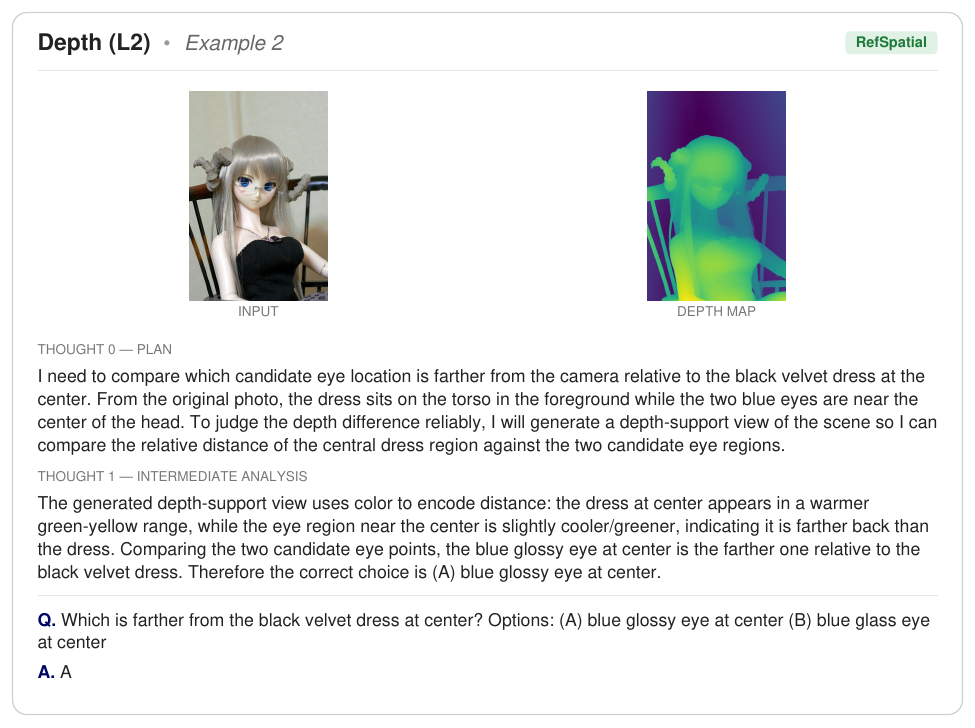}\par}\medskip

\medskip
\subsection*{Level 3 (mental world modeling)}
\noindent Tasks at this level construct a mental scene model that extrapolates beyond the observed input---novel viewpoints, top-down maps, ego/exocentric switches, and rigid-body rotations.
\par\medskip

\subsubsection*{BEV synthesis (L3)}

{\centering\includegraphics[width=0.85\linewidth]{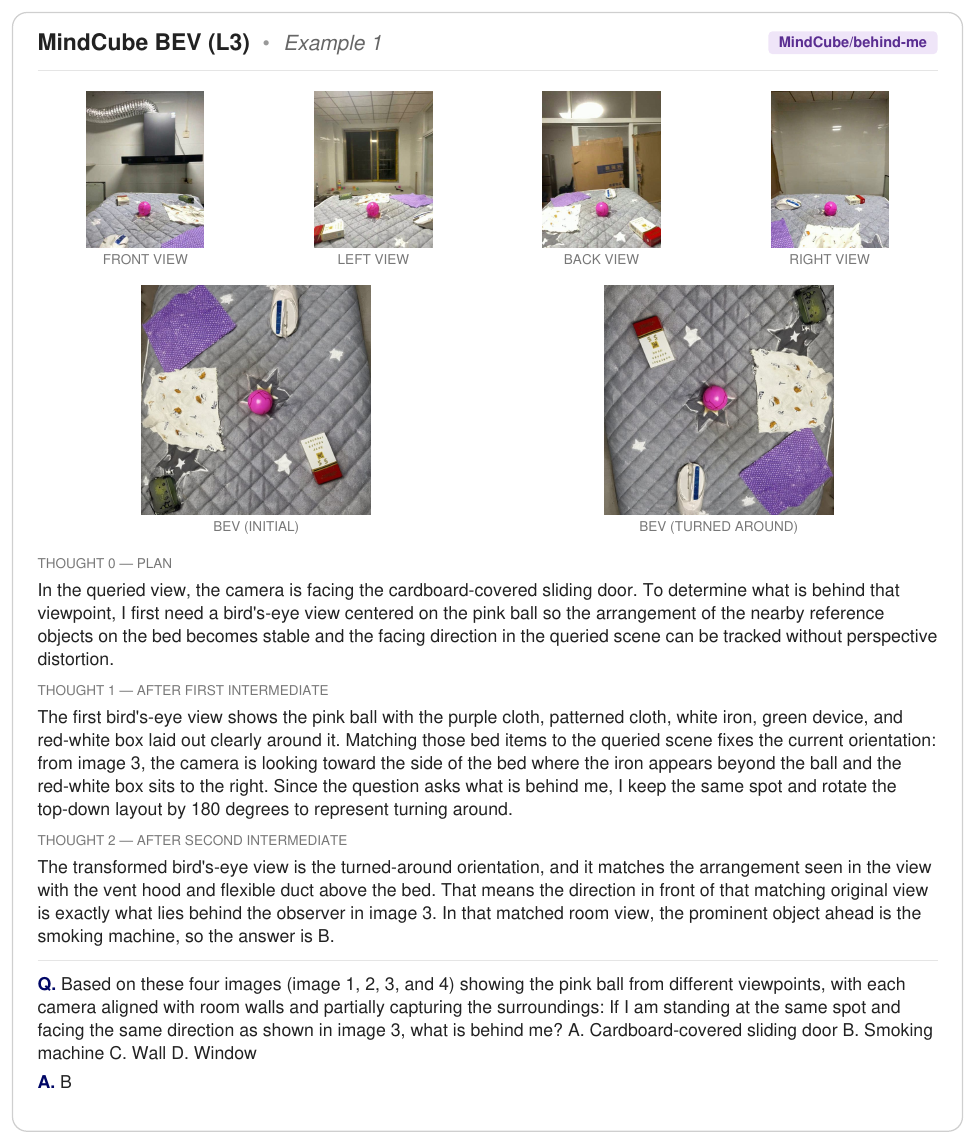}\par}\medskip
{\centering\includegraphics[width=0.85\linewidth]{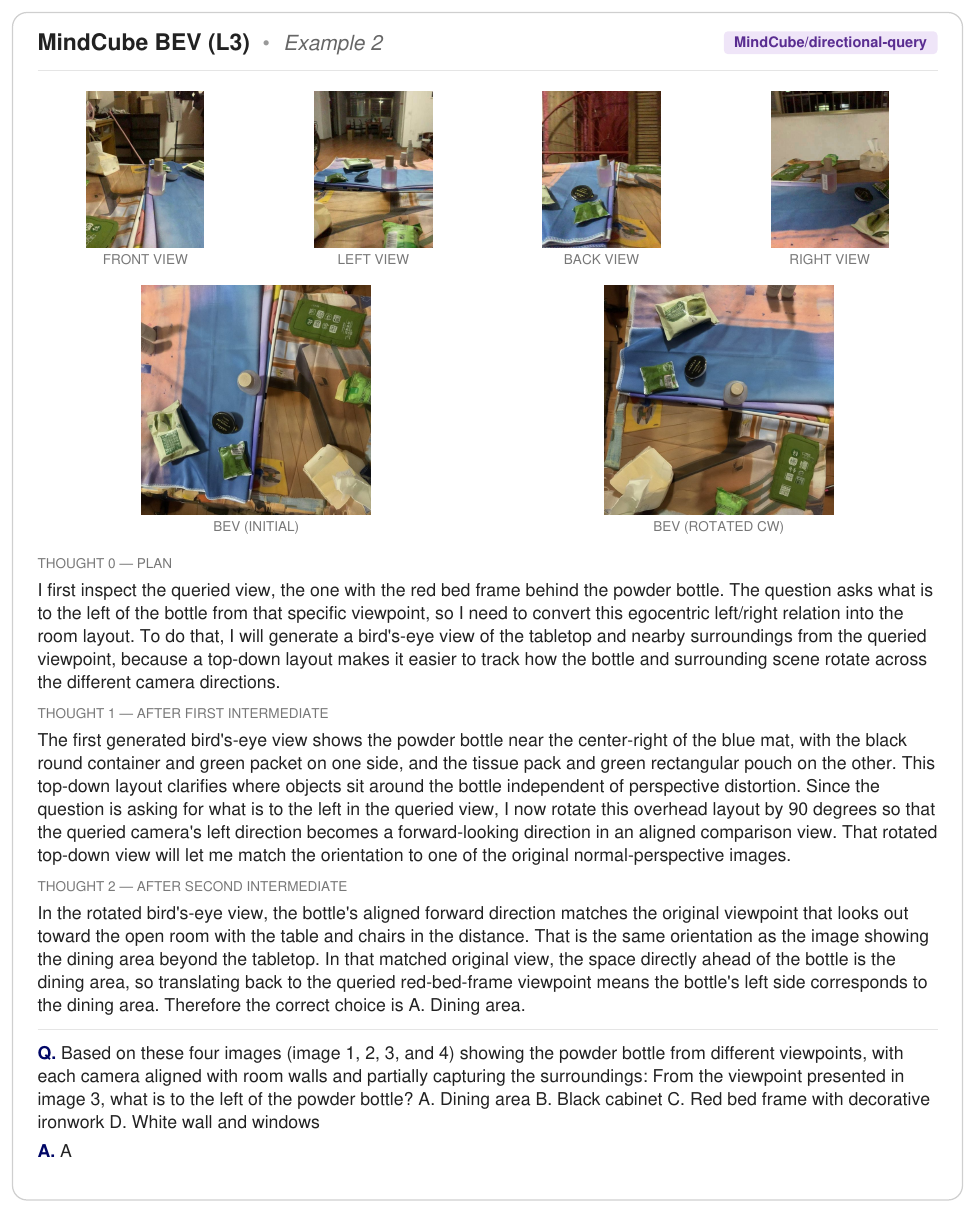}\par}\medskip

\medskip
\subsubsection*{Panorama (L3)}

{\centering\includegraphics[width=0.85\linewidth]{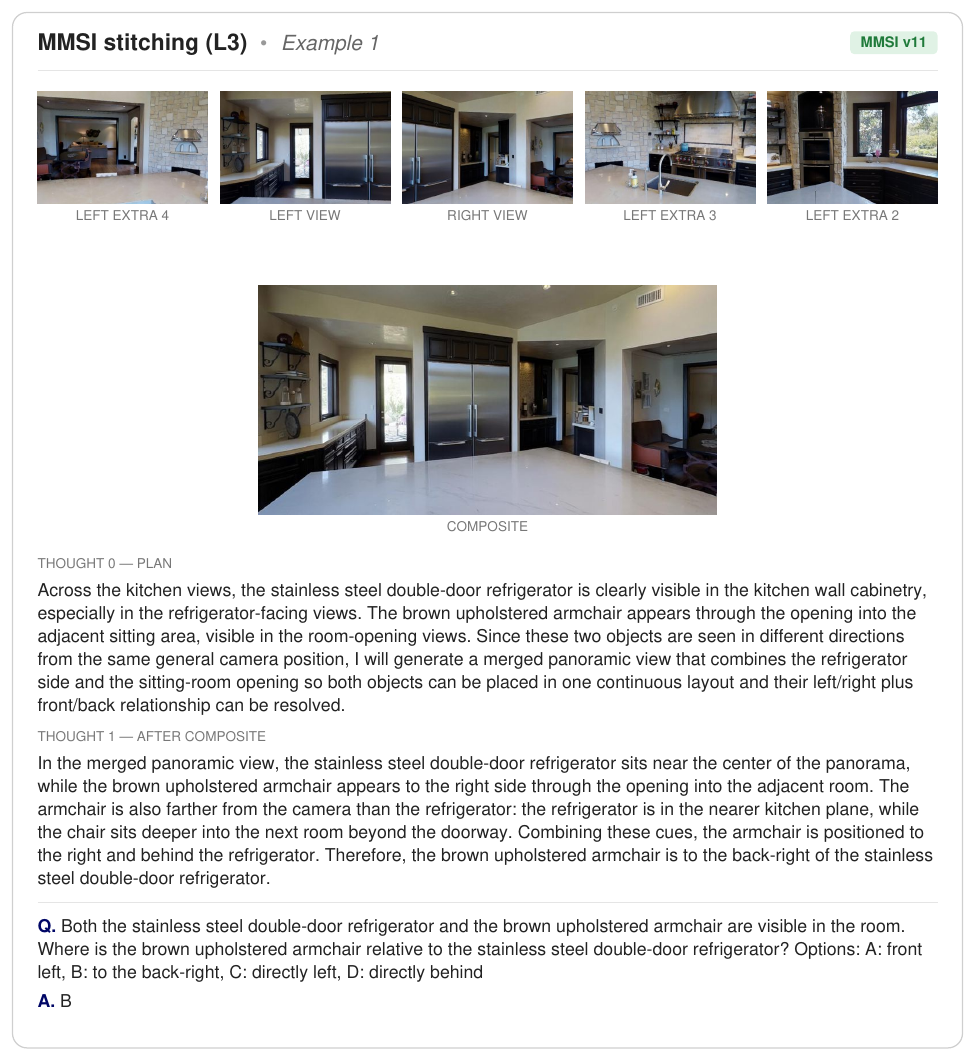}\par}\medskip
{\centering\includegraphics[width=0.85\linewidth]{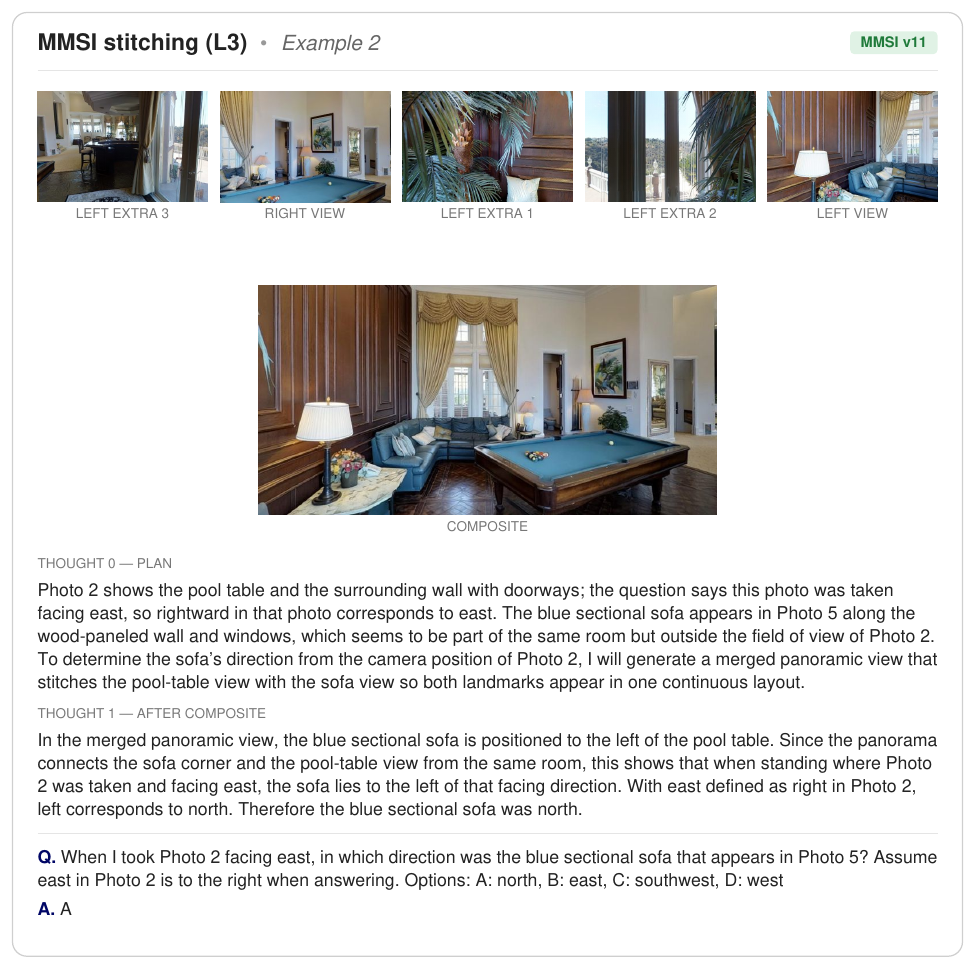}\par}\medskip

\medskip
\subsubsection*{Alternate-view rendering (L3)}

{\centering\includegraphics[width=0.85\linewidth]{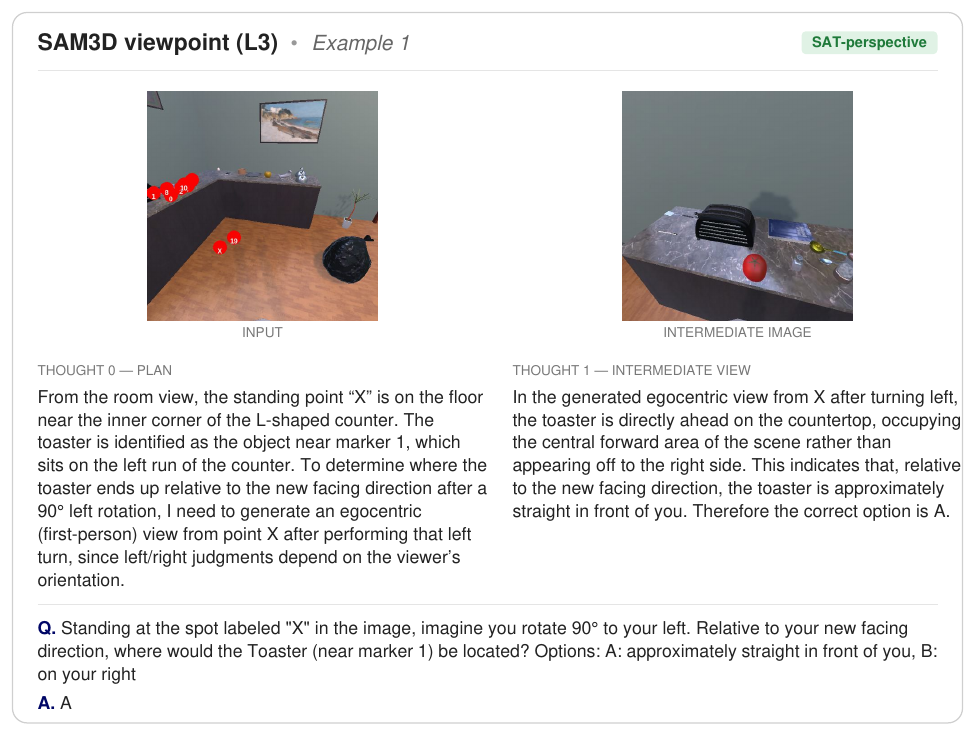}\par}\medskip
{\centering\includegraphics[width=0.85\linewidth]{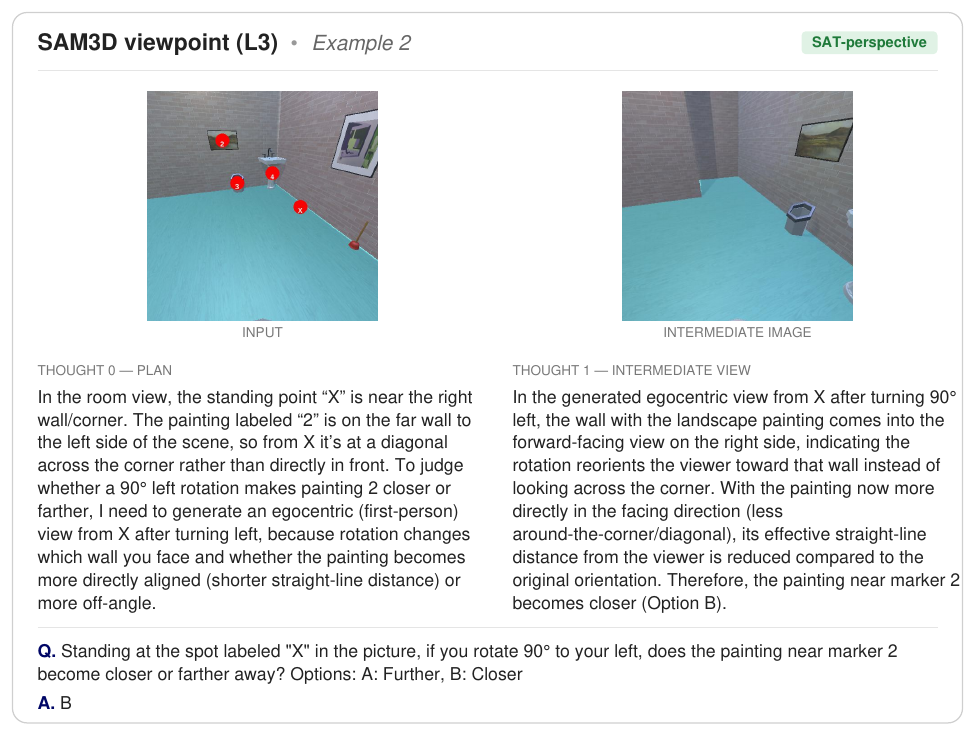}\par}\medskip

\medskip
\subsubsection*{Object rotation (L3)}

{\centering\includegraphics[width=0.85\linewidth]{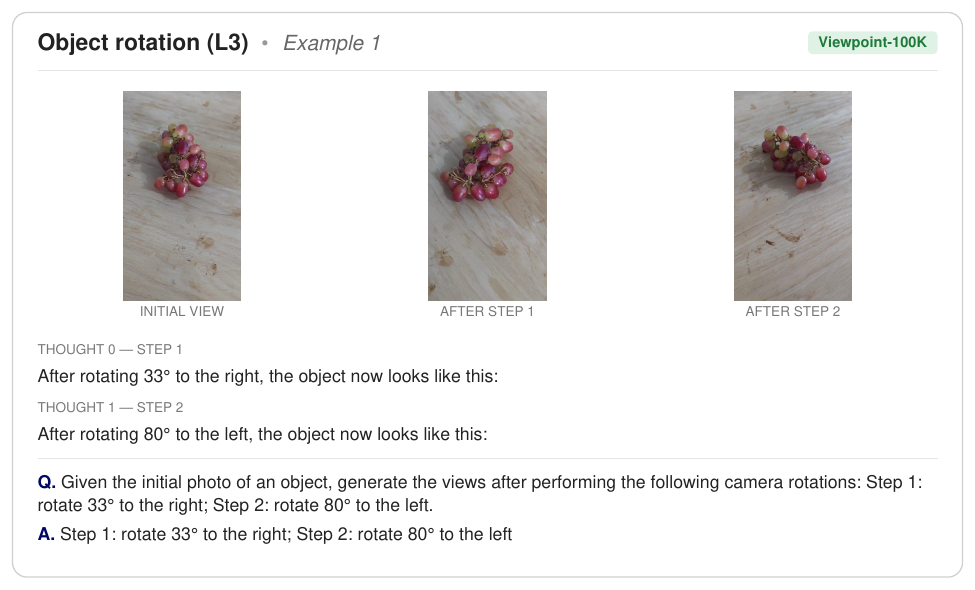}\par}\medskip
{\centering\includegraphics[width=0.85\linewidth]{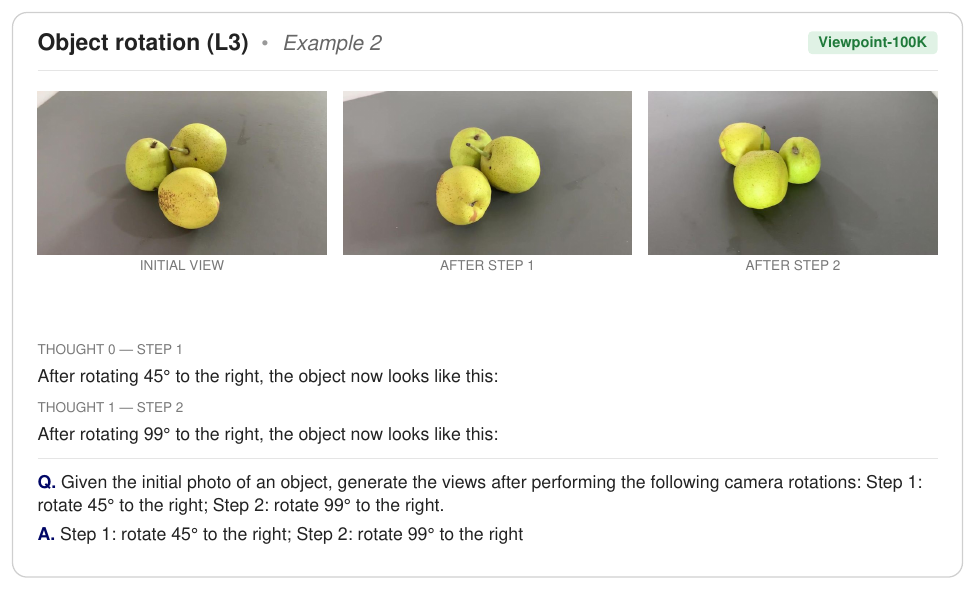}\par}\medskip

\medskip
\subsubsection*{Ego-exo transfer (L3)}

{\centering\includegraphics[width=0.85\linewidth]{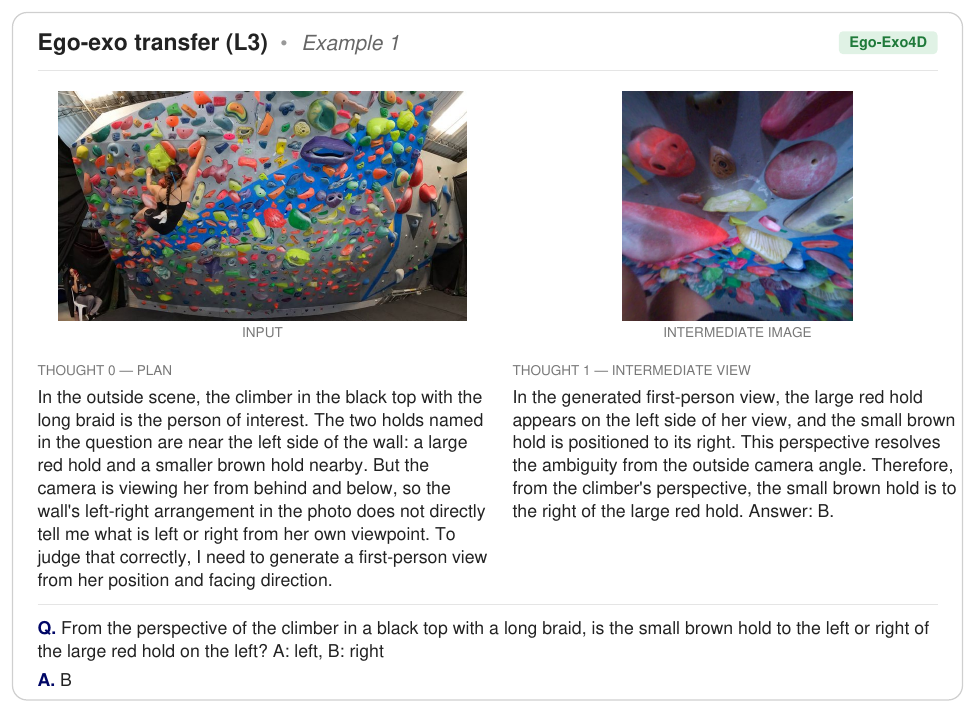}\par}\medskip
{\centering\includegraphics[width=0.85\linewidth]{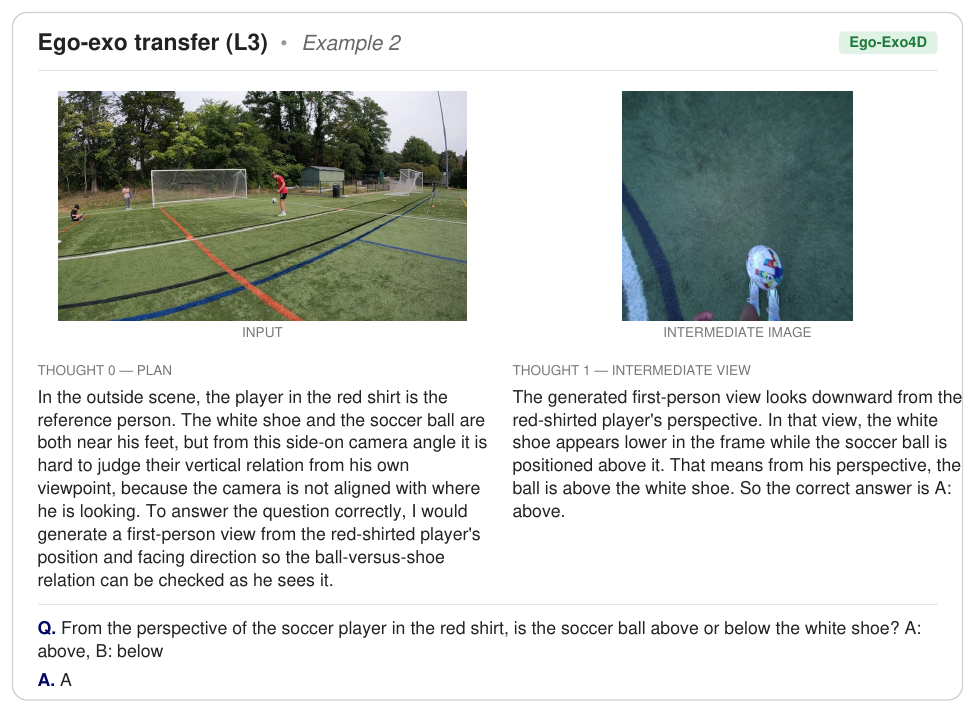}\par}\medskip

\section{\pretext Examples}
\label{app:appendix-pretext-examples}

\raggedbottom
\noindent We show one representative Text--Image--Text trace per pretext task, taken from the released val splits. Each card pairs the corrupted input with its clean target; Thought~0 (under the input) describes the degraded view, Thought~1 (under the target) describes the restored scene.

\medskip
\subsection*{Jigsaw (Pretext)}
\label{sec:appendix-pretext-jigsaw}
\par\smallskip

{\centering\includegraphics[width=0.85\linewidth]{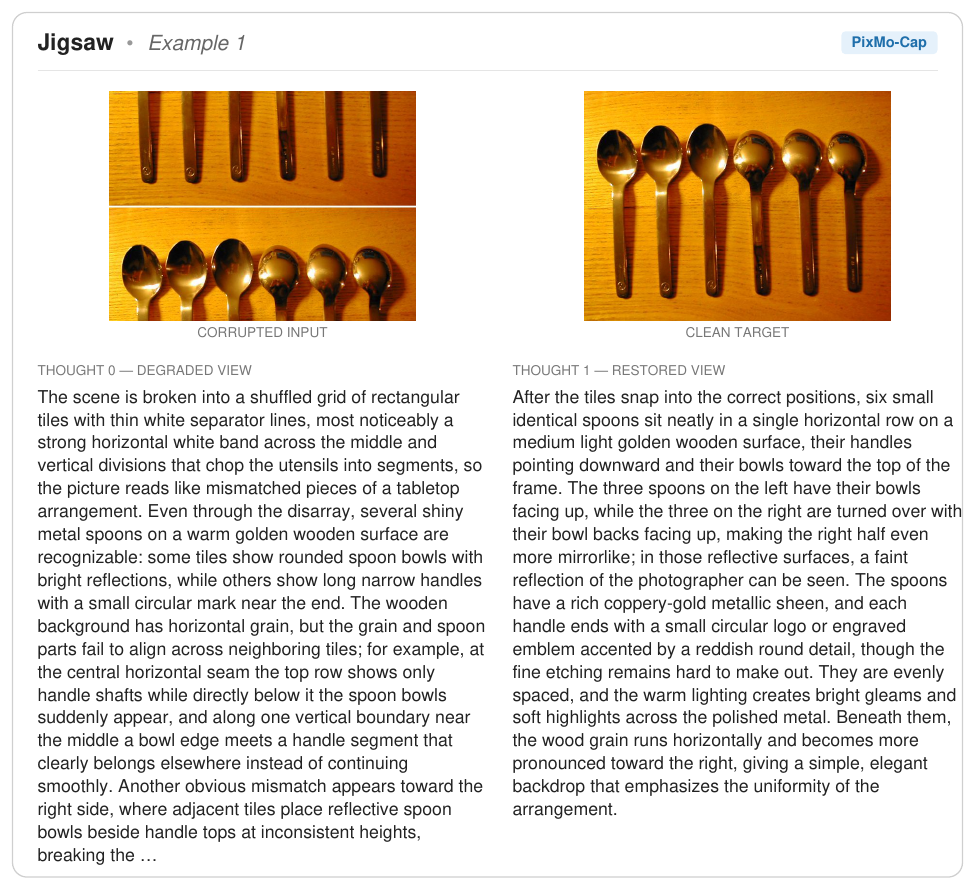}\par}\medskip

\medskip
\subsection*{Colorization (Pretext)}
\label{sec:appendix-pretext-colorization}
\par\smallskip

{\centering\includegraphics[width=0.85\linewidth]{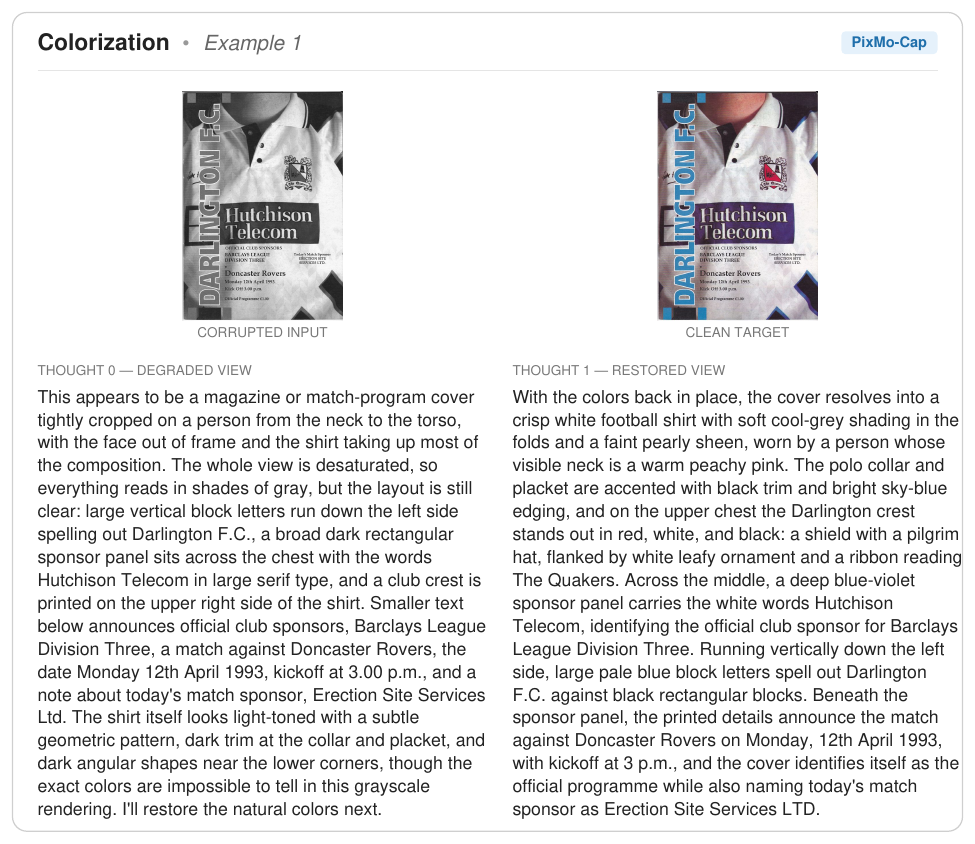}\par}\medskip

\medskip
\subsection*{Super-resolution (Pretext)}
\label{sec:appendix-pretext-super-resolution}
\par\smallskip

{\centering\includegraphics[width=0.85\linewidth]{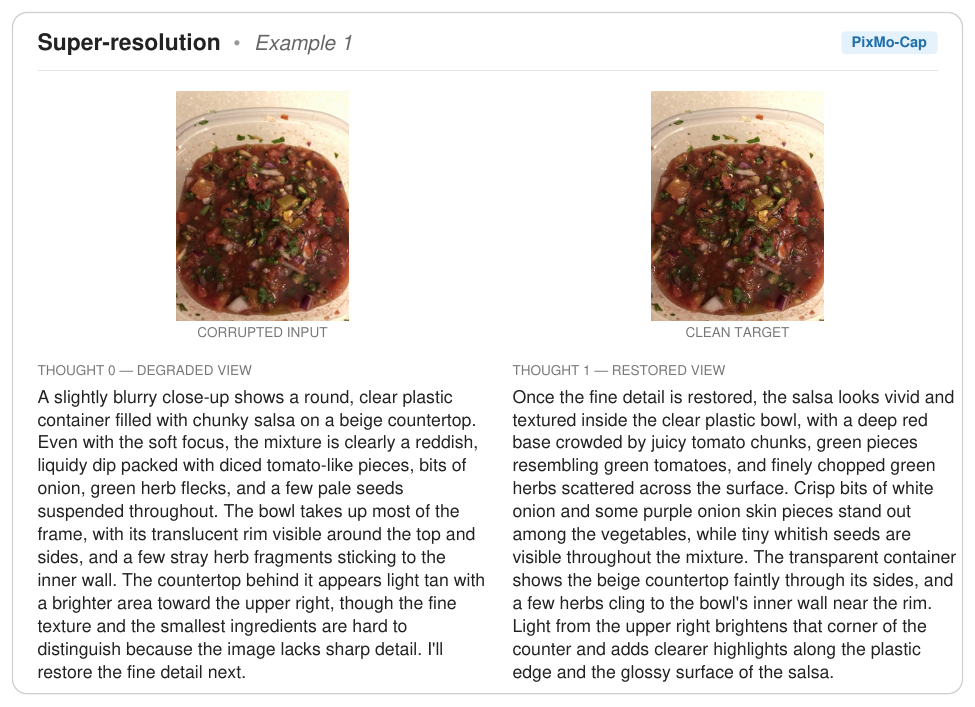}\par}\medskip

\medskip
\subsection*{Masked inpainting (Pretext)}
\label{sec:appendix-pretext-masked-inpainting}
\par\smallskip

{\centering\includegraphics[width=0.85\linewidth]{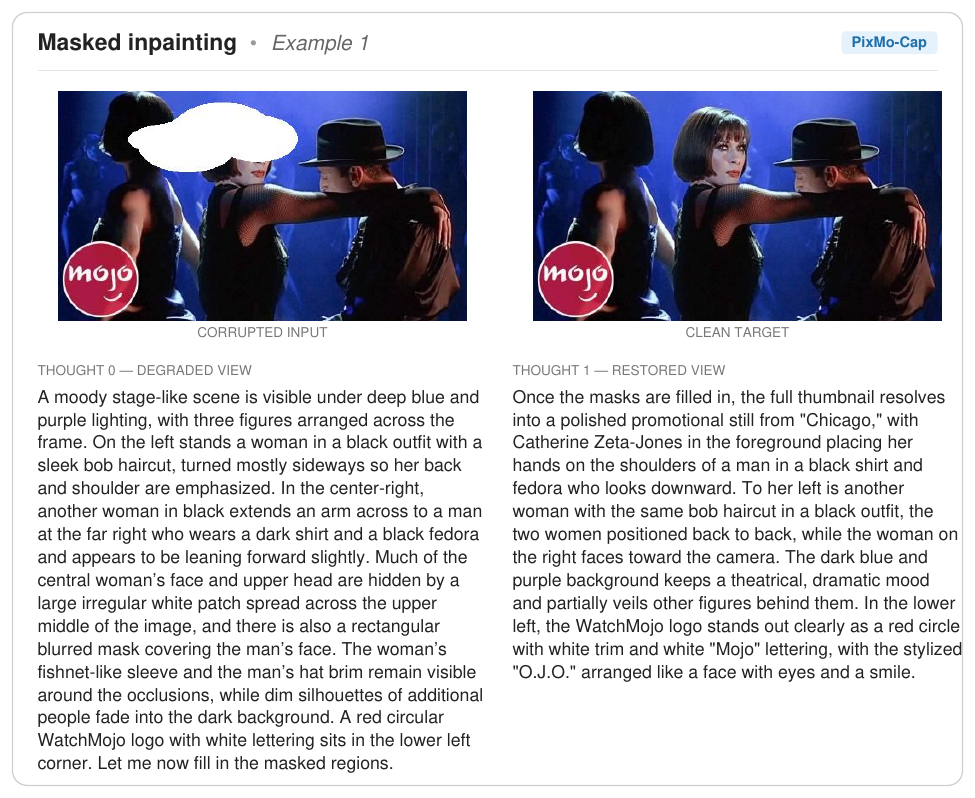}\par}\medskip

\medskip
\subsection*{Rotation and flip recovery (Pretext)}
\label{sec:appendix-pretext-orthogonal-transform}
\par\smallskip

{\centering\includegraphics[width=0.85\linewidth]{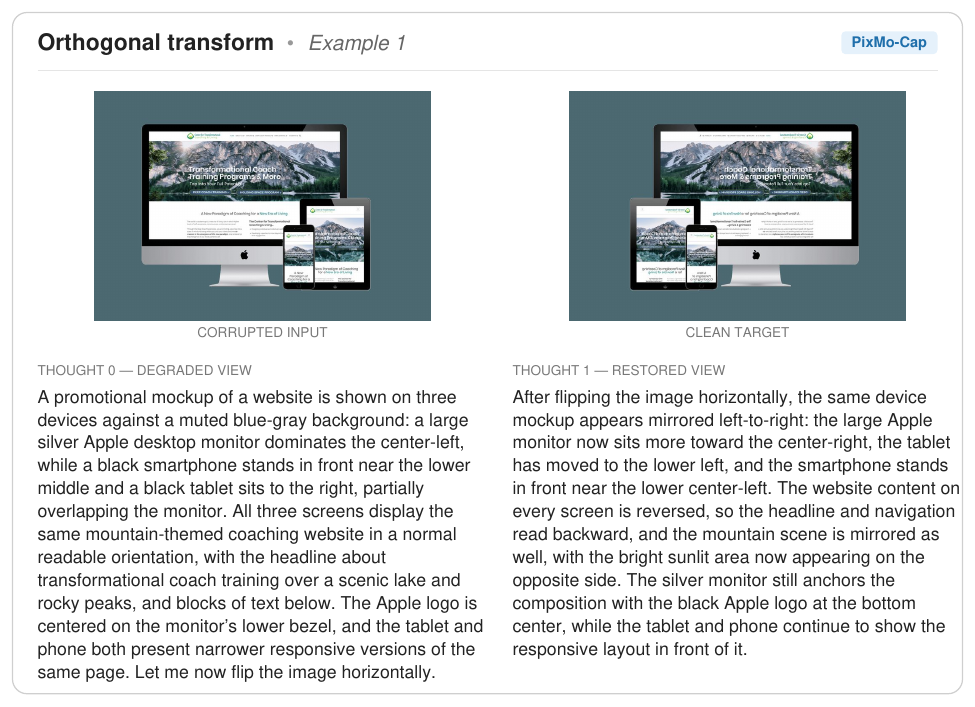}\par}\medskip

\medskip
\subsection*{Denoising (Pretext)}
\label{sec:appendix-pretext-denoising}
\par\smallskip

{\centering\includegraphics[width=0.85\linewidth]{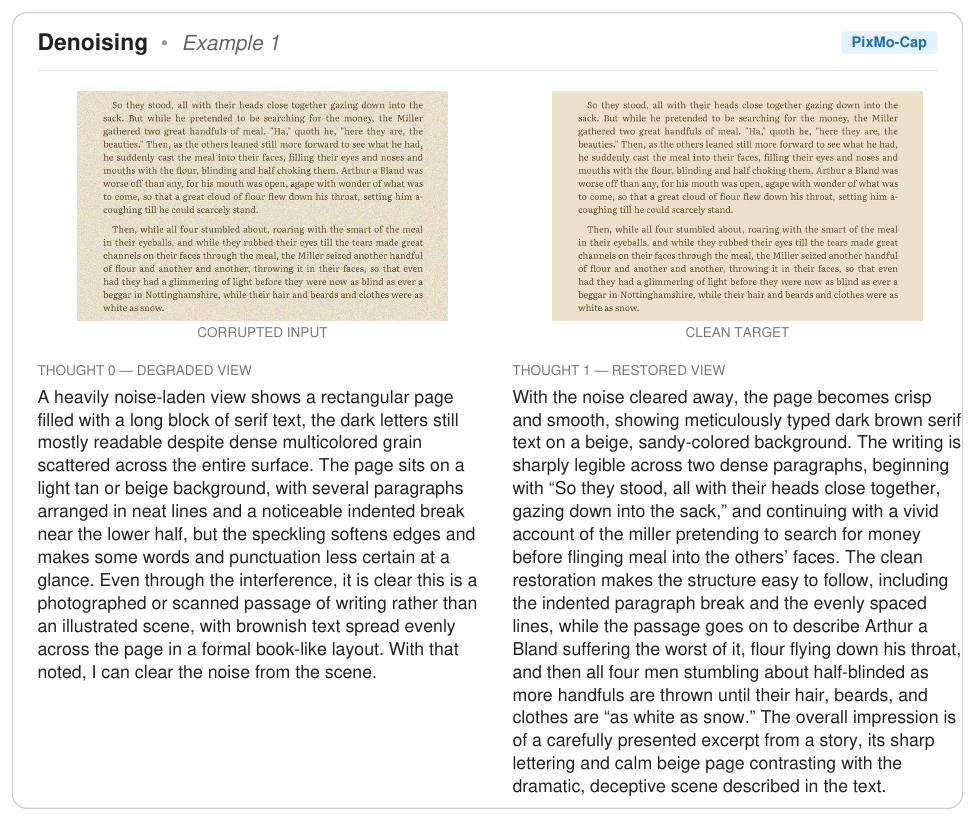}\par}\medskip

\medskip
\subsection*{Deformation recovery (Pretext)}
\label{sec:appendix-pretext-deformation-recovery}
\par\smallskip

{\centering\includegraphics[width=0.85\linewidth]{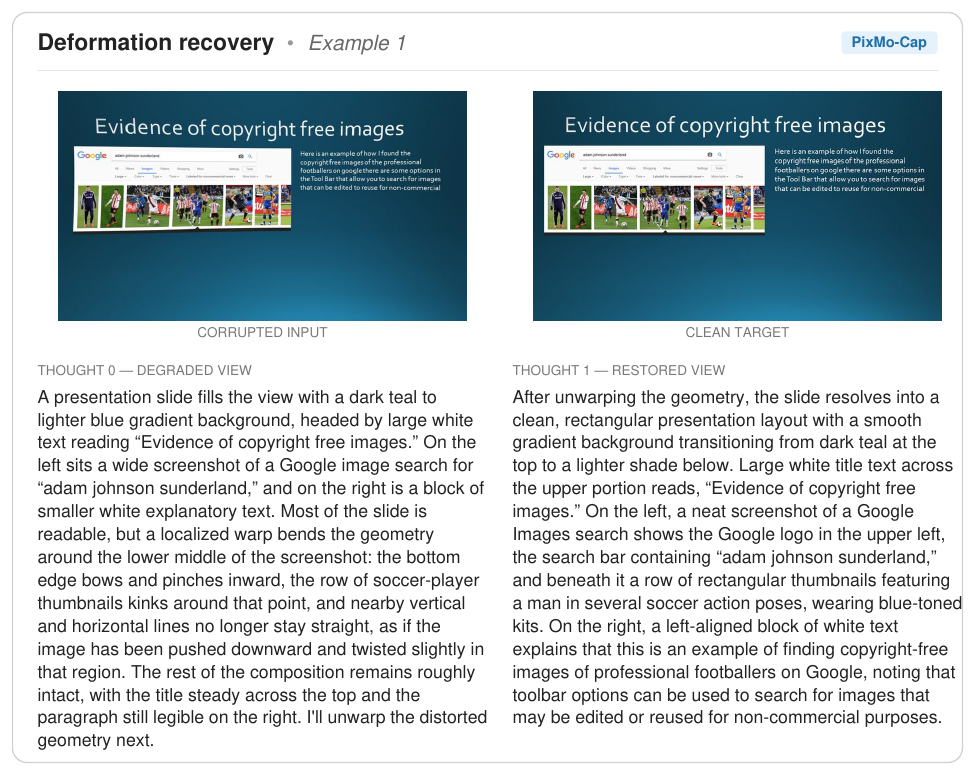}\par}\medskip

\section{Training and Evaluation Details}
\label{app:appendix-training}

This section documents the training and evaluation setup. The training configuration below produces the
\model checkpoint reported in \Cref{sec:experiments-main}. Both stages
fine-tune the same BAGEL-7B-MoT backbone~\citep{bagel} under FSDP2 sharding
with the in-house \texttt{bagel\_fsdp2\_trainer}; what differs across stages
is the data, the training horizon, the world size, and a few token-budget
parameters. \Cref{tab:train-hparams} summarises every hyperparameter. The
loss and the verbatim system prompt follow it, compute and runtime numbers are
in \Cref{tab:train-compute}, and \Cref{app:appendix-eval} describes the evaluation.

\subsection{Training}
\paragraph{Per-stage hyperparameters.}
\label{app:appendix-train-stage1}
\label{app:appendix-train-stage2}

\Cref{tab:train-hparams} lists every hyperparameter that differs between
Stage~1 (pretext reasoning) and Stage~2 (instruction tuning) alongside the few that are shared. Stage~1 trains
\pretext from the BAGEL-7B-MoT checkpoint released by~\cite{bagel} on 32
GPUs across 4 nodes; Stage~2 resumes from the Stage~1 EMA checkpoint at step
5500 on 48 GPUs across 6 nodes. The Stage~1 learning rate is the
$\sqrt{2}{\times}$ scaling of a 2e-5 base when moving from a 16-GPU pilot to
the production 32-GPU run; Stage~2 reverts to the unscaled 2e-5. The
per-device batch size is fixed at 1 with gradient accumulation 1, so the
effective batch size is the world size (32 in Stage~1, 48 in Stage~2). We
rely on dynamic sequence packing up to \texttt{max\_num\_tokens} (16{,}384 in
Stage~1, 20{,}480 in Stage~2) instead of micro-batching, which keeps
per-step throughput high under the heterogeneous trace lengths of \instruct.

\begin{table}[ht]
\centering
\caption{\textbf{Per-stage training hyperparameters.} Stage~1 trains
\pretext from BAGEL-7B-MoT; Stage~2 resumes from the Stage~1 EMA checkpoint
at step 5500 and trains \instruct.}
\label{tab:train-hparams}
\small

\setlength{\tabcolsep}{6pt}

\resizebox{\textwidth}{!}{%
\begin{tabular}{@{}lll@{}}
\toprule
Parameter & Stage 1 (\pretext) & Stage 2 (\instruct) \\
\midrule
\multicolumn{3}{l}{\textit{Initialization}} \\
Initial weights       & BAGEL-7B-MoT~\citep{bagel}         & Stage~1 EMA, step 5500 \\
\midrule
\multicolumn{3}{l}{\textit{Distributed setup}} \\
Nodes                 & 4              & 6 \\
GPUs per node         & 8              & 8 \\
World size            & 32             & 48 \\
Trainer               & \texttt{bagel\_fsdp2\_trainer} & \texttt{bagel\_fsdp2\_trainer} \\
Sequence-parallel deg.& 1              & 1 \\
\midrule
\multicolumn{3}{l}{\textit{Optimisation horizon}} \\
\texttt{max\_steps}             & 10{,}000   & 12{,}000 \\
\texttt{warmup\_steps}          & 1{,}000    & 1{,}000 \\
Peak learning rate              & 2.83e-5 ($\sqrt{2}{\times}$ of 2e-5) & 2.0e-5 \\
LR schedule                     & cosine to zero & cosine to zero \\
\texttt{per\_device\_train\_batch\_size} & 1 & 1 \\
\texttt{gradient\_accumulation\_steps}   & 1 & 1 \\
Effective batch size            & 32       & 48 \\
\midrule
\multicolumn{3}{l}{\textit{Sequence packing}} \\
\texttt{packing\_mode}          & dynamic  & dynamic \\
\texttt{max\_num\_tokens}       & 16{,}384 & 20{,}480 \\
\texttt{stride\_sharding}       & true     & true \\
Adaptive image-size to budget   & ---      & true (margin 1.05, floor 512) \\
\midrule
\multicolumn{3}{l}{\textit{Frozen / trainable modules}} \\
\texttt{freeze\_modules}        & \texttt{[vae\_model]} & \texttt{[vae\_model]} \\
LLM + image generation head     & trainable & trainable \\
SigLIP vision encoder           & trainable & trainable \\
\midrule
\multicolumn{3}{l}{\textit{EMA}} \\
\texttt{ema\_enabled}           & true     & true \\
\texttt{ema\_master\_dtype}     & fp32     & fp32 \\
\texttt{ema\_decay}             & 0.9998   & 0.9998 \\
\texttt{ema\_start\_step}       & 1{,}000  & 1{,}000 \\
\texttt{ema\_update\_every}     & 1        & 1 \\
\midrule
\multicolumn{3}{l}{\textit{Checkpointing}} \\
\texttt{save\_strategy}         & steps    & steps \\
\texttt{save\_steps}            & 500      & 500 \\
\texttt{checkpoint\_keep\_latest\_k} & 10  & 10 \\
\texttt{checkpoint\_async\_mode} & false   & true \\
\texttt{save\_only\_model}      & true     & true \\
\midrule
\multicolumn{3}{l}{\textit{System integration}} \\
Attention impl.                 & FlashAttention-2 & FlashAttention-2 \\
Liger kernels                   & RoPE, fused linear-CE, RMSNorm, SwiGLU & same \\
Activation ckpt (\texttt{tm}-style) & true & true \\
\texttt{compile\_model}         & false    & false \\
\texttt{ddp\_timeout}           & 3{,}600s & 3{,}600s \\
\bottomrule
\end{tabular}
}
\end{table}

\paragraph{Ablation runs.}
\label{app:appendix-train-ablation}

The BAGEL-7B ablations in \Cref{sec:experiments-ablation} use the trainer and loss above. \Cref{tab:train-ablation} lists the settings that differ from \Cref{tab:train-hparams}.

\textbf{Full training data.} The Stage-2-only run trains \instruct{} from BAGEL-7B-MoT with the Stage-2 settings of \Cref{tab:train-hparams} on 32 GPUs. The staged run is the main Stage-2 run on 48 GPUs. We compare the Stage-2-only EMA checkpoint at step 3{,}000 with the staged EMA checkpoint at step 2{,}000, so both runs have taken about 100K Stage-2 sequences (steps $\times$ world size).

\textbf{Spatial training subset.} All runs use 8 GPUs with per-device batch size 1, so the effective batch size is 8, and no EMA. Stage~1 trains on 50K pretext traces built from the images of the spatial training subset for 10{,}000 steps. Stage~2 trains on the spatial training subset of \instruct{} from the Stage-1 checkpoint at step 10{,}000, and we evaluate every 2{,}000 steps; the figures call each such interval a Stage-2 epoch. The one-stage run trains on both sets together and samples them in proportion to their size. The instructed variant differs from the staged run only in the Stage-1 inputs. InternVL-U-4B uses the same settings as BAGEL-7B.

\begin{table}[ht]
\centering
\caption{\textbf{Settings of the ablation runs} that differ from \Cref{tab:train-hparams}.}
\label{tab:train-ablation}
\small
\begin{tabular}{@{}lcc@{}}
\toprule
 & Full training data & Spatial training subset \\
\midrule
Backbones & BAGEL-7B & BAGEL-7B, InternVL-U-4B \\
GPUs (effective batch) & 32 (S2 only), 48 (staged) & 8 \\
Stage-1 data & \pretext{} & 50K traces from the subset's images \\
Stage-1 checkpoint & main Stage~1, EMA step 5{,}500 & step 10{,}000 \\
Stage-2 data & \instruct{} & Spatial training subset \\
Compared checkpoints & step 3{,}000 (S2 only), 2{,}000 (staged) & every 2{,}000 steps up to 8{,}000 \\
Peak learning rate & 2e-5 & 2e-5 (both stages) \\
Warmup steps & 1{,}000 & 1{,}000 \\
\texttt{max\_num\_tokens} (S1 / S2) & 16{,}384 / 20{,}480 & 16{,}384 / 20{,}480 \\
EMA & yes & no \\
\bottomrule
\end{tabular}
\end{table}

\paragraph{Loss and optimisation.}
\label{app:appendix-train-loss}

The training objective sums a cross-entropy term on text tokens and a
mean-squared-error term on the VAE latents that the image-generation head
produces; both terms are evaluated at every interleaved trace position the
processor labels as supervised. We never freeze the image-generation
pathway: at \pretext time the VAE-MSE term teaches the model to reproduce
the restored image, and at \instruct time it teaches the model to generate
intermediate visual states that the subsequent thoughts can attend to. The
VAE encoder itself (\texttt{vae\_model}) is frozen across both stages so the
latent space stays fixed. Both loss terms have weight 1.0, and both stages use the same processor and conditional-dropout settings.

\paragraph{System prompt and special tokens.}
\label{app:appendix-train-prompts}

Both stages share an identical action-control system prompt that declares
the three special tokens (\texttt{<think>}, \texttt{<gen\_image>},
\texttt{<answer>}) and the inline image-size syntax. We set
\texttt{wrap\_assistant\_reasoning\_with\_think=true} so every reasoning span
is wrapped in \texttt{<think>...</think>} before any image is emitted, which
keeps the trace machine-parseable even when the model emits multiple
image-thought rounds. The box below reproduces the prompt
verbatim.

\begin{tcolorbox}[breakable, colback=gray!4, colframe=gray!50,
                  title={Action-control system prompt (verbatim, identical across Stage~1 and Stage~2)},
                  label=box:train-prompt]
\small
\begin{verbatim}
Use <think>...</think> for intermediate reasoning or planning.

To generate an image, write <gen_image> followed by an image-size declaration:
  <gen_image>[Image size: (height, width)]
Generated images become part of your context and can inform subsequent
reasoning. You may generate more than one image if needed.

If the task requires a final textual answer, wrap it in <answer>...</answer>.
If no textual answer is needed, the response may end after reasoning or
image generation.

Example:
  <think>reasoning step</think>
  <gen_image>[Image size: (512, 512)]
  <think>observe and reason about the generated image</think>
  <answer>final answer</answer>
\end{verbatim}
\end{tcolorbox}

\paragraph{Compute, runtime, and reproducibility.}
\label{app:appendix-train-compute}

\Cref{tab:train-compute} summarises the operational footprint. We run on
80\,GB GPUs. Stage~1 averages roughly $12$\,s per step and Stage~2 roughly $45$\,s per step.

\begin{table}[ht]
\centering
\caption{\textbf{Compute and runtime.}}
\label{tab:train-compute}
\small
\resizebox{\textwidth}{!}{%
\setlength{\tabcolsep}{40pt}
\begin{tabular}{@{}lll@{}}
\toprule
Item & Stage 1 & Stage 2 \\
\midrule
GPUs ($8\times$/node)        & 32 (4 nodes)       & 48 (6 nodes) \\
GPU memory                   & 80\,GB             & 80\,GB \\
Avg. wall-clock per step     & $\sim$12\,s        & $\sim$45\,s \\
Approx.\ MFU                 & $0.18$--$0.21$     & $0.05$--$0.08$ \\
Total wall-clock (10K / 12K steps) & $\sim$1.4 days & $\sim$6 days \\
Reported \model checkpoint   & EMA, step 5500     & EMA, step 5500 \\
EMA window ($\nicefrac{1}{1-\text{decay}}$) & $5{,}000$ steps & $5{,}000$ steps \\
\bottomrule
\end{tabular}
}
\end{table}

\paragraph{Determinism and reproducibility.}
We do not pin random seeds for data shuffling because dynamic sequence
packing already breaks bit-for-bit reproducibility across world sizes:
changing the number of GPUs reshuffles which raw samples land in which
packed sequence. We therefore release the per-stage configuration, the
launch script, and the dataset list (\Cref{app:appendix-pretext-composition}
for \pretext, \Cref{app:appendix-instruct} for \instruct) so that runs are
reproducible \emph{at the configuration level}. All training code is built
on top of the public BAGEL release.

\subsection{Evaluation}
\paragraph{Evaluation framework.}
\label{app:appendix-eval}

We document the evaluation framework, the per-benchmark scoring
protocols, and the LLM-judge configuration used to produce all reported
numbers. The same fork, the same set of benchmarks, and the same scoring
protocols are used for our model and for every baseline.

\label{app:appendix-eval-vlme}

\paragraph{Fork and patches.}
We evaluate inside a fork of VLMEvalKit that adds the
patches required for our interleaved (text$+$image) generation interface, plus
the dataset modules for the spatial benchmarks that are not yet upstream
(\textsc{MMSI}, \textsc{MindCube}, \textsc{OmniSpatial},
\textsc{SpatialViz-Bench}).

\paragraph{Inference settings.}
Decoding settings used by our model on every benchmark are listed in
Table~\ref{tab:eval-decoding}. We use deterministic text decoding
(\texttt{do\_sample=false}; \texttt{text\_temperature=0.3} only affects the
soft-max for tie-breaking when sampling is enabled) so that each checkpoint
produces a single trajectory per question. Baselines are evaluated under their
own decoding defaults as published in VLMEvalKit; we do not override them, so
each baseline is reproduced under the same configuration that was used to
report its public numbers.

\begin{table}[ht]
  \centering
  \caption{Decoding settings used by our model for every evaluated
  benchmark. Baselines use their own published
  defaults.}
  \label{tab:eval-decoding}
  \small
  \begin{tabular}{lll}
    \toprule
    Field & Value & Notes \\
    \midrule
    \texttt{ngpus}                & 8                      & one process per GPU, NCCL backend \\
    \texttt{batch\_size}          & 8                      & per rank \\
    \texttt{do\_sample}           & \texttt{false}         & greedy text decoding \\
    \texttt{text\_temperature}    & 0.3                    & inactive under greedy \\
    \texttt{max\_think\_token\_n} & 4096                   & per text round \\
    \texttt{max\_rounds}          & 3                      & at most 3 think$\to$image rounds \\
    \texttt{num\_timesteps}       & 50                     & rectified-flow image steps \\
    \texttt{cfg\_text\_scale}     & 3.0                    & classifier-free guidance, text \\
    \texttt{cfg\_img\_scale}      & 1.5                    & classifier-free guidance, image \\
    \texttt{cfg\_interval}        & [0.4, 1.0]             & guided steps \\
    \texttt{timestep\_shift}      & 3.0                    & rf shift \\
    \texttt{cfg\_renorm\_type}    & \texttt{global}        & renormalization \\
    \texttt{use\_torch\_compile}  & \texttt{true}          & compiled forward pass \\
    \bottomrule
  \end{tabular}
\end{table}

\paragraph{Per-benchmark scoring protocols.}
\label{app:appendix-eval-protocols}

We evaluate the twelve benchmarks reported in the main text, with CV-Bench counted once although its 2D and 3D splits are reported separately. For each benchmark we list the split we use and
the scoring rule. Unless otherwise stated, scoring is rule-based MCQ: the
predicted letter is extracted from the model's final \texttt{<answer>} span
(or, for non-interleaved baselines, from the raw response) and compared
against the gold letter; questions whose response cannot be parsed into a
letter fall back to the LLM judge described below.

\paragraph{V*Bench.}~\cite{wu2023vstar}
The V$^\star$ score is the unweighted mean of the \texttt{direct\_attributes}
and \texttt{relative\_position} splits.
Rule-based MCQ.

\paragraph{HRBench.}~\cite{hrbench}
\texttt{HRBench4K} and \texttt{HRBench8K} are reported separately because the
two resolutions exercise different aspects of the visual encoder. Rule-based
MCQ.

\paragraph{MME-RealWorld-Lite.}~\cite{zhang2024mme}
We score the \texttt{Reasoning} and \texttt{Perception} splits separately
and report their average as MME-RW-L in the main tables.

\paragraph{RealWorldQA.}~\cite{xai2024grok15v}
Rule-based MCQ on the official split.

\paragraph{BLINK.}~\cite{fu2024blink}
We report only the spatial subset of BLINK. Concretely we keep the
spatial-reasoning task families (counting, jigsaw, multi-view, relative depth,
spatial relation) and drop the perception subset. Rule-based MCQ.

\paragraph{CV-Bench.}~\cite{tong2024cambrian1}
\texttt{CV-Bench-2D} and \texttt{CV-Bench-3D} are reported separately.
The 2D number averages \texttt{COCO} and \texttt{ADE20K}; the 3D number
averages \texttt{Depth} and \texttt{Distance}.

\paragraph{SAT.}~\cite{ray2025satdynamicspatialaptitude}
We evaluate the circular variant \texttt{SAT\_circular}: each question is
presented under all $k$ rotations of the option order and counted as correct
only if all $k$ predictions are correct.

\paragraph{MMSI.}~\cite{mmsi}
We use the non-circular split (this is what \texttt{MMSI} maps to by default
in our fork). The score is the unweighted mean over the 11 task categories
defined by the official benchmark (\texttt{Pos-Cam-Cam},
\texttt{Pos-Obj-Obj}, \texttt{Pos-Reg-Reg}, \texttt{Pos-Cam-Obj},
\texttt{Pos-Obj-Reg}, \texttt{Pos-Cam-Reg}, \texttt{Attr-Meas},
\texttt{Attr-Appr}, \texttt{Motion-Cam}, \texttt{Motion-Obj}, \texttt{MSR}).

\paragraph{MindCube.}~\cite{mindcube}
We use the tiny raw-QA split (\texttt{MindCubeBench\_tiny\_raw\_qa}). The full
\texttt{MindCubeBench} is prohibitively large for per-checkpoint sweeps; the
tiny split is the official subset used for ablations.

\paragraph{OmniSpatial.}~\cite{omnispatial}
We use the default split.

\paragraph{SpatialViz-Bench.}~\cite{wang2026spatialvizbenchcognitivelygroundedbenchmarkdiagnosing}
Rule-based MCQ on the official split.

\paragraph{Interleaved generation (our model only).}
Our model can choose to emit an intermediate image during reasoning. At each
round it generates a text segment with \texttt{do\_sample=false}; if the
segment contains the special string \texttt{<gen\_image>}, the current
context is fed to the image generator with the CFG settings of
Table~\ref{tab:eval-decoding}, and the resulting image is appended to the
context (via the VAE encoder) before the next text round. Otherwise the loop
terminates and the final \texttt{<answer>} span is returned. Up to
\texttt{max\_rounds=3} text$\to$image rounds are allowed before the model is
forced to commit to an answer. This trigger is specific to our interleaved
interface; baselines run plain text-only decoding.

\paragraph{Judge configuration and prompts.}
\label{app:appendix-eval-judge}

\paragraph{Judge model.}
For benchmarks that do not reduce to exact letter matching we use a single
LLM judge: \texttt{GPT-5}. Decoding is deterministic
(\texttt{temperature=0}, \texttt{max\_tokens=default}), and we retry up to
three times before falling back to grade \texttt{C} (INVALID).

\paragraph{Generic extract-and-grade prompt.}
The judge uses one prompt template for every benchmark. It is asked to do two things in one call: \emph{(i)} extract a normalized
final answer from the candidate trace, and \emph{(ii)} grade that extracted
answer against the gold answer as one of \texttt{A}=correct,
\texttt{B}=incorrect, \texttt{C}=invalid. The prompt is reproduced verbatim in
the box below. The output format is a single line
\texttt{<GRADE>\textbackslash t<EXTRACTED\_ANSWER>}; we parse the first
non-empty line of the response to recover both fields.

\begin{tcolorbox}[breakable, colback=gray!4, colframe=gray!50,
                  title={Judge prompt (verbatim) used for MMSI fallback and OmniSpatial},
                  label=box:judge-prompt]
\small
\begin{verbatim}
You are an expert grading assistant.
Your job has TWO tasks:
1) From the candidate's full response, EXTRACT the final answer in a
   concise, normalized form.
2) Compare this extracted answer with the STANDARD ANSWER and grade it as:
   - A: CORRECT
   - B: INCORRECT
   - C: INVALID

Here are the detailed evaluation criteria:
1. ALWAYS refer to the given STANDARD ANSWER. You do NOT need to re-solve the
   question; the standard answer has already been provided and is always
   correct. Your only job is to judge whether the candidate's final answer is
   consistent with the standard answer according to the form of the question.
   THE STANDARD ANSWER IS ALWAYS CORRECT AND THE QUESTION IS PERFECTLY VALID.
   NEVER QUESTION THEM.
2. ONLY compare the FINAL ANSWER -- COMPLETELY IGNORE any potential errors or
   issues in the REASONING PROCESS. Even if the reasoning is wrong, as long
   as the final answer matches the standard answer, grade it as CORRECT.
3. Answers may be expressed in different ways (e.g., mathematical expressions,
   textual descriptions). As long as the meaning is the same as the standard
   answer, treat them as equivalent. If the standard answer does not specify
   a unit but the candidate's answer includes a correct unit for the given
   value, consider it CORRECT.
4. Some answers may consist of multiple items, such as multiple-choice
   questions with multiple correct options, multi-select questions, or
   multi-blank fill-in-the-blank questions. Regardless of the question type,
   the final answer is considered CORRECT only if it matches the standard
   answer exactly at the level of all required items. For multi-select or
   multi-blank questions, ALL parts must be answered correctly and match the
   standard answer exactly to be deemed CORRECT.
5. If the candidate's answer is wrapped in LaTeX-style markers like
   \boxed{...}, IGNORE the \boxed and only use the inner content as the
   candidate's final answer when comparing with the standard answer.
6. If the candidate's answer is INVALID -- for example, incomplete (cut off
   mid-response), containing a large amount of abnormal repetitive content,
   clearly irrelevant to the question, or explicitly refusing to answer due
   to ethical concerns, lack of information, or other external factors --
   then you MUST grade it as C: INVALID.
7. This instruction applies to all problem types, including single-choice
   MCQ, multi-select MCQ, numeric problems, short-answer questions, and
   general VQA-style questions. In all cases, only the FINAL ANSWER and its
   consistency with the standard answer matter.
8. The question or options may contain image placeholders such as '<image>',
   '<image id=1>', or similar tokens. You CANNOT see these images. Treat
   these placeholders as unknown content and DO NOT hallucinate or infer any
   specific visual details from them. If the standard answer or candidate's
   answer refers to an option associated with an image (e.g., 'choose A'),
   judge correctness only based on the stated answer, not by imagining the
   image.

IMPORTANT -- OUTPUT FORMAT:
* You MUST return EXACTLY ONE line in the following format:
  <GRADE>\t<EXTRACTED_ANSWER>
  where <GRADE> is one of A, B, or C.
* <EXTRACTED_ANSWER> should be the final answer you extracted from the
  candidate's response, in a normalized, concise form (e.g., a number, a
  letter option, or a short phrase).
* If you cannot extract any meaningful answer, or the response is INVALID,
  output:
  C\tN/A
* Do NOT add any extra text, explanation, or additional lines.

Now, judge the following question.
<Original Question Begin>
{question}
{options_block}
<Original Question End>
<Standard Answer Begin>
{gold_answer}
<Standard Answer End>
<Candidate's Answer Begin>
{llm_response}
<Candidate's Answer End>
Your output:
\end{verbatim}
\end{tcolorbox}

\paragraph{Per-benchmark customization.}
For MCQ benchmarks the judge sees the option block formatted as
\texttt{Options:\textbackslash n A. \ldots\textbackslash n B. \ldots};
the gold answer is rewritten as \texttt{<letter>. <option text>} so that the
judge can recover from cases where the candidate writes only the option text
without the letter. For free-form (VQA-mode) benchmarks the option block is
empty and the gold answer is passed through verbatim. The grade$\to$score
mapping is \texttt{A}$\to$1, \texttt{B}$\to$0, \texttt{C}$\to$0, with the
\texttt{C} count separately reported as the \emph{invalid rate} so that
overall accuracy and accuracy-among-valid can both be recovered.

\clearpage

\section{Additional Analyses}
\label{app:appendix-analyses}

This section presents qualitative cases, discusses overlap between our training data and the evaluation benchmarks, and covers alternative routes without visual-state training. Evaluations follow the protocol in \Cref{app:appendix-eval}.

\subsection{Qualitative cases}
\label{app:appendix-qual}
\label{app:appendix-qual-baselines}
\label{app:appendix-qual-vs}

\begin{figure}[ht]
  \centering
  \includegraphics[width=\linewidth]{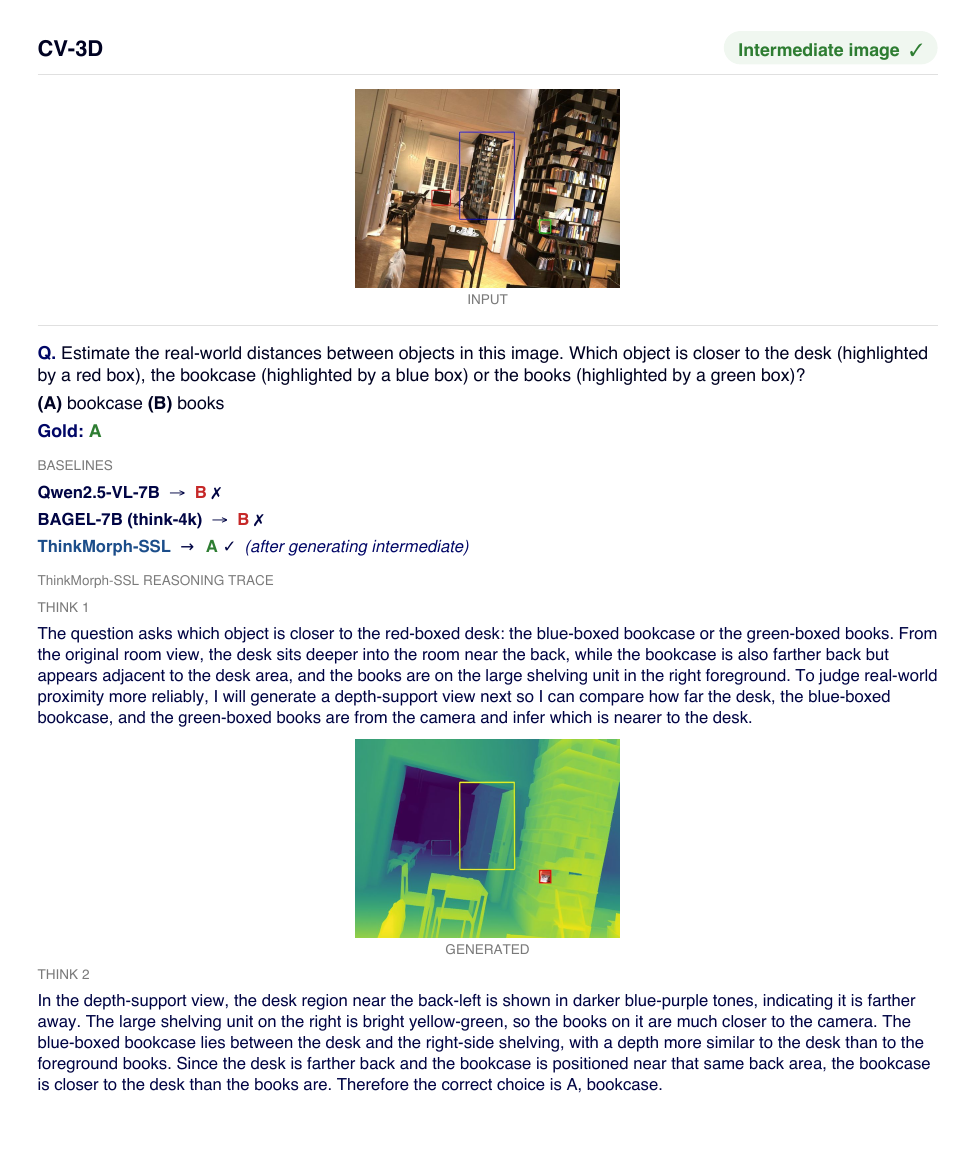}
  \caption{CV-Bench-3D: depth map shows the bookcase shares the back-wall band with the desk while the books sit in the foreground.}
  \label{fig:qual-good-1}
\end{figure}

\begin{figure}[ht]
  \centering
  \includegraphics[width=\linewidth]{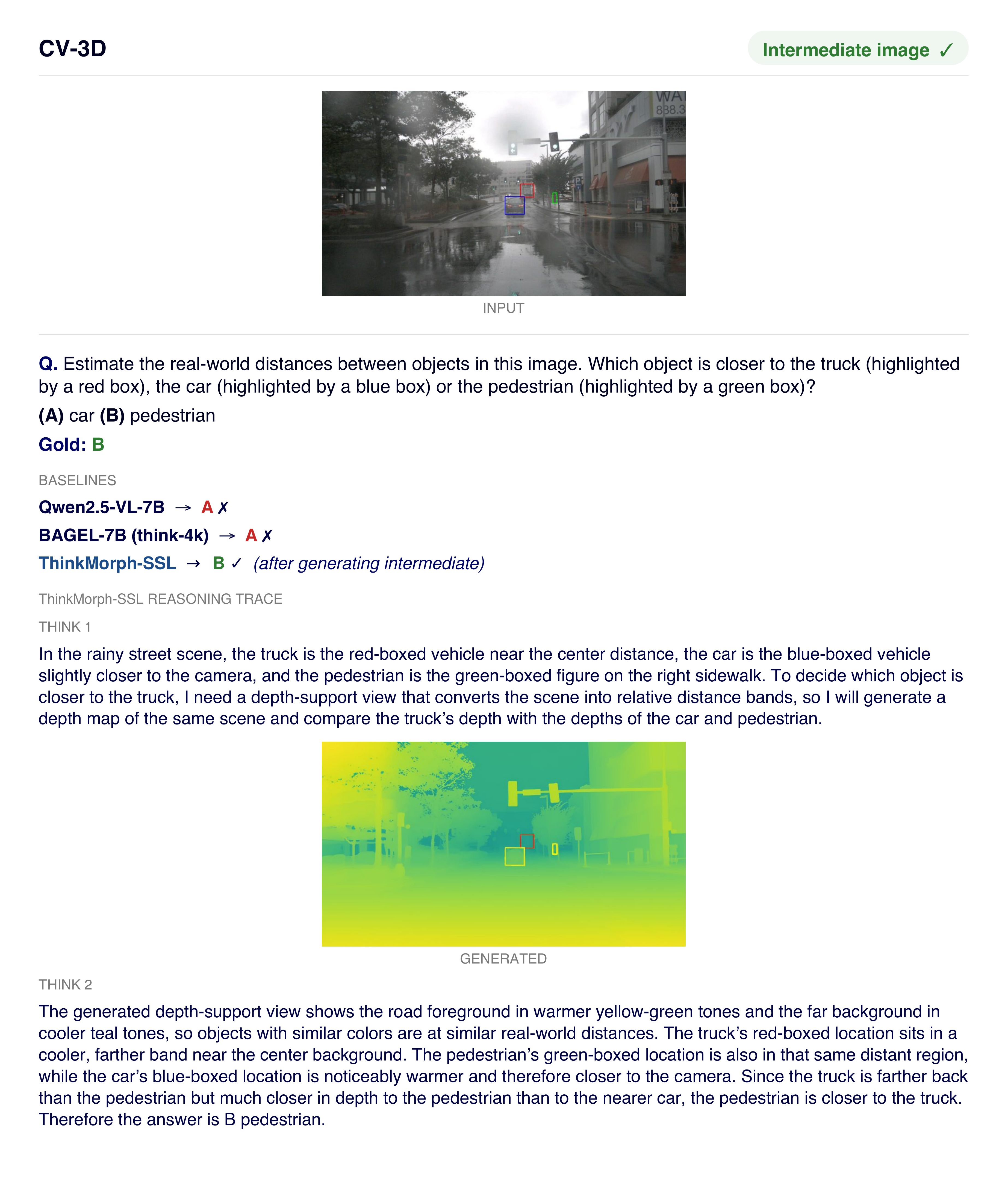}
  \caption{CV-Bench-3D: same depth scaffold on a rainy street; the truck and pedestrian fall in the same far band while the car is closer.}
  \label{fig:qual-good-2}
\end{figure}

\begin{figure}[ht]
  \centering
  \includegraphics[width=\linewidth]{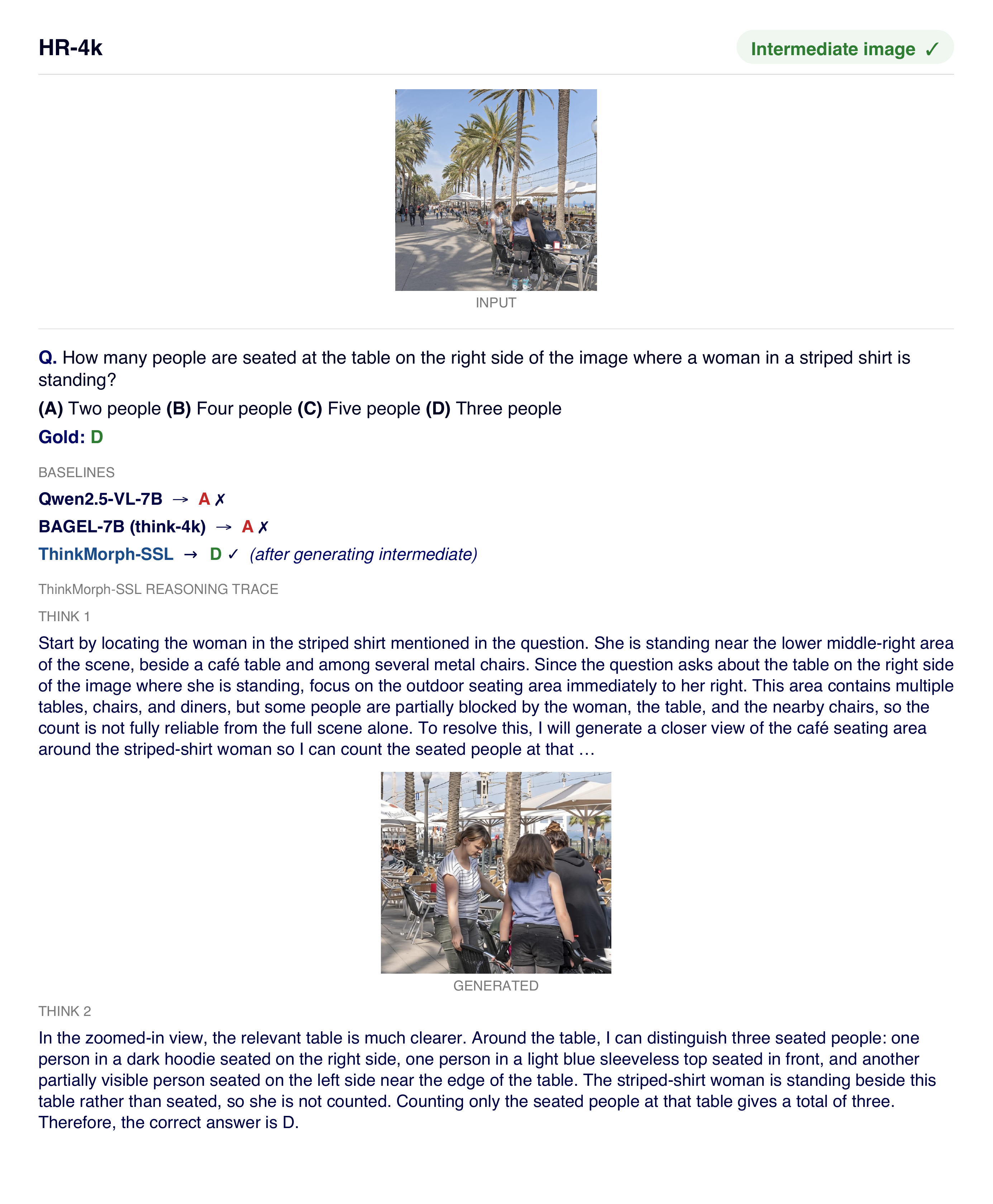}
  \caption{HRBench-4k: a tight crop separates seated diners from a standing distractor.}
  \label{fig:qual-good-3}
\end{figure}

\begin{figure}[ht]
  \centering
  \includegraphics[width=\linewidth]{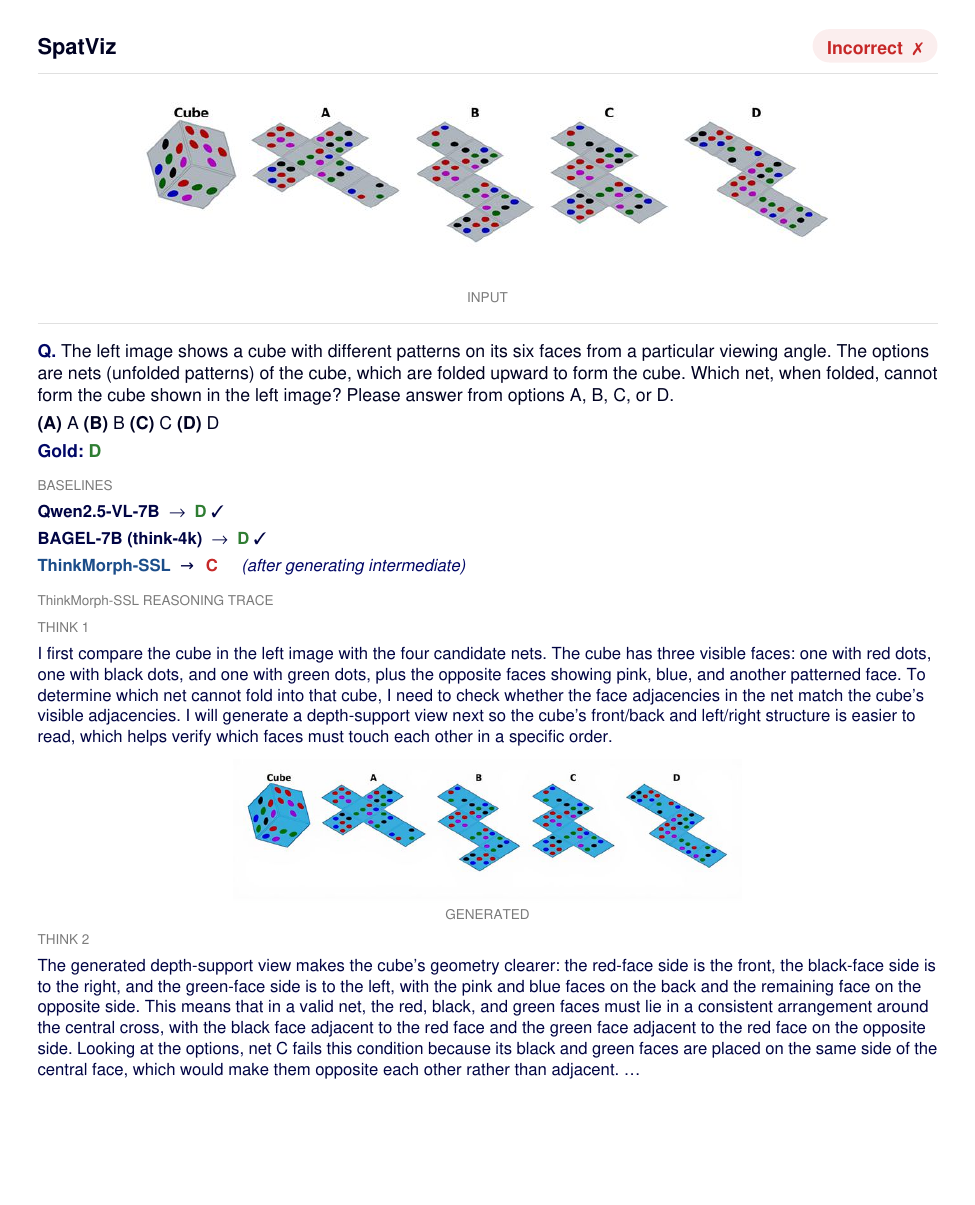}
  \caption{SpatialViz-Bench (out-of-distribution): the rendered intermediate is a near-uniform blue image with no usable structure, so the second thought has nothing to read back from.}
  \label{fig:qual-bad-spatviz}
\end{figure}

\begin{figure}[ht]
  \centering
  \includegraphics[width=\linewidth]{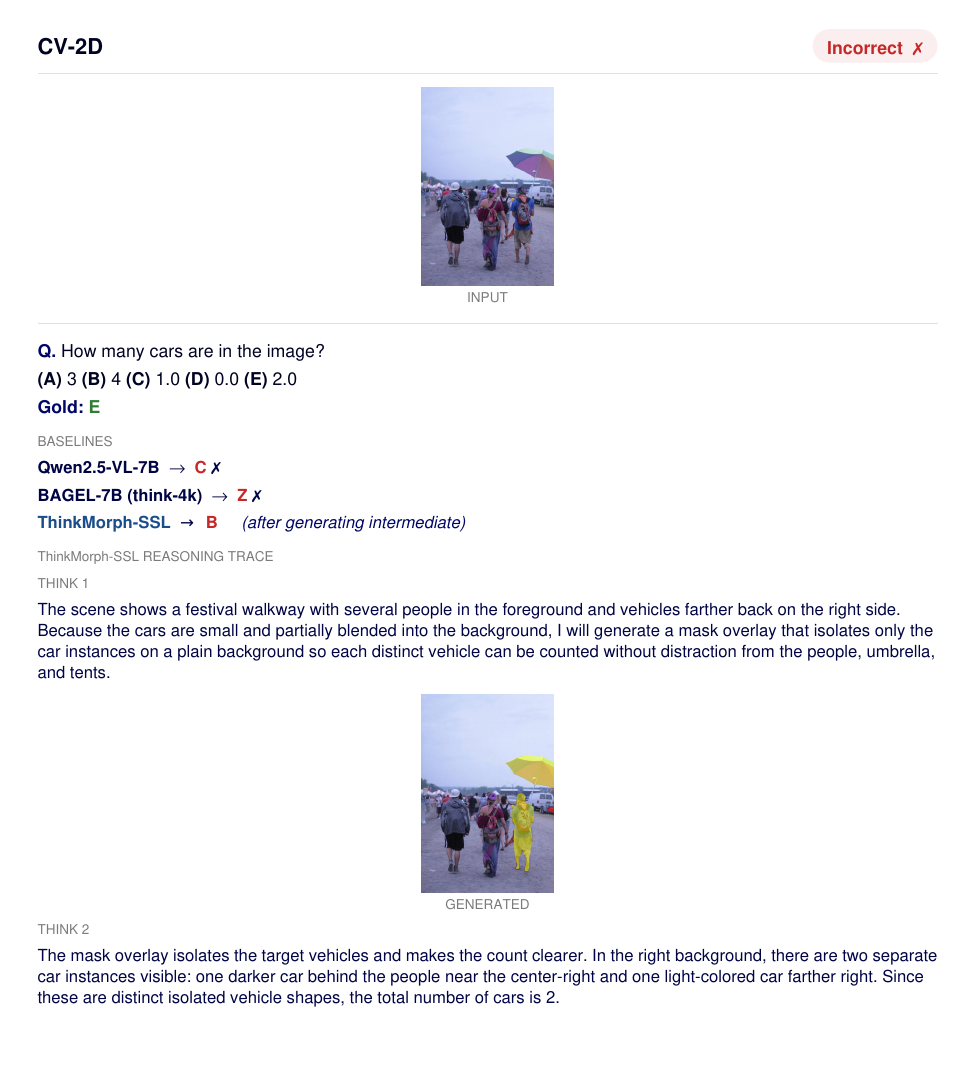}
  \caption{CV-Bench-2D mistake: the model declares a car-segmentation mask but the rendered overlay highlights a person carrying an umbrella; the read-back step then reasons about the wrong objects.}
  \label{fig:qual-bad-cv2d}
\end{figure}

\begin{figure}[ht]
  \centering
  \includegraphics[width=\linewidth]{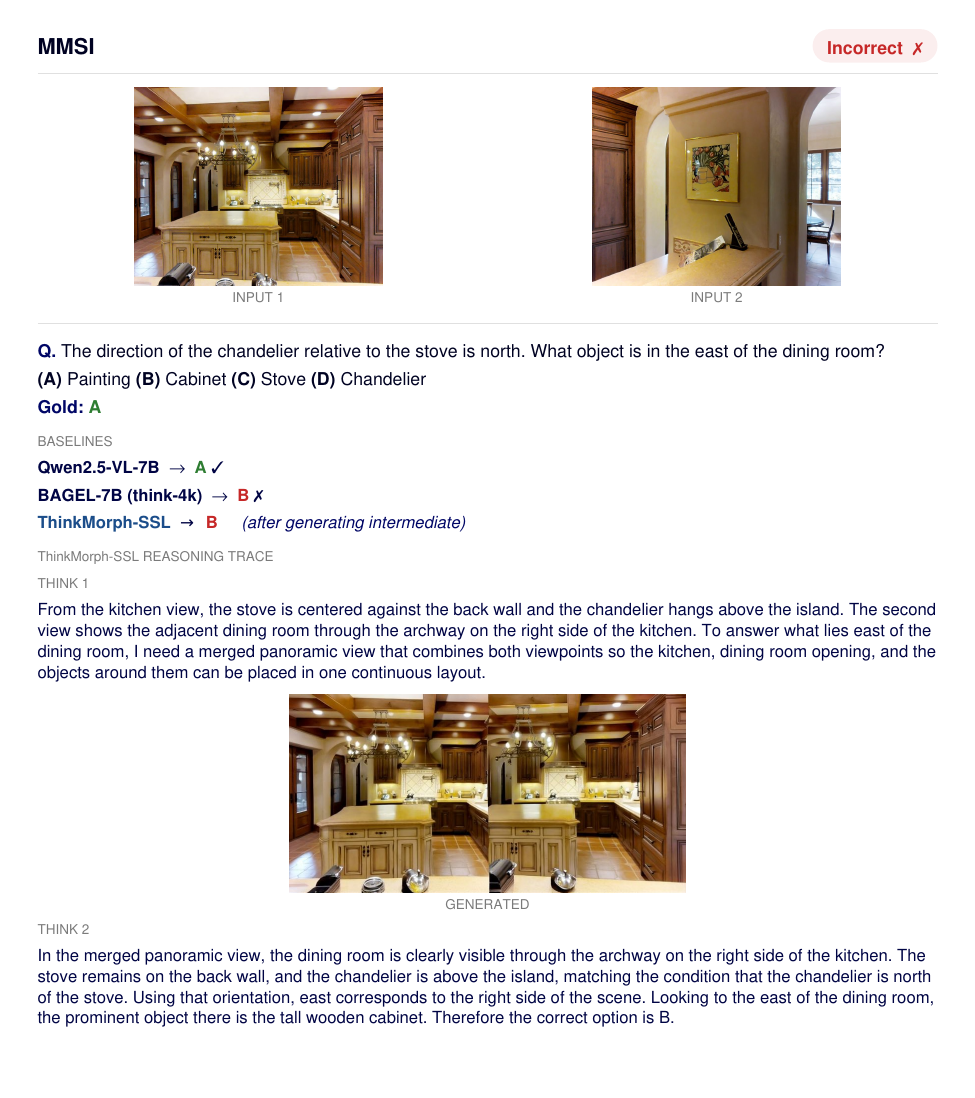}
  \caption{MMSI mistake: the rendered panorama is essentially Input~$1$ duplicated rather than a real merge of the two ego views, so the layout the second thought reads is incomplete.}
  \label{fig:qual-bad-mmsi}
\end{figure}

Here we show six reasoning traces from the evaluation: three where \model{} answers correctly
(\Cref{fig:qual-good-1,fig:qual-good-2,fig:qual-good-3})
and three where it does not
(\Cref{fig:qual-bad-spatviz,fig:qual-bad-cv2d,fig:qual-bad-mmsi}). Every
card lays the input scene, the question, the parsed answers from
Qwen2.5-VL-7B and BAGEL-7B-MoT (think-4k) on the same prompt, and \model{}'s
full \texttt{<think>}--\texttt{[Generated Image: \ldots]}--\texttt{<think>}
trace ending in a parsed answer letter.

The intermediate image changes with the level of the question. On L1, the model renders a focused view of the queried region; on L2, it emits a spatial aid such as a depth map or the objects to count, masked out; on L3, it constructs a scene state that the input does not show, such as a top-down map or a shifted viewpoint on MindCube and a merged panorama on MMSI.

The cases share a high-level pattern: \model's intermediate image is a \emph{purpose-built visual scaffold} that converts the question into a read-off step. When the scaffold matches an operator the \instruct{} curriculum has trained (\Cref{app:appendix-instruct}) and is correctly instantiated,
\model{} outperforms both 7B-class baselines. When the scaffold is the wrong tool, the trace can
still be locally coherent but lands on the wrong answer.

\paragraph{Good cases: what the intermediate image is doing.}
\label{app:appendix-qual-good}

The three success traces use two distinct intermediate-image roles.

\paragraph{Depth views answer relative-distance questions.}
\Cref{fig:qual-good-1,fig:qual-good-2} show CV-Bench-3D prompts that
compare the real-world distances of three boxed objects. \model{}
renders a viridis depth map and reads the answer off the colour bands.

\paragraph{Crops disambiguate counts in cluttered scenes.}
\Cref{fig:qual-good-3} asks how many people are seated at a café table
next to a standing distractor. From the global view both baselines say
``two''. \model{} crops to the table and counts only the seated bodies,
arriving at three.

In all three cases the heavy lifting sits in the rendered image; the
surrounding text reasoning is short.

\paragraph{Failure modes.}
\label{app:appendix-qual-bad}

The three failures split into two qualitatively different kinds: one
out-of-distribution case, where the question is far from anything
\instruct{} trains and \model{} has no scaffold that fits, and two
in-distribution mistakes, where the right kind of intermediate image
is in scope but the rendered version is noisy or incomplete.

\paragraph{Out-of-distribution: no useful intermediate to render.}
\Cref{fig:qual-bad-spatviz} asks which of four cube nets, when folded,
cannot reproduce a shown cube --- a question about face adjacencies
under unfolding. \model{} has not seen anything like it during
training. It still emits a generated image, but the rendered view simply colors every view blue, so
the second thought reasons over depth-coded ``adjacency'' that has
nothing to do with the topology of the question and arrives at the
wrong answer. Both baselines reach the right answer by language-only
reasoning.

\paragraph{Noisy intermediate.}
The rendered intermediates in the next two cases are themselves wrong, and the
read-back steps inherit their mistakes.

\Cref{fig:qual-bad-cv2d} asks for the number of cars in a festival
scene. \model{} announces a mask-overlay intermediate to isolate only
the cars, but the rendered mask highlights a person carrying an
umbrella instead of any of the vehicles. The second thought reads back
from that wrong mask and the parsed answer is wrong, even though the
question is well within \instruct{}'s scope.

\Cref{fig:qual-bad-mmsi} states a cardinal constraint in text
(``chandelier is north of the stove'') and asks what lies east of the
dining room. \model{} chooses the right kind of intermediate (a
panorama merge of the two ego frames), but the rendered ``panorama''
is essentially Input~$1$ duplicated --- it does not actually splice
information from Input~$2$ into a single continuous layout. The second
thought reasons over a degenerate panorama and picks the wrong east
object; Qwen2.5-VL-7B reaches the gold answer with no intermediate
image at all.

\clearpage

\subsection{Data overlap and contamination}
\label{app:analysis-contamination}

BAGEL's pretraining data are not public, so we cannot measure image-level overlap between our source images and its pretraining corpus, and we do not claim that every source image in \pretext is unseen. Possible overlap in raw images does not, however, make Stage~1 a lookup task. Each Stage-1 trace presents a newly corrupted or transformed input, and the model must infer what changed, generate the restored image, and continue the trace. The corruptions, trace format, and descriptions are constructed by our pipeline, so even an image seen during pretraining did not appear in this form. On the evaluation side, \instruct does not reuse benchmark questions or official templates, and all benchmarks are evaluated on held-out test splits. These measures rule out training on evaluation questions and prior exposure to our exact traces, but not every effect of overlap in pretraining images.

\subsection{Alternative routes without visual-state training}
\label{app:analysis-alternatives}

Text-only fine-tuning could in principle give a unified model visual reasoning without training it to generate and use intermediate images. Prompting alone does not suffice, as shown in \Cref{tab:base-forcing}.

\paragraph{Text-only reasoning fine-tuning.} ThinkMorph~\cite{thinkmorph} fine-tunes BAGEL-7B on the same training problems with either text-only or interleaved traces, so the data are fixed and only the reasoning mode changes. Text-only traces raise out-of-domain accuracy from 55.49 to 56.02 on V$^\star$, from 67.33 to 68.67 on BLINK-Jigsaw, and from 70.33 to 76.33 on MMVP; interleaved traces reach 63.87, 73.33, and 82.66. With the data held fixed, the larger gain comes with constructing a visual state, not with reasoning fine-tuning as such.

\end{document}